\def\paperID{457}
\def\confName{CVPR}
\def\confYear{2024}

\def\paperTitle{
\MethodName: Learning to Detect Images from Unseen \\ Text-to-Image Models by Inverting Stable Diffusion \vspace{-1em}
}

\def\authorBlock{
    George Cazenavette\textsuperscript{1}\thanks{Work done during an internship at Google Research.} 
    \qquad
    Avneesh Sud\textsuperscript{2}
    \qquad
    Thomas Leung\textsuperscript{2}
    \qquad
    Ben Usman\textsuperscript{2}\vspace{0.5em}
    \\
    \textsuperscript{1}Massachusetts Institute of Technology 
    \qquad
    \textsuperscript{2}Google Research\vspace{0.5em}
    \\
    \href{http://fake-inversion.github.io}{fake-inversion.github.io}
}

\newif\ifreview 
\newif\ifarxiv \newcommand{\arxiv}{\arxivtrue}
\newif\ifcamera 
\newif\ifrebuttal

\arxiv

\documentclass[10pt,twocolumn,letterpaper]{article}

\usepackage{abstract}

\newcommand{\eg}[1]{test}

\ifreview \usepackage[review]{cvpr} \fi
\ifarxiv \usepackage[pagenumbers]{cvpr} \fi
\ifarxiv \usepackage{cvpr} \fi
\ifrebuttal \usepackage[rebuttal]{cvpr} \fi
\ifcamera \usepackage{cvpr} \fi

\usepackage{graphicx}	
\usepackage{amsmath}	
\usepackage{amssymb}	
\usepackage{booktabs}
\usepackage{times}
\usepackage{microtype}
\usepackage{epsfig}
\usepackage[table,xcdraw,dvipsnames]{xcolor}
\usepackage{caption}
\usepackage{float}
\usepackage{placeins}
\usepackage{color}
\usepackage{colortbl}
\usepackage{stfloats}
\usepackage{enumitem}
\usepackage{tabularx}
\usepackage{xstring}
\usepackage{multirow}
\usepackage{xspace}
\usepackage{url}
\usepackage{subcaption}
\usepackage{xcolor}
\usepackage[hang,flushmargin]{footmisc}
\usepackage{duckuments}
\usepackage{graphbox}
\usepackage{diagbox}
\usepackage{lipsum}  
\usepackage{makecell}
\usepackage{afterpage}
\usepackage{contour}
\usepackage{contour}
\usepackage[normalem]{ulem}
\usepackage{colortbl}
\usepackage{chngcntr}
\usepackage{tablefootnote}
\usepackage{soul}
\contourlength{0.8pt}

\newcommand{\myuline}[1]{%
  \uline{\phantom{#1}}%
  \llap{\contour{white}{#1}}%
}

\ifcamera \usepackage[accsupp]{axessibility} \fi

\newcommand{\dalle}[1]{{DALL·E~3}}
\newcommand{\dallet}[1]{{DALL·E~2}}
\newcommand{\dalledot}[1]{·}
\newcommand{\sota}[1]{state-of-the-art}
\newcommand{\tti}[1]{text-to-image}
\newcommand{\pkmn}[1]{Pokémon}

\newcommand{\MethodName}{FakeInversion\xspace}

\newcommand{\bfparnodot}[1]{\vspace{.25em} \noindent \textbf{#1}}

\newcommand{\bfpar}[1]{\bfparnodot{#1.}}

\newcommand{\supp}{supplemental material\xspace}
\ifarxiv \renewcommand{\supp}{appendix\xspace} \fi

\newcommand{\citetodo}[1]{{\textcolor{red}{[?]}}}

\newcommand{\R}[1]{{%
    \textbf{%
        \ifstrequal{#1}{1}{\textcolor{red}{R#1}}{%
        \ifstrequal{#1}{2}{\textcolor{blue}{R#1}}{%
        \ifstrequal{#1}{3}{\textcolor{magenta}{R#1}}{%
        \ifstrequal{#1}{4}{\textcolor{teal}{R#1}}{%
                           \textcolor{cyan}{R#1}%
        }}}}%
    }%
}}

\DeclareMathOperator*{\argmin}{argmin}

\usepackage{xr-hyper}

\makeatletter
\newcommand*{\addFileDependency}[1]{
  \typeout{(#1)}
  \@addtofilelist{#1}
  \IfFileExists{#1}{}{\typeout{No file #1.}}
}

\makeatother
\newcommand*{\myexternaldocument}[1]{
    \externaldocument{#1}
    \addFileDependency{#1.tex}
    \addFileDependency{#1.aux}
}

\definecolor{cvprblue}{rgb}{0.21,0.49,0.74}
\definecolor{cvprblue}{rgb}{0.21,0.49,0.74}
\usepackage[pagebackref,breaklinks,
colorlinks,
citecolor=cvprblue,
linkcolor=red,
anchorcolor=red,
filecolor=red,
urlcolor=cvprblue
]{hyperref}
\usepackage[capitalize]{cleveref}
\crefname{section}{Sec.}{Secs.}
\crefname{table}{Table}{Tables}
\crefname{figure}{Figure}{Figures}
\ifarxiv \crefname{appendix}{App.}{Apps.}
\else \crefname{appendix}{Suppl.}{Suppls.} \fi
\unless\ifarxiv \myexternaldocument{_supplementary} \fi

\begin{document}
\title{\paperTitle}
\author{\authorBlock}

\newcommand{\AS}[1]{\textcolor{blue}{{\textbf{[Avneesh: #1]}}}}

\setlength{\abovedisplayskip}{3pt}
\setlength{\belowdisplayskip}{3pt}

\twocolumn[{
\maketitle
\vspace*{-0.25in}
\begin{tabular}{c}

    \centering
    \includegraphics[width=.97\linewidth]{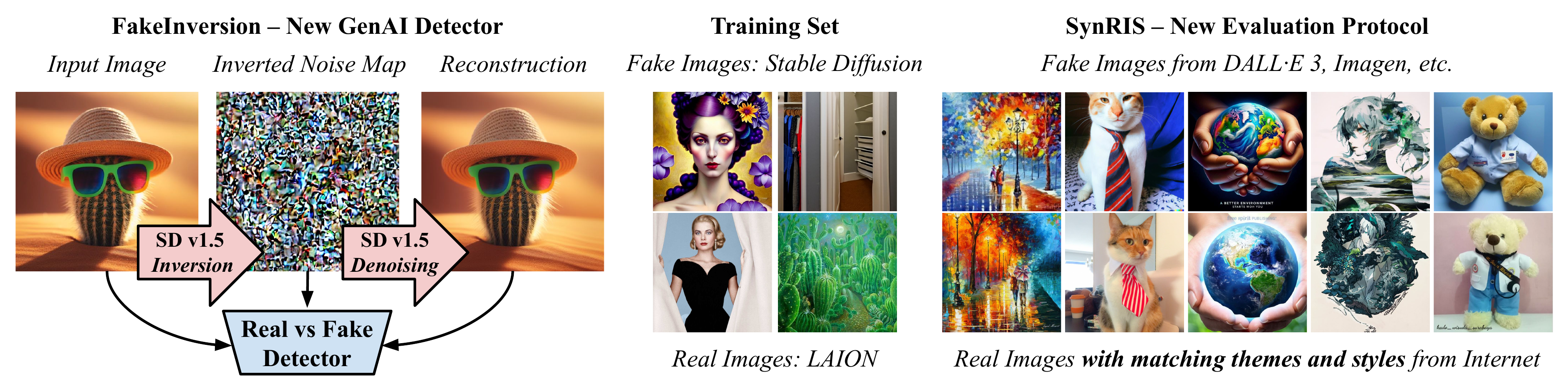}
    \end{tabular}

    \captionof{figure}{\textbf{(left)} We propose a  \protect\myuline{new synthetic image detector} that uses two additional input signals derived from a fixed pre-trained Stable Diffusion \cite{sd}: an inverted latent noise map and the reconstructed input image. 
    \textbf{(middle)} Our detector is trained using fake images generated using Stable Diffusion and real LAION images.
    It achieves \protect\myuline{state-of-the-art generalization performance} in detecting unseen text-to-image generators.
    \textbf{(right)} To ensure that the performance evaluation does not favor detectors that are biased towards particular themes or styles, we introduce a \protect\myuline{new thematically and stylistically aligned evaluation benchmark} 
    -- we measure detector's ability to discriminate fake images (\eg{}~\dalle{}, Imagen) from real images with matching content and style found on the Internet using reverse image search (RIS). 
    }
    \vspace{0.25in}
    \label{fig:cover}
}
]

\begin{abstract}

Due to the high potential for abuse of GenAI systems, the task of detecting synthetic images has recently become of great interest to the research community. 
Unfortunately, existing image-space detectors quickly become obsolete as new high-fidelity text-to-image models are developed at blinding speed. 
In this work, we propose a new synthetic image detector that uses features obtained by inverting an open-source pre-trained Stable Diffusion model. 
We show that these inversion features enable our detector to generalize well to unseen generators of high visual fidelity (e.g., \dalle{}) even when the detector is trained only on lower fidelity fake images generated via Stable Diffusion.
This detector achieves new state-of-the-art across multiple training and evaluation setups. 
Moreover, we introduce a new challenging evaluation protocol that uses reverse image search to mitigate stylistic and thematic biases in the detector evaluation. 
We show that the resulting evaluation scores align well with detectors' in-the-wild performance, and release these datasets as public benchmarks for future research. 
\end{abstract}

\saythanks
\section{Introduction}
\label{sec:intro}

\looseness=-1
Recent advances in text-to-image modeling have made it easier than ever to generate
harmful or misrepresentative content at scale.
Moreover, new versions of most photorealistic commercial models are being continuously updated and released behind closed APIs, making it harder to keep fake image detectors up to date.
In this work, we make significant strides towards building a GenAI detector that can reliably identify images from unseen photorealistic text-to-image models.
Specifically, we propose a model that can be trained using fake images \textit{only} from Stable Diffusion (SD)~\cite{sd} and 
reliably detect images generated by recent open (Kandinsky~\cite{kandinsky2_2}, W\"uerstchen~\cite{wurstchen}, PixArt-$\alpha$~\cite{chen2023pixart}, etc.) and closed-source text-to-image models (Imagen~\cite{imagen}, Midjourney~\cite{midjourney}, \dalle3{}~\cite{dalle3}, etc.) of \textit{significantly higher} visual fidelity.

\afterpage{
\begin{figure*}
\begin{center}
\centering
\includegraphics[width=\linewidth]{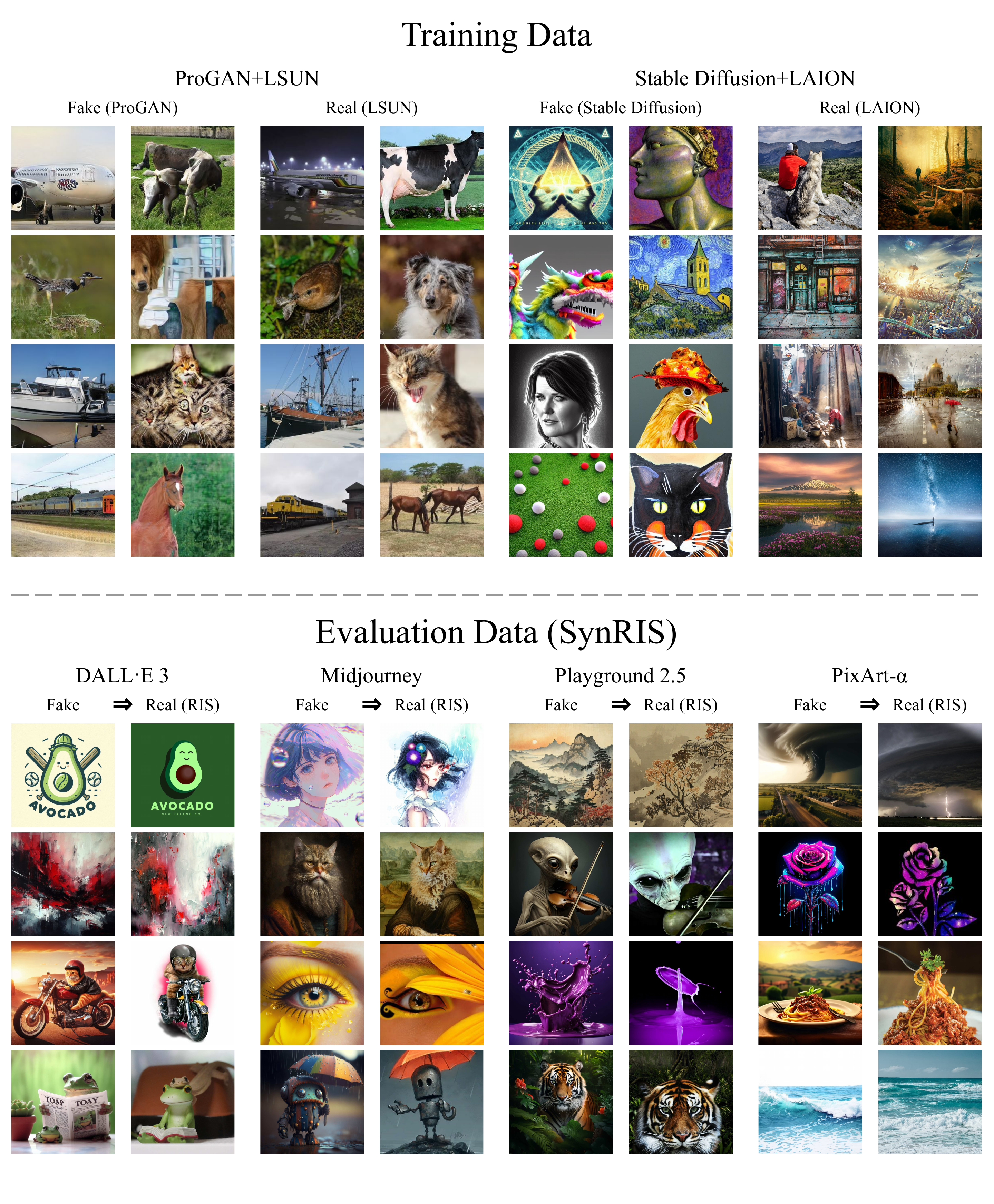}
\end{center}
\vspace{-2.5em}
\caption{
\looseness=-1
\textbf{Training and evaluation datasets.} We train all methods on \textbf{two training sets (top)}: ProGAN+LSUN from \cite{wang2019cnngenerated} and Stable Diffusion+LAION with fake images taken from DiffusionDB~\cite{wang2022diffusiondb}. We construct a new \textbf{evaluation benchmark SynRIS (bottom)} using reverse image search (RIS) on fake images generated by both proprietary (\eg{}, \dalle{}~\cite{dalle3}, Midjourney~\cite{midjourney}) and open-source (\eg{}, Playground~\cite{li2024playground}, PixArt-$\alpha$) models. Note that generators used for evaluation are of \textit{significantly higher visual fidelity} than those used for training. \vspace{10px}}
\label{fig:ris}
\end{figure*}
}

Existing methods~\cite{corvi2023naples,wang2019cnngenerated,ufd} focus primarily on detecting traces left by convolutional generators in a way that is robust to re-compression, resizing and other in-the-wild transformations. 
While these methods worked 
well for GANs and early diffusion models, we show that they, unfortunately, fail to generalize well to current photorealistic generative models, even when re-trained using better data.
Recent diffusion detectors that rely on CLIP embeddings~\cite{ufd} or inversions~\cite{wang2023dire} fail to generalize to challenging benchmarks. Drawing inspiration from recent works that showed that GANs tend to ``omit hard objects''~\cite{bau2019seeing} and that text-to-image models lean towards ``easily captionable'' images~\cite{sha2023defake}, in this paper, we focus on detecting fake images by analyzing internal representations of an existing off-the-shelf text-to-image model.

\looseness=-1
In this paper, we introduce a new \myuline{synthetic image} \myuline{detection method}: \textbf{\MethodName}. 
Our method uses features extracted from a lower-fidelity open-source text-to-image model (Stable Diffusion~\cite{sd}) to detect images generated by unseen text-to-image generators. 
Specifically, our model takes as input 1) the original image, 2) the approximate noise map recovered via text-conditioned DDIM~\cite{ddim} inversion with Stable Diffusion (SD), and 3) the reconstruction obtained by ``denoising'' the approximate noise map (Figure~\ref{fig:cover}).
We show that these additional signals significantly improve the performance of the resulting detector on unseen new proprietary 
and open-source 
photorealistic text-to-image models, attaining a new state-of-the-art.
We also provide an intuitive justification for why a diffusion detector needs such features to generalize well to unseen diffusion generators.

To deploy a synthetic image detector at scale, we need to make sure that it is not relying on content signals such as the presence of specific objects or styles in the image.
If left unmitigated, such bias towards particular themes or styles would 
disproportionately marginalize particular groups when applied to detecting healthcare misinfo~\cite{axios2023healthdeepfake} or forged art~\cite{nyt2022mjart} at scale.
Unfortunately, existing evaluation protocols that measure a detector's ability to differentiate between real and fake images drawn from very visually and thematically different distributions can not be used to test for the presence of such bias in the detector.
For example, evaluating a fake detector using real COCO~\cite{lin2014microsoft} images and fake images generated by \dalle{}~\cite{dalle3} could favour a detector that assigns higher fakeness score to digital art, since COCO contains mostly natural images.
To circumvent these issues, we propose a \myuline{new evaluation protocol}:  \textbf{SynRIS}.
For each set of synthetic images, we
evaluate a given detector against a set of real images obtained by applying reverse image search (Figure~\ref{fig:cover}) to given fake images -- the resulting evaluation does not favor models biased towards any particular topic, theme, or style.
We show that
the proposed evaluation protocol is more reliable at evaluating the quality of the synthetic image detector, especially when applied to closed-source text-to-image models. 
We will release our evaluation benchmark (including datasets) for future research. 

To summarize our contributions: 1) we introduce a new synthetic image detector that uses text-conditioned inversion maps extracted from Stable Diffusion; 2) we show that this additional feature improves the detector's ability to detect images generated by unseen text-to-image models, achieving new state-of-the-art; 3) we propose a new challenging evaluation protocol that uses reverse image search to ensure that the classifier is not biased towards any particular theme or style; 4) we verify that this evaluation protocol reliably measures detector generalization to closed text-to-image models; and 5) we release our challenging benchmark for future research.

\section{Related Work}
\label{sec:related}

In this section we first give a brief overview of the state-of-art in image-space detectors, then we discuss how recent works attempt to detect semantic inconsistencies in generated images, and finally discuss how our evaluation protocol compares to evaluation protocols used in prior work.

\bfpar{Artifact Detectors}~\citet{wang2019cnngenerated} were among the first to show that a CNN detector (CNNDet) generalized well from more powerful GANs (\eg ProGAN) to less powerful ones. Soon after,~\citet{patchforensics} extended this idea with a fully-convolutional network that classified individual image patches,~\citet{ju2022fusing} explored fusing global and local image features,~\citet{corvi2023naples} explored better augmentation and downsampling strategies,~\citet{zhang2019detecting} and~\citet{frank2020leveraging} explored artifacts in the spectrum of GAN-generated images, and~\citet{marra2019gans} explored GAN fingerprinting.

\bfpar{Generation Inconsistencies} 
Several works have focused on understanding the semantic properties of generated images. 
For example,~\citet{bau2019seeing} showed that GANs avoid generating ``hard objects'' such as mirrors and TVs -- that both humans and discriminators fail to notice missing. Recently, 
\citet{ufd} showed that image CLIP~\cite{radford2021clip} embeddings are highly predictive of whether an image is fake, and~\citet{sha2023defake} showed that images generated using text-to-image models tend to have higher CLIP similarity to their automatically inferred captions, suggesting that images generated by text-to-image models can often be described more fully by short text captions compared to images naturally occurring on the web. 
Inspired by these works, we also focus on properties of images beyond their low-level convolutional traces by examining internal representations of diffusion models obtained using DDIM inversion~\cite{ddim}.
A concurrent work~\cite{wang2023dire} found that DDIM image reconstruction residuals are predictive of whether an image is fake. In this paper we justify why image-space residuals are insufficient, and empirically verify that a detector that uses internal representations of a diffusion model generalizes better. 
We evaluate our model against the official DIRE checkpoint and perform an ablation using only reconstruction residuals.

\bfpar{Evaluation Protocols} 
\looseness=-1
Given that internal representations of diffusion models lack the low-level features necessary to perform generator trace detection, we need a way to ensure that the learned classifier does not overfit to particular objects or styles. 
Unfortunately, prior works focus either on open-source models with known training sets but lower visual fidelity or use dataset pairs of real and in-the-wild fake images that are too different both in style and content to ensure the lack of such bias. 
For example, recent works of~\citet{corvi2023naples} and~\citet{ufd} measure detectors' ability to discriminate between \dallet{} images and a mix of Imagenet, COCO and UCID~\cite{schaefer2003ucid} or LAION respectively, and DIRE~\cite{wang2023dire} focuses only on open-source lower fidelity generators such as SD and older generators trained on ImageNet and LSUN-Bedrooms~\cite{yu2015lsun}.
In a concurrent work,~\citet{epstein2023online} evaluate how adding training data from older models affects the performance of the classifier on newer fakes, which is an important problem but different from the one we address in this paper.

\looseness=-1
To summarize, we are the first to show that text-conditioned DDIM inversion feature maps extracted from one diffusion model improve the ability of a detector to identify images generated by other higher-fidelity diffusion models. Moreover, we are the first to propose an evaluation procedure for GenAI detectors that ensures that the learned detector is not biased towards any style or theme, and to quantitatively verify that the resulting evaluation is more reliable. 

\section{Method}

\label{sec:method}
In this section, we first provide a background on diffusion models and DDIM inversion. 
Then, we introduce our detection method that makes use of text-conditioned DDIM inversion and give an intuitive justification for why having this signal is helpful for generalization.

\bfparnodot{Latent Diffusion Models}. LDMs~\cite{sd} first map high-resolution (in our case, 512$\times$512$\times$3) RGB images $x$ into low-resolution (64$\times$64$\times$4) latent images $z$ using a pre-trained encoder $E: \mathcal X \to \mathcal Z$. The original image can be recovered almost perfectly via a pre-trained decoder $D: \mathcal Z \to \mathcal X$. 
In the derivation below $z_*$ correspond to such \textit{latent images}, rather than RGB images. 

\looseness=-1
\bfpar{Conditional Diffusion Models and DDIM Inversion} To generate a new latent image $z$ conditioned on some vector $c$, a conditional denoising diffusion model~\cite{ddpm} starts from a random noise map $z_T$ of the same shape and iteratively stochastically denoises it using a learned denoising network $\epsilon_\theta$ for a fixed number of steps, until a clean latent image $z_0$ is obtained.
The process of sampling from a pre-trained diffusion model can be discretized into fewer steps and made deterministic through the use of DDIM sampling~\cite{ddim}.
Notably, this sampling procedure enables ``inverting'' a clean image $z_0$ into a corresponding noise map $z_T$, such that when $z_T$ is denoised via DDIM sampling, we obtain a new latent $\hat{z}_0$ that is very close to the original $z_0$. Formally, to invert an image $z_0$, \ie, to obtain a corresponding noise map $z_T$, we iteratively add noise to its current estimate $z_t$ via the following \textit{conditional forward process} starting from a clean latent image $z_0$:
\begin{equation}
    z_{t+1} = \sqrt{\Bar{\alpha}_{t+1}}f_\theta(z_t, t, c) + \sqrt{1-\Bar{\alpha}_{t+1}}\epsilon_\theta(z_t,t,c)
\end{equation}
where $z_t$ is the noisy latent at time $t$, vector $c$ is the conditioner, value $\Bar{\alpha}$ is the DDIM noise scaling factor~\cite{ddim}, noise $\epsilon_\theta(z_t, t, c)$ is the prediction of the learned denoising function $\epsilon_\theta$ at time $t$, and $f_\theta(z_t, t, c)$ is the best current estimate of the clean latent $z_0$:
\begin{equation}
    f_\theta(z_t, t, c) = \frac{z_t - \sqrt{1-\Bar{\alpha}_t}\epsilon_\theta(z_t,t,c)}{\sqrt{\Bar{\alpha}_t}}
\end{equation}

\noindent A imperfect reconstruction $\hat{z}_0$ can then be obtained via the deterministic \textit{conditional reverse process} starting from $z_T$:
\begin{equation}
    \hat{z}_{t-1} = \sqrt{\Bar{\alpha}_{t-1}}f_\theta(\hat{z}_t, t, c) + \sqrt{1-\Bar{\alpha}_{t-1}}\epsilon_\theta(\hat{z}_t,t,c).
\end{equation}
We will refer to such full forward and reverse mapping as: 
\begin{equation}
    \hat{z}_{T} = F_{\theta}(z_0, c), \quad \hat{z}_{0} = R_{\theta}(\hat{z}_{T}, c).
\end{equation}

\bfpar{Text Conditioning} In our case, the conditioning vector $c$ used to modulate the forward and reverse sampling processes is the embedding of the text prompt describing an image. In this work, we use an off-the-shelf captioner (BLIP 2~\cite{blip}) to obtain a text prompt describing an input image, and CLIP~\cite{radford2021clip} to embed this text. Prior work showed that the realism of generated images can be improved through the use of classifier-free guidance~\cite{ho2022cfg}. 
Later,~\citet{mokady2023null} showed that classifier-free guidance leads to instability in DDIM inversion
and proposed a mitigation strategy through fine-tuning parts of the model. Since we 
cannot afford fine-tuning on each incoming image, in this paper we do not use classifier-free guidance during inversion and sampling and use the original conditional update rules described above.

\begin{figure*}
    \centering
    \hspace*{40px}\includegraphics[width=0.85\linewidth]{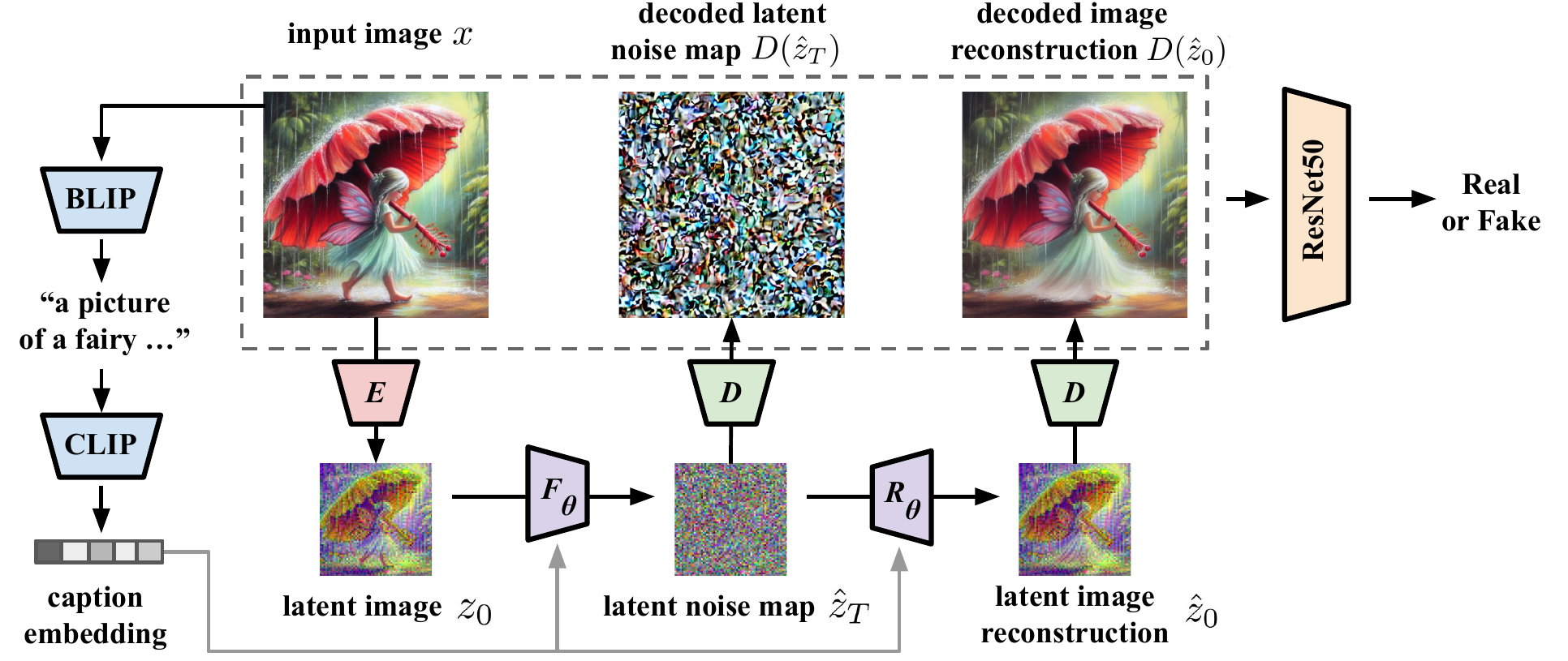}
    \caption{\textbf{Proposed method.} 
    In addition to the original image itself ($x$), we also train our detector using the (decoded) noise map $D(\hat{z}_T)$ and reconstruction $D(\hat{z}_0)$ obtained by inverting the image through Stable Diffusion using DDIM.
    The original image is first mapped to the latent space with Stable Diffusion's {\textcolor[HTML]{cc0000}{\textbf{VAE Encoder}}}. The latent image is then inverted and reconstructed through Stable Diffusion's {\textcolor[HTML]{674ea7}{\textbf{U-Net}}} using DDIM while conditioned on the {\textcolor[HTML]{3d85c6}{\textbf{CLIP}}} embedding of the image's predicted {\textcolor[HTML]{3d85c6}{\textbf{BLIP}}} caption. The latent noise map and reconstruction are mapped back to image space using Stable Diffusion's {\textcolor[HTML]{6aa84f}{\textbf{VAE Decoder}}}. The original image, decoded noise map, and decoded reconstruction are then concatenated and used as input for our {\textcolor[HTML]{e69138}{\textbf{ResNet Classifier}}}.
    }
    \vspace{-12px}
    \label{fig:method_diag}
\end{figure*}

\bfpar{GenAI Detector} As shown in Figure~\ref{fig:method_diag}, given an input image $x$, we first caption that image using BLIP~\cite{blip} and embed that caption into a vector $c$ using CLIP~\cite{radford2021clip}. Then we compute the corresponding latent image $z_0 = E(x)$ using a pre-trained encoder and obtain a latent DDIM noise map $\hat{z}_T$ using text-conditioned DDIM inversion with a pre-trained diffusion model. Then, we obtain a reconstructed latent image $\hat{z}_0$ using text-conditioned DDIM sampling and decode both the latent noise map $\hat{z}_T$ and the reconstruction $\hat{z}_0$ to image space using the decoder $D$. Finally, we apply a learned mapping $h_{\phi}: \mathcal{X}^3 \to \mathbb R$ to these ``images'' to get a prediction logit. We learn the parameters of this function $\phi$ via backpropagation of the binary cross-entropy loss $\ell(\hat y, y)$ on the training set of known fake and real images $\{(x, y)\}$:
\begin{gather}
    c = \operatorname{CLIP}(\operatorname{BLIP}(x)) \\
    \hat{z}_{T} = F_{\theta}(E(x), c), \quad \hat{z}_{0} = R_{\theta}(\hat{z}_{T}, c) \\
    \phi^* = \argmin_{\phi} \mathbb E_{x,y}  [\, \ell\left(h_{\phi}\left(x, D(\hat{z}_T), D(\hat{z}_0)\right), y\right)]
\end{gather}

\bfpar{Intuition}
But why would a diffusion detector benefit from having access to DDIM inversion of an image if it already has access to the image itself? 
Recent works~\cite{lu2022dpm,salimans2022progressive} showed that DDIM can be viewed as a first-order discretization of a neural probability-flow ODE. 
The bijection between observations $z_0$ and noise maps $z_t$ induced by this ODE can be used to evaluate the likelihood of the data via the change of variable. 
If we view the forward DDIM mapping $F_\theta$ %
as an approximation of that true bijective mapping between the $z_0$ and $z_T$ that introduces a discretization error $\delta$ into the inverted noise maps $\hat{z}_T$, causing the resampled image $\hat{z}_0$ to deviate from the original image $z_0$, it can be shown (see \cref{sup:log_prob_derivation}) that, in the first-order approximation, the log-likelihood of the data given that underlying model can be estimated from the input image $z_0$, its imperfect reconstruction $\hat{z}_0$, and the noise map $z_T$ alone:
\begin{equation}
    \label{eq:log_prob}
    \log p(z_0) \propto \log p_z(z_T) - \langle \delta, \hat{z}_0  - z_0 \rangle / \|\delta\|^2
\end{equation}
\looseness=-1
Notably, the expression above does not explicitly depend on the parameters $\theta$ other than through these three signals. 
Given that, under model misspecification, likelihood-based models tend to overgeneralize~\cite{huszar2015overgeneralization}, producing samples that are unlikely under the true data distributions, and assuming that the class of diffusion-based models is similar enough to overgeneralize in similar ways, we propose that a model that has access to all signals required to internally perform some form of likelihood testing on input data against a particular text-to-image model (the image, its imperfect reconstruction, and the intermediate noise map) would generalize better to detect images generated via other diffusion models. Text conditioning $c$ further amplifies differences between log-likelihoods of distributions of fake and real images, making the corresponding test more powerful and consequently making inversions even more useful for detecting fake images. 
To sum up, GenAI detectors find discrepancies between the real data distribution and the approximation learned by the GenAI model. 
The equation above shows that proposed features enable a detector to get a rough estimate of whether a given image is of high probability under the approximate distribution learned by Stable Diffusion. 
We empirically verify that this ``SD-likelihood'' signal helps in detecting other generators.

\section{Experiments}

In this section, we discuss our training and evaluation sets, which baseline methods we use to compare, and the details of how we train our classifier. 

\bfpar{Baselines}
We compare our model to a representative set of the most recent state-of-art baselines (all published in \myuline{2023}) that open-sourced their training or inference code. \myuline{DMDet}~\cite{corvi2023naples} is a state-of-art RGB-only method that achieved significant generalization performance through the use of augmentations and a modified down-sampling strategy; authors released only the inference checkpoint. \myuline{UFD}~\cite{ufd} is another recent state-of-the-art method that trains a linear classification head on top of the CLIP~\cite{radford2021clip} embeddings of real and fake images. We use the official checkpoint and also retrain it from scratch on each of our training sets using the official code. \myuline{DIRE}~\cite{wang2023dire} is a concurrent work that showed that using image-space DDIM reconstruction residuals helps detection. The official checkpoint open-sourced by the authors has an issue causing its performance to be much lower than the performance reported in the paper; we discuss this in more detail in \cref{sup:dire}. We also provide an ablation of our method that uses only DDIM residuals. 
This serves as a close approximation of what a DIRE-like method \textit{could} achieve. 
We also include an older convolutional baseline \myuline{CNNDet}~\cite{wang2019cnngenerated} using its official checkpoint and code to retrain on our data.

\begin{table}[t]
\centering
\setlength{\tabcolsep}{1pt}
\begin{tabular}{cccc}

\toprule
\textbf{Dataset} & \textbf{Data Size} & \textbf{Real Data} & \textbf{Real/Fake Source} \\
\midrule
\footnotesize{\dallet{}} \cite{ramesh2022hierarchical} & 700/700 & RIS (ours) & fakes from~\cite{corvi2023naples} \\
\arrayrulecolor{black!10}\midrule
\footnotesize{\dalle{}}   \cite{dalle3}& 3.3k/3.3k & RIS (ours) & fakes from~\cite{dalle3hf} \\
\arrayrulecolor{black!10}\midrule
\footnotesize{Midjourney} \cite{midjourney} & 4.4k/4.4k & RIS (ours) & \makecell{fakes from~\cite{wanng_midjourney_v5}} \\
\arrayrulecolor{black!10}\midrule
\footnotesize{Imagen \cite{imagen}} & 700/700 & RIS (ours) & \makecell{our fakes \\ (see \cref{sup:data})} \\
\arrayrulecolor{black!10}\midrule
\footnotesize{Open-Source\hyperref[foot]{$^{\color{red}1}$}} & \makecell{3.5k/3.5k\\(x11)} & RIS (ours) & \makecell{our fakes \\ (see \cref{sup:data})} \\ 
\arrayrulecolor{black!100}\midrule
\footnotesize{\dallet{} (A)} & 1k/5k & \makecell{Imagenet,\\\footnotesize COCO, UCID \cite{schaefer2003ucid}} & \makecell{both reals/fakes \\ from~\cite{corvi2023naples}} \\
\arrayrulecolor{black!10}\midrule
\footnotesize{Craiyon  (A)} & 1k/1k & LAION & both from~\cite{ufd} \\
\arrayrulecolor{black!10}\midrule
\footnotesize{LDM  (A)} & 1k/1k & LAION & both from~\cite{ufd} \\
\arrayrulecolor{black!100}\bottomrule
\end{tabular}
\caption{\textbf{Evaluation Datasets.}
We evaluate using 15 new RIS-based evaluation benchmarks (\textbf{SynRIS (ours)} -- top) as well as existing academic (A) text-to-image evaluation dataset (bottom).
\label{tab:datasets_v2}
}
\vspace{-1em}
\end{table}

\begin{table}[]
\centering

\renewcommand{\arraystretch}{0.95}
\setlength{\tabcolsep}{1pt}
\newlength{\characterlength}
\settowidth{\characterlength}{0}
\begin{tabular}{cccccc}
\toprule
\multicolumn{2}{c}{Eval Data}
& \multicolumn{2}{c}{Training Data} & \multirow{2}{*}{FID} & \multirow{2}{*}{\makecell{KID \\ $\times 10^{-2}$}} \\
\cmidrule(lr){1-2} \cmidrule(lr){3-4}
Fake & Real & {\scriptsize{ProGAN+LSUN}} & {\scriptsize{SD+LAION}} \\ 
\midrule
\multirow{2}{*}{\footnotesize{\dallet{} %
}} & {\footnotesize LAION} & 0.233 & 0.043 & 163.5 & 2.7\\
 & \footnotesize{\textbf{RIS (ours)}} & \textbf{0.457} & \textbf{0.406} & \hspace{\characterlength}\textbf{88.5} & \textbf{0.3} \\
\arrayrulecolor{black!10}\midrule
\multirow{2}{*}{\footnotesize{\dalle{} 
}} & {\footnotesize LAION} & 0.794 & 0.360 & \hspace{\characterlength}126.1 & 2.6 \\
 & \footnotesize{\textbf{RIS (ours)}} & \textbf{0.920} & \textbf{0.795} & \hspace{\characterlength}\textbf{93.6} & \textbf{0.4} \\
\arrayrulecolor{black!10}\midrule
\multirow{3}{*}{Imagen
} & {\footnotesize LAION} & 0.406 & 0.360 & \hspace{\characterlength}127.1 & 2.1 \\
 & WebLI & 0.559 & 0.664 & \hspace{\characterlength}101.9 & 1.2 \\
 & \footnotesize{\textbf{RIS (ours)}} & \textbf{0.620} & \textbf{0.720} & \hspace{\characterlength}\textbf{83.6} & \textbf{0.2} \\
\arrayrulecolor{black!10}\midrule
\multirow{2}{*}{\footnotesize{Kandinsky 2 
}} & {\footnotesize LAION} & 0.655	& 0.189 & \hspace{\characterlength}118.0 & 2.1 \\
 & \footnotesize{\textbf{RIS (ours)}} & \textbf{0.857} & \textbf{0.686} & \hspace{\characterlength}\textbf{88.9} & \textbf{0.4} \\
\arrayrulecolor{black!10}\midrule
\multirow{2}{*}{\footnotesize{\footnotesize{SDXL 
}}} & {\footnotesize LAION} & 0.689 & 	0.106 & \hspace{\characterlength}108.7 & 1.6 \\
 & \footnotesize{\textbf{RIS (ours)}} & \textbf{0.874} & \textbf{0.551} & \hspace{\characterlength}\textbf{93.0} & \textbf{0.4} \\
\arrayrulecolor{black!100}\bottomrule
\end{tabular}
\caption{\textbf{Difficulty of RIS vs LAION eval.} FPR@0.8 recall for the state-of-the-art detector (UFD \cite{ufd}) evaluated using fakes from respective generators and real images from LAION, reverse image search (RIS), and WebLI (Imagen's training set \cite{yu2022parti}) - higher FPR is harder; FID and KID between reals and fakes - lower is closer.
\label{tab:ris_eval_is_harder}
}
\vspace{-1em}
\end{table}

\bfpar{Training data -- ProGAN+LSUN}
\looseness=-1
Most prior works use the ProGAN training set introduced in CNNDet~\cite{wang2019cnngenerated}. 
This training set consists of 350k images from class-conditioned pre-trained ProGAN~\cite{karras2019progan} combined with a set of real images from LSUN~\cite{yu2015lsun}.
Training on these images has yielded good results in the detection of GAN-generated images \cite{wang2019cnngenerated,corvi2023naples,ufd}, and we find that this set continues to show promise when applied to images from newer diffusion models.

\bfpar{Training data -- Stable Diffusion+LAION} Similar to the concurrent work of~\citet{epstein2023online}, we first train detectors using 300k fake Stable Diffusion v1 images from DiffusionDB~\cite{wang2022diffusiondb} and 300k random real LAION~\cite{schuhmann2022laion} images. 
While state-of-the-art at the time of its release, Stable Diffusion (v1) has since been eclipsed in quality by many new text-to-image models. 
We find that training on fake images from this relatively old diffusion model still yields models capable of identifying fakes from much newer and more powerful generators.

\bfpar{Evaluation data (fakes)} 
We obtain several thousand images generated by closed-source photorealistic text-to-image models using APIs (Imagen~\cite{imagen}), using existing databases of fakes on HuggingFace (Midjourney~\cite{wanng_midjourney_v5}, \dalle{}~\cite{dalle3hf}) and by taking fake images from prior academic benchmarks (\dallet{} from~\cite{corvi2023naples}). 
We also generate several thousand images using high-fidelity open-source text-to-image \nobreakdash models\footnote{\label{foot}Open-Source dataset includes fake images from Kandinsky~2~\cite{kandinsky2_2}, Kandinsky~3~\cite{kandinsky3}, PixArt\nobreakdash-$\alpha$~\cite{chen2023pixart}, SDXL\nobreakdash-DPO~\cite{dpo}, SDXL~\cite{sdxl}, \mbox{SegMoE}~\cite{segmoe}, SSD\nobreakdash-1B~\cite{segmind}, Stable\nobreakdash-Cascade~\cite{wurstchen}, Segmind\nobreakdash-Vega~\cite{segmind}, and W\"urstchen~2~\cite{wurstchen}.} conditioned on Midjourney prompts~\cite{mj_prompts}. 

\bfpar{Evaluation data (reals) -- Reverse Image Search (RIS)} 
To ensure that detectors are not biased toward any particular theme or style, we need sets of real and fake images that are themselves stylistically and thematically aligned.
We address this issue using a reverse image search API to find a visually and thematically similar real image for each fake image from the eval fake set defined above. 
Examples of images found using this procedure can be found in Figure~\ref{fig:ris}. 
We define real images as images not generated using a text-to-image model, even if other tools (such as Photoshop) were used. 
To ensure that our real images are not contaminated with similar images generated by text-to-image models, we include only matches found on pages created before January 1, 2021. 
As a result, our real sets include only images published before the original DALL\dalledot{}E~\cite{ramesh2021dalle} was announced. 
The exact sizes of all evaluation and training sets can be found in Table~\ref{tab:datasets_v2}.

\bfpar{Evaluation data -- prior academic benchmarks}
We evaluate competing methods on academic text-to-image benchmarks from published prior work~\cite{ufd,corvi2023naples} that evaluate methods' abilities to differentiate between a set of fakes from a text-to-image model (\eg, \dallet{}) and an unrelated set of real images (\eg, Imagenet, COCO). 
Consequently, these benchmarks can not be used to test whether a detector focuses on the styles and themes of a particular generator.

\bfpar{Detector architecture} We use ResNet50 trained from scratch as a detector backbone. In each experiment, we select the best checkpoint via validation on the held-out set sampled from the same source as the training set. We augment each image via a suite of random transforms \textit{before} performing DDIM inversion: flip, crop, color jitter, grayscale, cutout, noise, blur, jpeg, and rotate. We use BLIP~\cite{blip} to compute image captions. See~\cref{sup:training_details} for more details.

\bfpar{Metrics} \looseness=-1 We report detection AUCROC as the main metric.
We also provide tables with average precision and accuracy, along with PR, ROC and DET curves in the \supp{}. To ensure that trained and evaluated detectors can not exploit differences in image resolutions and aspect ratios, each image is resized to 256px along the shortest side and saved losslessly.

\section{Results}

\begin{table*}[!htp]\centering

\begin{tabular}{l!{\color{gray!20}\vrule}ccccc!{\color{gray!20}\vrule}cccc}\toprule
\multicolumn{1}{r}{Train Data}&\multicolumn{5}{c}{ProGAN + LSUN} &\multicolumn{4}{c}{Stable Diffusion + LAION} \\\cmidrule(lr){1-1}\cmidrule(lr){2-6} \cmidrule(lr){7-10}
\multicolumn{1}{c}{Eval Set\hspace{1em}\textbar{}\hspace{1em}Model} &DIRE &CNNDet &DMDet &UFD &Ours &CNNDet\textsuperscript{\textdagger} &DMDet$^*$ &UFD\textsuperscript{\textdagger} &Ours \\ \midrule
DALL·E 2 \cite{ramesh2022hierarchical} &0.561 &0.455 &0.656 &\cellcolor[HTML]{ffffff}0.728 &\cellcolor[HTML]{D2E3FC}\textbf{0.854} &0.680 &0.672 &\cellcolor[HTML]{D2E3FC}\textbf{0.776} &\cellcolor[HTML]{ffffff}0.747 \\
DALL·E 3 \cite{dalle3} &\cellcolor[HTML]{ffffff}0.524 &0.378 &0.409 &0.323 &\cellcolor[HTML]{D2E3FC}\textbf{0.642} &\cellcolor[HTML]{ffffff}0.716 &0.415 &0.480 &\cellcolor[HTML]{D2E3FC}\textbf{0.759} \\
Midjourney v5/6 \cite{midjourney} &0.538 &0.473 &\cellcolor[HTML]{ffffff}0.544 &0.397 &\cellcolor[HTML]{D2E3FC}\textbf{0.750} &\cellcolor[HTML]{ffffff}0.630 &0.484 &0.592 &\cellcolor[HTML]{D2E3FC}\textbf{0.664} \\
Imagen \cite{imagen} &0.562 &0.452 &0.502 &\cellcolor[HTML]{ffffff}0.637 &\cellcolor[HTML]{D2E3FC}\textbf{0.776} &\cellcolor[HTML]{ffffff}0.714 &0.573 &0.575 &\cellcolor[HTML]{D2E3FC}\textbf{0.807} \\ \midrule
Kandinsky 2 \cite{kandinsky2_2} &0.463 &\cellcolor[HTML]{ffffff}0.492 &0.468 &0.474 &\cellcolor[HTML]{D2E3FC}\textbf{0.758} &\cellcolor[HTML]{ffffff}0.600 &0.478 &0.562 &\cellcolor[HTML]{D2E3FC}\textbf{0.699} \\
Kandinsky 3 \cite{kandinsky3} &0.491 &0.480 &\cellcolor[HTML]{ffffff}0.593 &0.469 &\cellcolor[HTML]{D2E3FC}\textbf{0.845} &\cellcolor[HTML]{ffffff}0.659 &0.614 &0.637 &\cellcolor[HTML]{D2E3FC}\textbf{0.743} \\
PixArt-$\alpha$ \cite{chen2023pixart} &0.478 &0.487 &\cellcolor[HTML]{ffffff}0.599 &0.506 &\cellcolor[HTML]{D2E3FC}\textbf{0.854} &0.627 &0.580 &\cellcolor[HTML]{ffffff}0.647 &\cellcolor[HTML]{D2E3FC}\textbf{0.730} \\
Playground 2.5 \cite{li2024playground} &0.453 &0.528 &\cellcolor[HTML]{ffffff}0.661 &0.466 &\cellcolor[HTML]{D2E3FC}\textbf{0.778} &0.582 &0.517 &\cellcolor[HTML]{ffffff}0.587 &\cellcolor[HTML]{D2E3FC}\textbf{0.625} \\
SDXL-DPO \cite{dpo} &0.458 &0.486 &\cellcolor[HTML]{ffffff}0.603 &0.464 &\cellcolor[HTML]{D2E3FC}\textbf{0.841} &\cellcolor[HTML]{ffffff}0.843 &0.563 &0.702 &\cellcolor[HTML]{D2E3FC}\textbf{0.881} \\
SDXL \cite{sdxl} &0.459 &0.525 &\cellcolor[HTML]{ffffff}0.667 &0.464 &\cellcolor[HTML]{D2E3FC}\textbf{0.764} &\cellcolor[HTML]{D2E3FC}\textbf{0.814} &0.568 &0.663 &\cellcolor[HTML]{ffffff}0.807 \\
Seg-MOE \cite{segmoe} &0.459 &0.429 &\cellcolor[HTML]{ffffff}0.467 &0.401 &\cellcolor[HTML]{D2E3FC}\textbf{0.796} &\cellcolor[HTML]{ffffff}0.663 &0.476 &0.620 &\cellcolor[HTML]{D2E3FC}\textbf{0.713} \\
SSD-1B \cite{segmind} &0.449 &0.589 &\cellcolor[HTML]{ffffff}0.689 &0.515 &\cellcolor[HTML]{D2E3FC}\textbf{0.827} &\cellcolor[HTML]{ffffff}0.726 &0.556 &0.628 &\cellcolor[HTML]{D2E3FC}\textbf{0.794} \\
Stable-Cascade \cite{wurstchen} &0.465 &0.447 &\cellcolor[HTML]{ffffff}0.603 &0.341 &\cellcolor[HTML]{D2E3FC}\textbf{0.882} &\cellcolor[HTML]{ffffff}0.705 &0.565 &0.682 &\cellcolor[HTML]{D2E3FC}\textbf{0.749} \\
Segmind Vega \cite{segmind} &0.471 &0.556 &\cellcolor[HTML]{ffffff}0.645 &0.468 &\cellcolor[HTML]{D2E3FC}\textbf{0.823} &\cellcolor[HTML]{ffffff}0.742 &0.540 &0.623 &\cellcolor[HTML]{D2E3FC}\textbf{0.811} \\
W\"urstchen 2 \cite{wurstchen} &0.456 &0.510 &\cellcolor[HTML]{ffffff}0.671 &0.616 &\cellcolor[HTML]{D2E3FC}\textbf{0.792} &0.610 &0.675 &\cellcolor[HTML]{ffffff}0.697 &\cellcolor[HTML]{D2E3FC}\textbf{0.705} \\ \midrule
DALL·E 2 \cite{ramesh2022hierarchical} (A) &0.554 &0.466 &\cellcolor[HTML]{ffffff}0.646 &\cellcolor[HTML]{D2E3FC}\textbf{0.662} &0.623 &0.566 &\cellcolor[HTML]{D2E3FC}\textbf{0.727} &\cellcolor[HTML]{ffffff}0.590 &0.571 \\
Craiyon \cite{Dayma_DALL·E_Mini_2021} (A) &0.523 &0.660 &\cellcolor[HTML]{ffffff}0.941 &\cellcolor[HTML]{D2E3FC}\textbf{0.974} &0.874 &0.763 &\cellcolor[HTML]{D2E3FC}\textbf{0.988} &\cellcolor[HTML]{ffffff}0.918 &0.886 \\
LDM \cite{sd} (A) &0.512 &0.653 &0.854 &\cellcolor[HTML]{D2E3FC}\textbf{0.924} &\cellcolor[HTML]{ffffff}0.878 &0.913 &\cellcolor[HTML]{D2E3FC}\textbf{1.000} &0.919 &\cellcolor[HTML]{ffffff}0.979 \\ \midrule
Average &0.493 &0.504 &\cellcolor[HTML]{ffffff}0.623 &0.546 &\cellcolor[HTML]{D2E3FC}\textbf{0.798} &\cellcolor[HTML]{ffffff}0.697 &0.611 &0.661 &\cellcolor[HTML]{D2E3FC}\textbf{0.759} \\
\bottomrule
\end{tabular}
\caption{\textbf{Main Results -- Detector AUCROC.} Detectors trained on ProGAN+LSUN and SD+LAION are evaluated using proprietary (first panel) and open-source (second panel) generators, and academic (A) benchmarks from prior work (last panel). 
$^*$Note: This DMDet classifier was trained with fakes from an LDM checkpoint rather than Stable Diffusion. \textsuperscript{\textdagger}These models were re-trained by us.
\label{tab:main_v2} 
}
\vspace{-0.0em}
\end{table*}

In this section we discuss following key findings: 1) the proposed thematically and stylistically aligned RIS-based evaluation protocol is harder and is more reliable then protocols used in prior work; 2) the proposed detector outperforms prior work on both prior academic and this new RIS-based evaluation;
3) the DDIM inversion features were crucial in achieving high generalization in all cases.

\bfpar{RIS-based evaluation is harder and more reliable} Table~\ref{tab:ris_eval_is_harder} 
compares False Positive Rate (FPR) of the state-of-the-art detector~\cite{ufd} on different evaluation sets at the threshold that attains 80\% recall (fake images are the same, so the threshold at given recall is the same as well). 
Results show that LAION-based evaluation significantly underestimates the false positive rate of the detector when evaluating its ability to discriminate fakes from closed-source text-to-image models (Imagen, \dallet{}/3) across both training sets. 
We also obtained real examples from the multimodal dataset used to train Imagen (WebLI~\cite{yu2022parti}), and evaluated the detector against these real examples and these results closely align with our RIS-based eval (see Fig.~\ref{fig:ris-curve} for PR curves). 
The FID~\cite{fid} and KID \cite{kid} between real and fake images is also lower for RIS eval, and matches FID/KID between WebLI and Imagen fakes, suggesting better stylistic and thematic alignment.
Similar trends can be seen on open-source models (Kadnisky, SDXL) and across both training sets. 
These results suggest that our RIS-based eval is a more reliable way to estimate a model's ability to detect images from closed-sourced text-to-image models trained on unknown data.

\bfpar{\MethodName~achieves state-of-the-art performance} \mbox{Table}~\ref{tab:main_v2} shows that our method consistently scores best at detecting both closed and open-source methods across various training sets. It also matches the performance of prior work on academic benchmarks. 
On average, our method outperforms prior work by \textbf{at least 4pp} on \myuline{both training sets}. 

\begin{table}[t]
\centering
\small
\setlength{\tabcolsep}{3pt}
\vspace*{-7px}
\begin{tabular}{lcccccc}\toprule
\multicolumn{1}{r}{Train Data}&\multicolumn{3}{c}{ProGAN + LSUN} &\multicolumn{3}{c}{SD + LAION} \\\cmidrule(lr){1-1} \cmidrule(lr){2-4} \cmidrule(lr){5-7}
\multicolumn{1}{c}{Eval Set\hspace{1em}\textbar{}\hspace{1em}Model}&RGB &Res &Ours &RGB &Res &Ours \\ \midrule
DALL·E 2 \cite{ramesh2021dalle} &0.410 &0.650 &\cellcolor[HTML]{D2E3FC}\textbf{0.854} &0.592 &0.650 &\cellcolor[HTML]{D2E3FC}\textbf{0.747} \\
DALL·E 3 \cite{dalle3} &0.399 &\cellcolor[HTML]{D2E3FC}\textbf{0.672} &0.642 &0.676 &0.672 &\cellcolor[HTML]{D2E3FC}\textbf{0.759} \\
Midjourney v5 \cite{midjourney} &0.434 &0.590 &\cellcolor[HTML]{D2E3FC}\textbf{0.750} &0.530 &0.590 &\cellcolor[HTML]{D2E3FC}\textbf{0.664} \\
Imagen \cite{imagen} &0.530 &0.670 &\cellcolor[HTML]{D2E3FC}\textbf{0.776} &0.729 &0.670 &\cellcolor[HTML]{D2E3FC}\textbf{0.807} \\ \midrule
Kandinsky 2 \cite{kandinsky2_2} &0.462 &0.600 &\cellcolor[HTML]{D2E3FC}\textbf{0.716} &0.614 &0.607 &\cellcolor[HTML]{D2E3FC}\textbf{0.714} \\
Kandinsky 3 \cite{kandinsky3}&0.434 &0.617 &\cellcolor[HTML]{D2E3FC}\textbf{0.824} &0.606 &0.679 &\cellcolor[HTML]{D2E3FC}\textbf{0.774} \\
PixArt-$\alpha$ \cite{chen2023pixart}&0.470 &0.604 &\cellcolor[HTML]{D2E3FC}\textbf{0.647} &0.594 &0.570 &\cellcolor[HTML]{D2E3FC}\textbf{0.707} \\
Playground 2.5 \cite{li2024playground}&0.439 &0.604 &\cellcolor[HTML]{D2E3FC}\textbf{0.726} &0.510 &0.533 &\cellcolor[HTML]{D2E3FC}\textbf{0.660} \\
SDXL-DPO \cite{dpo}&0.338 &0.643 &\cellcolor[HTML]{D2E3FC}\textbf{0.704} &0.738 &0.711 &\cellcolor[HTML]{D2E3FC}\textbf{0.837} \\
SDXL \cite{sdxl}&0.410 &0.612 &\cellcolor[HTML]{D2E3FC}\textbf{0.691} &0.784 &0.709 &\cellcolor[HTML]{D2E3FC}\textbf{0.884} \\
Seg-MOE \cite{segmoe}&0.416 &0.585 &\cellcolor[HTML]{D2E3FC}\textbf{0.799} &0.607 &0.611 &\cellcolor[HTML]{D2E3FC}\textbf{0.781} \\
SSD-1B \cite{segmind}&0.494 &0.672 &\cellcolor[HTML]{D2E3FC}\textbf{0.775} &0.690 &0.648 &\cellcolor[HTML]{D2E3FC}\textbf{0.813} \\
Stable-Cascade \cite{wurstchen}&0.448 &0.674 &\cellcolor[HTML]{D2E3FC}\textbf{0.743} &0.557 &0.686 &\cellcolor[HTML]{D2E3FC}\textbf{0.766} \\
Segmind Vega \cite{segmind}&0.465 &0.677 &\cellcolor[HTML]{D2E3FC}\textbf{0.781} &0.683 &0.631 &\cellcolor[HTML]{D2E3FC}\textbf{0.829} \\
W\"urstchen 2 \cite{wurstchen}&0.563 &0.624 &\cellcolor[HTML]{D2E3FC}\textbf{0.664} &0.588 &0.605 &\cellcolor[HTML]{D2E3FC}\textbf{0.702} \\
\bottomrule
\end{tabular}
\caption{\looseness=-1\textbf{Input Signal Ablation -- AUCROC}. Detectors trained on ProGAN+LSUN and Stable Diffusion+LAION and evaluated on proprietary and open generators. Using the original images, inversion maps, and reconstructions (\textbf{Ours}) yields better performance than RGB or DDIM Residuals (as in DIRE \cite{wang2023dire}) alone. \label{tab:rgb_res_abl}}

\end{table}

\begin{table}[!htp]\centering
\small
\setlength{\tabcolsep}{3pt}
\begin{tabular}{lrcccccc}\toprule
& &\multicolumn{3}{c}{ProGAN + LSUN} &\multicolumn{3}{c}{SD + LAION} \\\cmidrule(lr){3-5} \cmidrule(lr){6-8}
& &\scriptsize{CNNDet} &UFD &Ours &\scriptsize{CNNDet} &UFD &Ours \\\midrule
\multirow{3}{*}{\rotatebox[origin=c]{90}{noise}} &Imagen &0.477 &0.579 &\cellcolor[HTML]{cfe2f3}\textbf{0.758} &0.730 &0.529 &\cellcolor[HTML]{cfe2f3}\textbf{0.822} \\
&MJ &0.481 &0.383 &\cellcolor[HTML]{cfe2f3}\textbf{0.665} &0.600 &0.533 &\cellcolor[HTML]{cfe2f3}\textbf{0.624} \\
&\scriptsize{\dalle{}} &0.390 &0.315 &\cellcolor[HTML]{cfe2f3}\textbf{0.598} &0.700 &0.449 &\cellcolor[HTML]{cfe2f3}\textbf{0.750} \\\arrayrulecolor{black!30}\specialrule{.2pt}{0.8pt}{1.5pt}
\multirow{3}{*}{\rotatebox[origin=c]{90}{blur}} &Imagen &0.447 &0.595 &\cellcolor[HTML]{cfe2f3}\textbf{0.793} &0.730 &0.570 &\cellcolor[HTML]{cfe2f3}\textbf{0.812} \\
&MJ &0.463 &0.379 &\cellcolor[HTML]{cfe2f3}\textbf{0.747} &0.639 &0.583 &\cellcolor[HTML]{cfe2f3}\textbf{0.658} \\
&\scriptsize{\dalle{}} &0.375 &0.315 &\cellcolor[HTML]{cfe2f3}\textbf{0.639} &0.738 &0.501 &\cellcolor[HTML]{cfe2f3}\textbf{0.756} \\\arrayrulecolor{black!30}\specialrule{.2pt}{0.8pt}{1.5pt}
\multirow{3}{*}{\rotatebox[origin=c]{90}{JPEG}} &Imagen &0.463 &0.651 &\cellcolor[HTML]{cfe2f3}\textbf{0.769} &0.715 &0.555 &\cellcolor[HTML]{cfe2f3}\textbf{0.804} \\
&MJ &0.466 &0.383 &\cellcolor[HTML]{cfe2f3}\textbf{0.743} &0.624 &0.610 &\cellcolor[HTML]{cfe2f3}\textbf{0.654} \\
&\scriptsize{\dalle{}} &0.372 &0.327 &\cellcolor[HTML]{cfe2f3}\textbf{0.632} &0.713 &0.477 &\cellcolor[HTML]{cfe2f3}\textbf{0.754} \\\arrayrulecolor{black!30}\specialrule{.2pt}{0.8pt}{1.5pt}
\multirow{3}{*}{\rotatebox[origin=c]{90}{crop}} &Imagen &0.436 &0.561 &\cellcolor[HTML]{cfe2f3}\textbf{0.781} &0.704 &0.546 &\cellcolor[HTML]{cfe2f3}\textbf{0.797} \\
&MJ &0.471 &0.383 &\cellcolor[HTML]{cfe2f3}\textbf{0.742} &0.623 &0.597 &\cellcolor[HTML]{cfe2f3}\textbf{0.680} \\
&\scriptsize{\dalle{}} &0.375 &0.298 &\cellcolor[HTML]{cfe2f3}\textbf{0.642} &0.702 &0.457 &\cellcolor[HTML]{cfe2f3}\textbf{0.769} \\
\arrayrulecolor{black}\bottomrule
\end{tabular}
\caption{\textbf{Performance on Corrupted Images -- AUCROC}. Our method remains robust to common image degredations~\cite{frank2020leveraging}.}\label{tab:robustness}
\vspace{-1em}
\end{table}

\bfpar{Inversions are crucial for generalization} 
To ensure that the observed gains are not coming from a particular choice of hyperparameters in our detector, we performed an ablation training the exact same network using only RGB images and only absolute DDIM image reconstruction residuals \hbox{$\operatorname{Res} = |x - D(\hat{z}_0)|$} (similar to DIRE~\cite{wang2023dire}). Table~\ref{tab:rgb_res_abl} shows that both RGB and reconstruction residual-based models perform significantly worse than the proposed method that uses both the input image, its reconstruction, and the inversion map, confirming all three are essential to achieve state-of-art generalization to unseen detectors. In the \supp{} we show that text conditioning also helps generalization.

\begin{figure}
    \centering
    \vspace{-10px}
    \includegraphics[width=\linewidth,trim=2in 0 2in 0,clip]{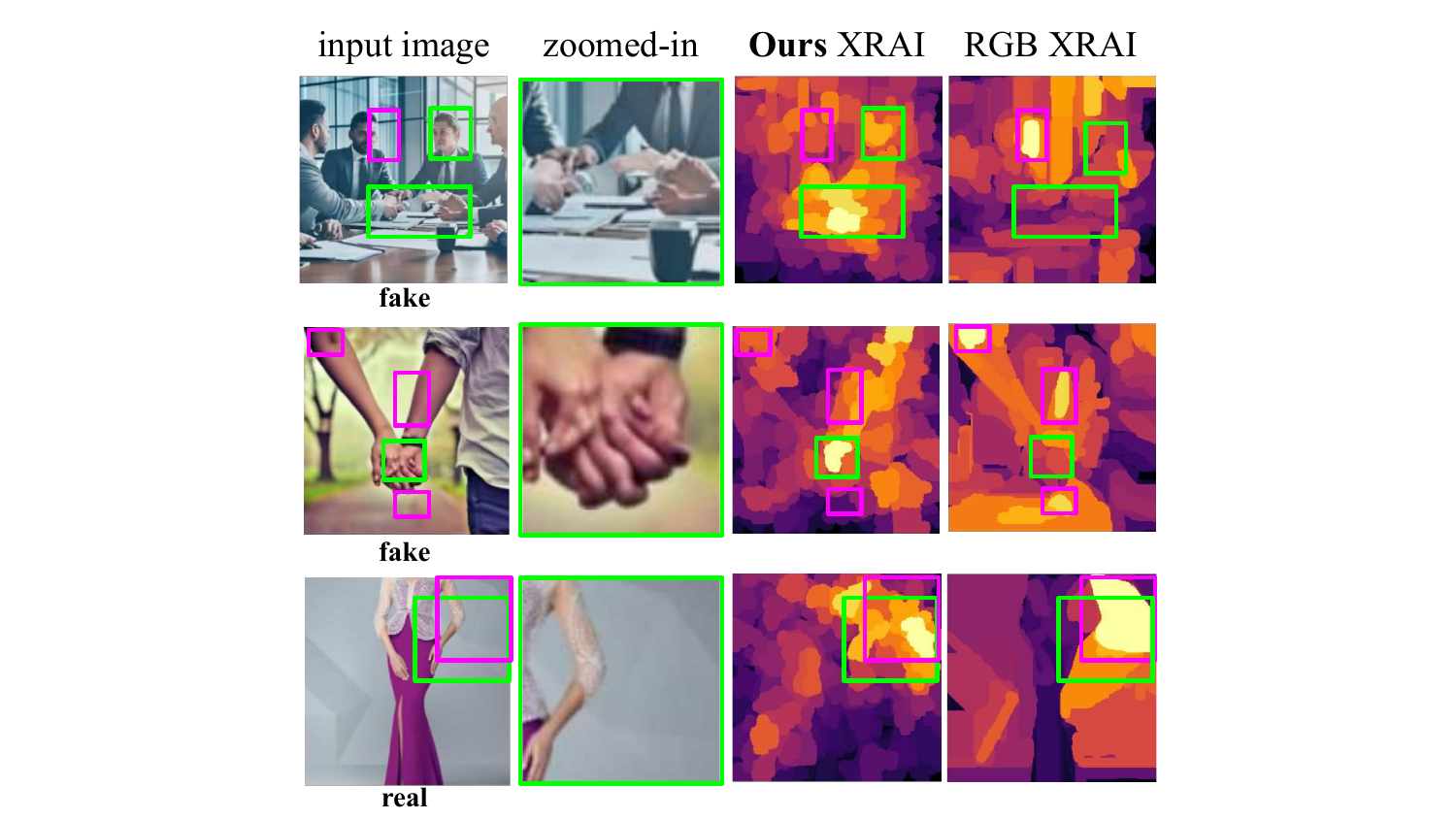}\vspace{-10px}
    \caption{\textbf{Saliency Analysis}. Green boxes highlight the most salient regions according to our model and purple boxes for an equivalent RGB-only model. We use a post-hoc explainability technique, XRAI \cite{kapishnikov2019xrai}. The regions of anatomical inconsistencies in fakes are most salient in our model.
    \label{fig:xrai}}
\end{figure}

\bfpar{Robustness and Interpretability} Table~\ref{tab:robustness} shows that our method is sufficiently robust to in-the-wild transformations \cite{frank2020leveraging} such as JPEG re-compression and blur. 
Figure~\ref{fig:xrai} shows that a model that uses inversion maps not only generalizes better but also focuses more on features that humans recognize as GenAI artifacts (\eg malformed hands). 

\bfpar{Discussion} Our results suggest that, while both our method and recent methods (UFD~\cite{ufd}, DMDet~\cite{corvi2023naples}) consistently outperform older prior methods (CNNDet~\cite{wang2019cnngenerated}) on prior academic benchmarks, recent methods struggle to maintain the same level of exceptional performance when evaluated against our new RIS-based evaluation benchmark, even when retrained on better data. Our method and some of the older baselines, on the other hand, perform well on both. We attribute this discrepancy to the drastic shift between real and fake images used in prior evaluation -- suggesting that some of the recent methods were in part overfitting to the distributions of styles and content of natural images, which appears to be less of an issue for our method.

\section{Conclusion}
\label{sec:conclusion}

In this paper, we introduce \textbf{\MethodName}: a GenAI detection method that uses text-conditioned inversion maps extracted from a pre-trained Stable Diffusion to achieve a new state-of-the-art at detecting images generated via unseen text-to-image diffusion models. We also propose \textbf{SynRIS}: a new challenging evaluation protocol that uses reverse image search to ensure that the evaluation is not biased towards any styles and themes. We show that the new protocol is also more reliable at evaluating detectors on images generated using proprietary models trained on unknown data. While \MethodName improves upon the state-of-the-art on this challenging benchmark, there clearly remains \textbf{much work to be done}; the detection performance on the new evaluation benchmark is far from saturated. We invite future researchers to use these new datasets to explore, build, and deploy better GenAI detectors at scale, with confidence that their solutions will not favor any content and style.

\clearpage 

\bibliographystyle{ieeenat_fullname}
\bibliography{_main}

\ifarxiv 
\clearpage 
\appendix 
\counterwithin{figure}{section}
\counterwithin{table}{section}
\label{sec:appendix_section}

\section*{Supplementary -- Table of Contents}
In this supplementary material, we provide following additional details regarding the proposed model and data:
\begin{itemize}
    \item In Section~\ref{sup:data} we describe \textbf{how traing and eval data was obtained}. 
    Visualizations of real and fake image pairs for all evaluation datasets in the proposed \textbf{SynRIS benchmark} can be found in the end of the document in Fig.~\ref{fig:imagen_samples}-\ref{fig:pixart_samples}.
    \item In Section \ref{supp:derivation} we provide a \textbf{derivation} of the relationship (\ref{eq:log_prob}) from the main paper - connecting likelihood with inversion maps, DDIM discretization and reconstruction errors.
    \item In Section~\ref{sup:training_details} we provide technical details of how our model was \textbf{trained}: architecture, inversion and captioning models we used.
    \item In Section~\ref{sup:extended_results} we provide \textbf{extended evaluation} results: Average Precision (AP) and overall accuracy for all evaluated methods, ROC, PR and DET curves, evaluation of robustness to prompt shift  and fine-tuning, and the effect of text conditioning.
    \item In Section~\ref{sup:baselines} we discuss how baselines were trained and issues we encountered when evaluating DIRE.
    \item In Section \ref{sup:data_viz}, we include sample visualizations from the various datasets of our \textbf{SynRIS benchmark}.
\end{itemize}

\section*{Ethics and Limitations}
The ultimate goal of this work is to prevent abuse and the spread of misinformation, an inherently ethical task. When generating our datasets, we sourced our prompts from an existing database. 
As such, some of the generated images may inadvertently contain inappropriate content.
We explicitly address misalignment between fake and real images in our training and evaluation to ensure that the detector is not favouring any styles or themes to avoid marginalization of any groups.

While our method performs well at detecting images from existing diffusion models, this same performance may not transfer well to text-to-image models that do \textit{not} make use of diffusion, such as text-to-image GANs (GigaGAN \cite{kang2023scaling}) and transformers (Muse \cite{muse}). Training our model also requires significantly more compute than similar methods (CNNDetect~\cite{wang2019cnngenerated}, DMDetect~\cite{corvi2023naples}) since we must first pass all training images through our inversion pipeline.

\section*{Acknowledgements}
We would like to thank J.~P.~Lewis, David Marwood, Shumeet Baluja, Sergey Ioffe and Arkanath Pathak for their feedback and technical advise.

\section{Data}\label{sup:data}
In this section we discuss how training and evaluation datasets were generated. Our new training set along with all our RIS-based evaluation benchmark can be found at our project page: [will be released with camera ready].
\subsection{Training Data}\label{sup:train_data}
We train our method and other baselines on two different training sets: ProGAN + LSUN and DiffusionDB + LAION (DDB-L).
\subsubsection{ProGAN+LSUN}
Authors of CNNDetect \cite{wang2019cnngenerated} introduced this dataset along with their GAN detection method. 
This dataset consists of 360k real images from LSUN~\cite{yu2015lsun} and 360k fake images generated by ProGAN~\cite{karras2019progan}, each composed of 20 different classes.
All real and fake images are 256x256 resolution.
\subsubsection{Stable Diffusion+LAION}\label{sec:ddb-laion}
For this training set, we took random 300k fake images from DiffusionDB~\cite{wang2022diffusiondb} and random 300k images from LAION~\cite{schuhmann2022laion} with predicted aesthetic scores of 6.25 or higher. Note that DiffusionDB consists of images from Stable Diffusion v1.

\subsection{Evaluation Data - Fakes}\label{sup:eval_fake}
This sections provides detail how each evaluation dataset's \textit{fake} images were obtained.
\subsubsection{Imagen}\label{sup:imagen}
We obtained Imagen images from authors of Imagen -- they generated them using an internal closed API using the same prompt distribution that was used to train the Imagen model.

\subsubsection{Midjourney}
For our Midjourney images, we use this dataset on Hugging Face (\href{https://huggingface.co/datasets/wanng/midjourney-v5-202304-clean}{link}).
These images were scraped from the Midjourney Discord server. 
This dataset includes a tag indicating whether or not the image was ``upscaled'' by the user.
We choose images that had been upscaled since they are presumably of higher quality (since the user spent additional credits to upscale them).

\subsubsection{\dalle{}}
For our \dalle{} images, we use this dataset on Hugging Face (\href{https://huggingface.co/datasets/laion/dalle-3-dataset}{link}).
These images were generated by users and shared on the LAION Discord server.

\subsubsection{Kandinsky 2}
For our Kandinsky 2 images, we use the Kandinsky 2.2 \cite{kandinsky2_2} model from Hugging Face (\href{https://huggingface.co/kandinsky-community/kandinsky-2-2-decoder}{link}), using the default parameters given in their usage example:
\begin{itemize}
    \item \texttt{prior\_guidance\_scale=1.0}
    \item \texttt{height=768}
    \item \texttt{width=768}
    \item \texttt{negative\_prompt=``low quality, bad quality''}
\end{itemize}

\subsubsection{Kandinsky 3}
For our Kandinsky 3 images, we use the Kandinsky 3 \cite{kandinsky3} model from Hugging Face (\href{https://huggingface.co/kandinsky-community/kandinsky-3}{link}), using the default parameters given in their usage example:
\begin{itemize}
    \item \texttt{num\_inference\_steps=50}
\end{itemize}

\subsubsection{PixArt-$\alpha$}
For our PixArt-$\alpha$ \cite{chen2023pixart} images, we use the 1024 resolution model from Hugging Face (\href{https://huggingface.co/PixArt-alpha/PixArt-XL-2-1024-MS}{link}).
All parameters are left as their defaults.

\subsubsection{Playground 2.5}
For our Playground 2.5 images, we use the Playground 2.5 \cite{li2024playground} model from Hugging Face (\href{https://huggingface.co/playgroundai/playground-v2.5-1024px-aesthetic}{link}), using the default parameters given in their usage example:
\begin{itemize}
    \item \texttt{num\_inference\_steps}
    \item \texttt{guidance\_scale=3}
\end{itemize}

\subsubsection{SDXL Direct Preference Optimization}
For our SDXL-DPO images, we use the SDXL-DPO \cite{dpo} model from Hugging Face (\href{https://huggingface.co/mhdang/dpo-sdxl-text2image-v1}{link}), with the default parameters given in their usage example:
\begin{itemize}
    \item \texttt{guidance\_scale=5}
\end{itemize}

\subsubsection{Stable Diffusion XL}
For our SDXL images, we use the Stable Diffusion XL \cite{sdxl} model from Hugging Face (\href{https://huggingface.co/stabilityai/stable-diffusion-xl-base-1.0}{link}), using both the base and refiner models with the default parameters given in their usage example:
\begin{itemize}
    \item \texttt{num\_inference\_steps=40}
    \item \texttt{denoising\_end=0.8}
    \item \texttt{denoising\_start=0.8}
\end{itemize}

\subsubsection{Segmind Mixture of Experts}
For our Seg-MoE images, we use the SegMoE-4x2-v0 \cite{segmoe} model from Hugging Face (\href{https://huggingface.co/segmind/SegMoE-4x2-v0}{link}), with the default parameters given in their usage example:
\begin{itemize}
    \item \texttt{negative\_prompt=``nsfw, bad quality, worse quality''}
    \item \texttt{height=1024}
    \item \texttt{width=1024}
    \item \texttt{num\_inference\_steps=25}
    \item \texttt{guidance\_scale=7.5}
\end{itemize}

\subsubsection{Segmind Stable Diffusion 1B}
For our SSD-1B \cite{segmind} images, we use the SSD-1B model from Hugging Face (\href{https://huggingface.co/segmind/SSD-1B}{link}), using the default parameters given in their usage example:
\begin{itemize}
    \item \texttt{negative\_prompt="ugly, blurry, poor quality"}
\end{itemize}

\subsubsection{Stable Cascade}
For our Stable Cascade \cite{wurstchen} images, we use the Stable Cascade model from Hugging Face (\href{https://huggingface.co/stabilityai/stable-cascade}{link}), using the default parameters given in their usage example for the prior model:
\begin{itemize}
    \item \texttt{height=1024}
    \item \texttt{width=1024}
    \item \texttt{guidance\_scale=4.0}
    \item \texttt{num\_inference\_steps=20}
\end{itemize}
and decoder model:
\begin{itemize}
    \item \texttt{guidance\_scale=0.0}
    \item \texttt{num\_inference\_steps=10}
\end{itemize}

\subsubsection{Segmind Vega}
For our Segmind Vega \cite{segmind} images, we use the Segmind Vega model from Hugging Face (\href{https://huggingface.co/segmind/Segmind-Vega}{link}), using the default parameters given in their usage example:
\begin{itemize}
    \item \texttt{negative\_prompt="(worst quality, low quality, illustration, 3d, 2d, painting, cartoons, sketch)"}
\end{itemize}

\subsubsection{W\"urstchen 2}
For our W\"uerstchen \cite{wurstchen} images, we use the W\"uerstchen v2 model from Hugging Face (\href{https://huggingface.co/warp-ai/wuerstchen}{link}), using the default parameters given in their usage example:
\begin{itemize}
    \item \texttt{height=1024}
    \item \texttt{width=1024}
    \item \texttt{prior\_guidance\_scale=4.0}
    \item \texttt{decoder\_guidance\_scale=0.0}
\end{itemize}
We will also include images and results from W\"uerstchen~v3, which is currently in beta development.

\subsection{Evaluation Data - Reals\label{sup:eval_eval}}

The corresponding \textit{real} images for all of our evaluation sets were found via a reverse image search API provided by one of the major image search engines.

\begin{figure*}
    \centering
    \includegraphics[width=0.9\linewidth]{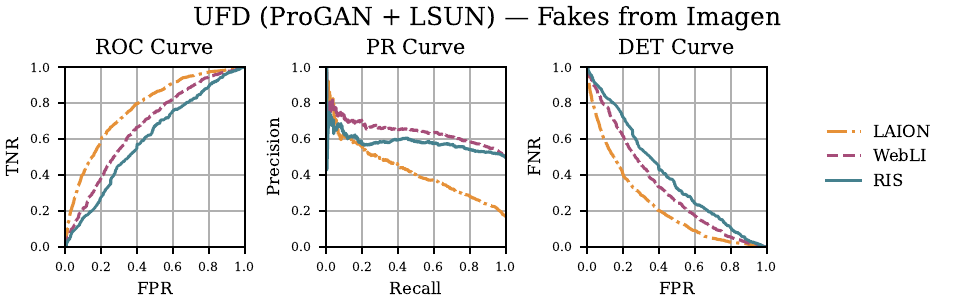}
    \caption{Receiver Operating Characteristic (\textbf{left}) Precision-Recall (\textbf{middle}) and Detection Error Tradeoff \textbf{(right)} curves for detecting Imagen versus real images from its training set WebLI \cite{yu2022parti} ({\color[HTML]{A64D79}{red}}), Reverse Image Search ({\color[HTML]{45818E}{green}}) and LAION ({\color[HTML]{E69138}{orange}}). These curves show that Imagen versus RIS is indeed a significantly harder task than Imagen versus LAION~\cite{schuhmann2022laion} and matches Imagen versus WebLI. }
    \label{fig:ris-curve}
\end{figure*}

\section{Derivations}\label{supp:derivation}
In this section we show the relationship between likelihood, DDIM inversion and DDIM reconstruction error. Notably, in the first approximation, it can be expressed using these terms without explicit dependency on model parameters $\theta$.

\subsection{Derivation of \cref{eq:log_prob}\label{sup:log_prob_derivation}}
Given an appropriate change of variable $\bf{f}$ we know:
\begin{equation*}
        \log p(\bf{x}) = \log p_z(\bf{f}^{-1}(x)) + \log\det\operatorname{\bf J}[\bf f^{-1}](x)
\end{equation*}

Rewriting the negative log Jacobain determinant as:
{
\newlength{\eqlength}
\settowidth{\eqlength}{$=$}

\begin{align*}
    & - \log\det\operatorname{\bf J} [\bf f^{-1}](x) \\
    = & \log\det\operatorname{\bf J}[\bf f](f^{-1}(x)) \\
    = & \operatorname{\bf Tr}\left(\log\operatorname{\bf J}[\bf f](f^{-1}(x))\right) \\
    = & \operatorname{\bf Tr}\left(\sum_{k=1}^{\infty}(-1)^{k+1}\frac{\left(\operatorname{\bf J}[\bf f](f^{-1}(x))- I)\right)^k}{k}\right) \\
    & \text{(take the first term)} \\
    \approx & \operatorname{\bf Tr}\left(\operatorname{\bf J}[\bf f](f^{-1}(x)) - I\right) \\
    \propto& \operatorname{\bf Tr}\left(\operatorname{\bf J}[\bf f](f^{-1}(x))\right) \\ & \text{(use Hutchinson estimator, assuming } \mu_v = 0, \Sigma_v = I) \\
    = & \mathbb{E}_{\bf v} \langle{\bf v}, \operatorname{\bf J}[\bf f](f^{-1}(x)) \ \bf{v} \rangle \\ 
    & \text{(reparametrize with $\delta$ such that } \mu_\delta = 0, \Sigma_\delta = \sigma_\delta I) \\
    = & \mathbb{E}_{\delta} \langle {\delta}, \operatorname{\bf J}[\bf f](f^{-1}(x)) \ \delta \rangle / \sigma_\delta^2 \\
    & \text{(take a single sample estimate)}\\
    \approx & \langle {\delta}, \operatorname{\bf J}[\bf f](f^{-1}(x)) \ \delta \rangle / \|\delta\|^2 \\
    & \text{(Taylor expansion of $\bf{f}$ around } \bf{f}^{-1}(x)\text{)} \\
    \approx & \langle \delta,  \bf{f}(\bf{f}^{-1}(x) + \delta) - \bf{f}(\bf{f}^{-1}(x))\rangle  / \|\delta\|^2 \\
    = & \langle \delta,  \bf{f}(\bf{f}^{-1}(x) + \delta) - x\rangle  / \|\delta\|^2
\end{align*}
}
and substituting 
\begin{gather*}
    \bf{x} = x_0 \\
    \bf{f}^{-1}(x) = x_T \\
    \bf{f}^{-1}(x) + \delta = \hat{x}_T \\
    \bf{f}(\bf{f}^{-1}(x) + \delta) = \hat{x}_0
\end{gather*}
and assuming a small enough isotropic $\delta$, in the first approximation, we get:
\begin{equation*}
        \log p(\bf{x}_0) \propto \log p_z(\bf{x}_T) - \langle \delta,  \hat{\bf{x}}_0 - \bf{x}_0\rangle  / \|\delta\|^2
\end{equation*}

\section{Training Details\label{sup:training_details}}
In this section, we detail the various components of the pipeline used to train our detector.

\subsection{Captioning}
We use the BLIP-2, OPT-2.7b \cite{blip} model from Hugging Face (\href{https://huggingface.co/stabilityai/stable-diffusion-2-1-unclip}{link}) to caption our images \textit{before} inversion and \textit{after} and augmentations. 
Captioning is done \textit{after} any augmentation since the augmentations could significantly change the caption (\eg{}, an RGB image converted to grayscale). 
\subsection{Inversion}
We base our inversion process on Pix-to-Pix Zero \cite{parmar2023zero}, making use of the Hugging Face implementation (\href{https://huggingface.co/docs/diffusers/api/pipelines/pix2pix_zero}{link}) without the additional attention map guidance and other regularizers. 

We first resize to 512$\times$512 and then invert \textit{all} images (training and evaluation) using the same Stable Diffusion 1.5 \cite{sd} checkpoint from Hugging Face (\href{https://huggingface.co/runwayml/stable-diffusion-v1-5}{link}).
\subsection{Training}
The original images, inverted noise maps, and denoised reconstructions are then concatenated along the channel dimension and used as input to a ResNet-50 \cite{he2016deep}.
We train our detector for 25 Epochs, but most of the performance is gained in the first few.
We otherwise use the same hyper-parameters as CNNDet \cite{wang2019cnngenerated}.

\section{Extended Results\label{sup:extended_results}}
\subsection{Extended metrics}
In Table \ref{tab:ap} and Table \ref{tab:eer} we show AP and Acc@EER metrics for all experiments in addition to AUCROC. Figures \cref{fig:roc_progan,fig:roc_ddblaion,fig:pr_progan,fig:pr_ddblaion,fig:det_progan,fig:det_ddblaion} show ROC, PR and DET curves for all compared classifiers.
\begin{table*}[!htp]\centering
\begin{tabular}{lcccccc}\toprule
\multicolumn{1}{r}{Train Data} &\multicolumn{3}{c}{ProGAN + LSUN} &\multicolumn{3}{c}{Stable Diffusion + LAION} \\ \cmidrule(lr){1-1}\cmidrule(lr){2-4} \cmidrule(lr){5-7}
Eval Set\hspace{1em}\textbar{}\hspace{1em}Model &CNNDet &UFD &Ours &CNNDet\textsuperscript{\textdagger} &UFD\textsuperscript{\textdagger} &Ours \\ \midrule
DALL·E 2 \cite{ramesh2022hierarchical} &0.499 &0.694 &\cellcolor[HTML]{cfe2f3}\textbf{0.867} &0.653 &0.750 &\cellcolor[HTML]{cfe2f3}\textbf{0.751} \\
DALL·E 3 \cite{dalle3} &0.473 &0.384 &\cellcolor[HTML]{cfe2f3}\textbf{0.625} &0.703 &0.474 &\cellcolor[HTML]{cfe2f3}\textbf{0.756} \\
Midjourney v5/6 \cite{midjourney} &0.498 &0.419 &\cellcolor[HTML]{cfe2f3}\textbf{0.736} &0.602 &0.555 &\cellcolor[HTML]{cfe2f3}\textbf{0.643} \\
Imagen \cite{imagen} &0.474 &0.582 &\cellcolor[HTML]{cfe2f3}\textbf{0.759} &0.705 &0.553 &\cellcolor[HTML]{cfe2f3}\textbf{0.791} \\ \midrule
Kandinsky 2 \cite{kandinsky2_2} &0.493 &0.479 &\cellcolor[HTML]{cfe2f3}\textbf{0.764} &0.592 &0.547 &\cellcolor[HTML]{cfe2f3}\textbf{0.695} \\
Kandinsky 3 \cite{kandinsky3} &0.491 &0.470 &\cellcolor[HTML]{cfe2f3}\textbf{0.860} &0.654 &0.605 &\cellcolor[HTML]{cfe2f3}\textbf{0.755} \\
PixArt-$\alpha$ \cite{chen2023pixart} &0.490 &0.502 &\cellcolor[HTML]{cfe2f3}\textbf{0.871} &0.623 &0.625 &\cellcolor[HTML]{cfe2f3}\textbf{0.744} \\
Playground 2.5 \cite{li2024playground} &0.510 &0.462 &\cellcolor[HTML]{cfe2f3}\textbf{0.778} &0.556 &0.571 &\cellcolor[HTML]{cfe2f3}\textbf{0.617} \\
SDXL-DPO \cite{dpo} &0.495 &0.475 &\cellcolor[HTML]{cfe2f3}\textbf{0.849} &0.829 &0.683 &\cellcolor[HTML]{cfe2f3}\textbf{0.874} \\
SDXL \cite{sdxl} &0.506 &0.470 &\cellcolor[HTML]{cfe2f3}\textbf{0.777} &\cellcolor[HTML]{cfe2f3}\textbf{0.799} &0.651 &0.792 \\
Seg-MOE \cite{segmoe} &0.490 &0.428 &\cellcolor[HTML]{cfe2f3}\textbf{0.800} &0.644 &0.611 &\cellcolor[HTML]{cfe2f3}\textbf{0.704} \\
SSD-1B \cite{segmind} &0.544 &0.509 &\cellcolor[HTML]{cfe2f3}\textbf{0.840} &0.714 &0.613 &\cellcolor[HTML]{cfe2f3}\textbf{0.787} \\
Stable-Cascade \cite{wurstchen} &0.508 &0.399 &\cellcolor[HTML]{cfe2f3}\textbf{0.892} &0.712 &0.656 &\cellcolor[HTML]{cfe2f3}\textbf{0.766} \\
Segmind Vega \cite{segmind} &0.524 &0.475 &\cellcolor[HTML]{cfe2f3}\textbf{0.834} &0.723 &0.612 &\cellcolor[HTML]{cfe2f3}\textbf{0.796} \\
W\"urstchen 2 \cite{wurstchen} &0.508 &0.592 &\cellcolor[HTML]{cfe2f3}\textbf{0.803} &0.600 &0.670 &\cellcolor[HTML]{cfe2f3}\textbf{0.712} \\ \midrule
DALL·E 2 \cite{ramesh2022hierarchical} (A) &0.160 &0.256 &\cellcolor[HTML]{cfe2f3}\textbf{0.620} &0.202 &0.216 &\cellcolor[HTML]{cfe2f3}\textbf{0.587} \\
Craiyon \cite{Dayma_DALL·E_Mini_2021} (A) &0.621 &\cellcolor[HTML]{cfe2f3}\textbf{0.977} &0.893 &0.737 &\cellcolor[HTML]{cfe2f3}\textbf{0.922} &0.876 \\
LDM \cite{sd} (A) &0.598 &\cellcolor[HTML]{cfe2f3}\textbf{0.933} &0.897 &0.901 &0.924 &\cellcolor[HTML]{cfe2f3}\textbf{0.976} \\ \midrule
Average &0.493 &0.528 &\cellcolor[HTML]{cfe2f3}\textbf{0.804} &0.664 &0.624 &\cellcolor[HTML]{cfe2f3}\textbf{0.757} \\
\bottomrule
\end{tabular}
\caption{\textbf{Main Results -- Detector Average Precision.}
 In the main paper, we report the AUCROC metric when evaluating our classifiers. We report AP (Average Precision) here for completeness as well. We observe similar trends: our method performs best across nearly all datasets. $^*$Note: This DMDet classifier was trained with fakes from an LDM checkpoint rather than Stable Diffusion. \textsuperscript{\textdagger}These models were re-trained by us.}\label{tab:ap}
\end{table*}

\begin{table*}[!htp]\centering

\begin{tabular}{lcccccc}\toprule
\multicolumn{1}{r}{Train Data} &\multicolumn{3}{c}{ProGAN + LSUN} &\multicolumn{3}{c}{Stable Diffusion + LAION} \\\cmidrule(lr){1-1}\cmidrule(lr){2-4} \cmidrule(lr){5-7}
Eval Set\hspace{1em}\textbar{}\hspace{1em}Model &CNNDet &UFD &Ours &CNNDet\textsuperscript{\textdagger} &UFD\textsuperscript{\textdagger} &Ours \\ \midrule
DALL·E 2 \cite{ramesh2022hierarchical} &0.470 &0.674 &\cellcolor[HTML]{cfe2f3}\textbf{0.773} &0.624 &\cellcolor[HTML]{cfe2f3}\textbf{0.700} &0.678 \\
DALL·E 3 \cite{dalle3} &0.435 &0.371 &\cellcolor[HTML]{cfe2f3}\textbf{0.592} &0.659 &0.473 &\cellcolor[HTML]{cfe2f3}\textbf{0.698} \\
Midjourney v5/6 \cite{midjourney} &0.490 &0.413 &\cellcolor[HTML]{cfe2f3}\textbf{0.661} &0.595 &0.558 &\cellcolor[HTML]{cfe2f3}\textbf{0.606} \\
Imagen \cite{imagen} &0.470 &0.580 &\cellcolor[HTML]{cfe2f3}\textbf{0.706} &0.674 &0.538 &\cellcolor[HTML]{cfe2f3}\textbf{0.720} \\ \midrule
Kandinsky 2 \cite{kandinsky2_2} &0.490 &0.483 &\cellcolor[HTML]{cfe2f3}\textbf{0.687} &0.574 &0.541 &\cellcolor[HTML]{cfe2f3}\textbf{0.652} \\
Kandinsky 3 \cite{kandinsky3} &0.481 &0.478 &\cellcolor[HTML]{cfe2f3}\textbf{0.766} &0.609 &0.600 &\cellcolor[HTML]{cfe2f3}\textbf{0.684} \\
PixArt-$\alpha$ \cite{chen2023pixart} &0.485 &0.504 &\cellcolor[HTML]{cfe2f3}\textbf{0.769} &0.591 &0.606 &\cellcolor[HTML]{cfe2f3}\textbf{0.669} \\
Playground 2.5 \cite{li2024playground} &0.508 &0.477 &\cellcolor[HTML]{cfe2f3}\textbf{0.707} &0.553 &0.562 &\cellcolor[HTML]{cfe2f3}\textbf{0.591} \\
SDXL-DPO \cite{dpo} &0.486 &0.473 &\cellcolor[HTML]{cfe2f3}\textbf{0.764} &0.761 &0.647 &\cellcolor[HTML]{cfe2f3}\textbf{0.801} \\
SDXL \cite{sdxl} &0.497 &0.473 &\cellcolor[HTML]{cfe2f3}\textbf{0.688} &0.735 &0.620 &\cellcolor[HTML]{cfe2f3}\textbf{0.737} \\
Seg-MOE \cite{segmoe} &0.472 &0.429 &\cellcolor[HTML]{cfe2f3}\textbf{0.725} &0.625 &0.586 &\cellcolor[HTML]{cfe2f3}\textbf{0.664} \\
SSD-1B \cite{segmind} &0.545 &0.506 &\cellcolor[HTML]{cfe2f3}\textbf{0.748} &0.665 &0.585 &\cellcolor[HTML]{cfe2f3}\textbf{0.724} \\
Stable-Cascade \cite{wurstchen} &0.477 &0.383 &\cellcolor[HTML]{cfe2f3}\textbf{0.802} &0.652 &0.633 &\cellcolor[HTML]{cfe2f3}\textbf{0.694} \\
Segmind Vega \cite{segmind} &0.522 &0.476 &\cellcolor[HTML]{cfe2f3}\textbf{0.741} &0.676 &0.587 &\cellcolor[HTML]{cfe2f3}\textbf{0.733} \\
W\"urstchen 2 \cite{wurstchen} &0.504 &0.580 &\cellcolor[HTML]{cfe2f3}\textbf{0.715} &0.580 &0.640 &\cellcolor[HTML]{cfe2f3}\textbf{0.658} \\ \midrule
DALL·E 2 \cite{ramesh2022hierarchical} (A) &\cellcolor[HTML]{cfe2f3}\textbf{0.656} &0.620 &0.590 &0.556 &\cellcolor[HTML]{cfe2f3}\textbf{0.562} &0.558 \\
Craiyon \cite{Dayma_DALL·E_Mini_2021} (A) &0.610 &\cellcolor[HTML]{cfe2f3}\textbf{0.917} &0.783 &0.699 &\cellcolor[HTML]{cfe2f3}\textbf{0.836} &0.810 \\
LDM \cite{sd} (A) &0.591 &\cellcolor[HTML]{cfe2f3}\textbf{0.845} &0.793 &0.837 &0.840 &\cellcolor[HTML]{cfe2f3}\textbf{0.933} \\ \midrule
Average &0.510 &0.538 &\cellcolor[HTML]{cfe2f3}\textbf{0.723} &0.648 &0.617 &\cellcolor[HTML]{cfe2f3}\textbf{0.700} \\
\bottomrule
\end{tabular}
\caption{\textbf{Main Results -- Acc @ Equal Error Rate.}
 We further report Acc@EER here as an additional metric. We observe similar trends: our method performs best across nearly all datasets. $^*$Note: This DMDet classifier was trained with fakes from an LDM checkpoint rather than Stable Diffusion. \textsuperscript{\textdagger}These models were re-trained by us.}\label{tab:eer}
\end{table*}

\subsection{Very Out-of-Distribution Data\label{sup:tuning}}
The generators analyzed thus far are trained to be \textit{general} aesthetic text-to-image models.

Here, we evaluate our model on models that have been fine-tuned generate images from \textit{very} specific domains: Anime and \pkmn{}.
Table \ref{tab:pkmn_anime} shows that even when using low-quality training data (ProGAN), our model still generalizes to these very out-of-distribution domains while existing methods (UFD~\cite{ufd}, CNNDet~\cite{wang2019cnngenerated}) completely fail, even performing worse than random guessing.
\begin{table}[]
    \centering
    \begin{tabular}{cccc}\toprule
        & \multicolumn{3}{c}{Detection Method}\\
        Eval Set & \textbf{Ours} & UFD & CNNDet \\ \midrule
        Anime & \textbf{0.67} & 0.13 & 0.34 \\
        \pkmn{} & \textbf{0.83} & 0.32 & 0.48\\
        \bottomrule
    \end{tabular}
    \caption{\textbf{AUCROC}. When trained on lower-quality data (\hbox{ProGAN}), existing methods (UFD~\cite{ufd}, CNNDet~\cite{wang2019cnngenerated}) \textit{completely} fail to generalize to our extremely out-of-distribution datasets, even doing worse than randomly guessing. On the other hand, our method continues to generalize well.}
    \label{tab:pkmn_anime}
\end{table}

\subsection{Text-Conditioning}
In Figure~\ref{fig:text_conditioning}, we show the effect of text-conditioning on the inversion-reconstruction process. 
By using text-conditioning, we recover a more faithful reconstruction of the input image, providing better signal for our model.
\begin{figure}
    \centering
    \includegraphics[width=\linewidth]{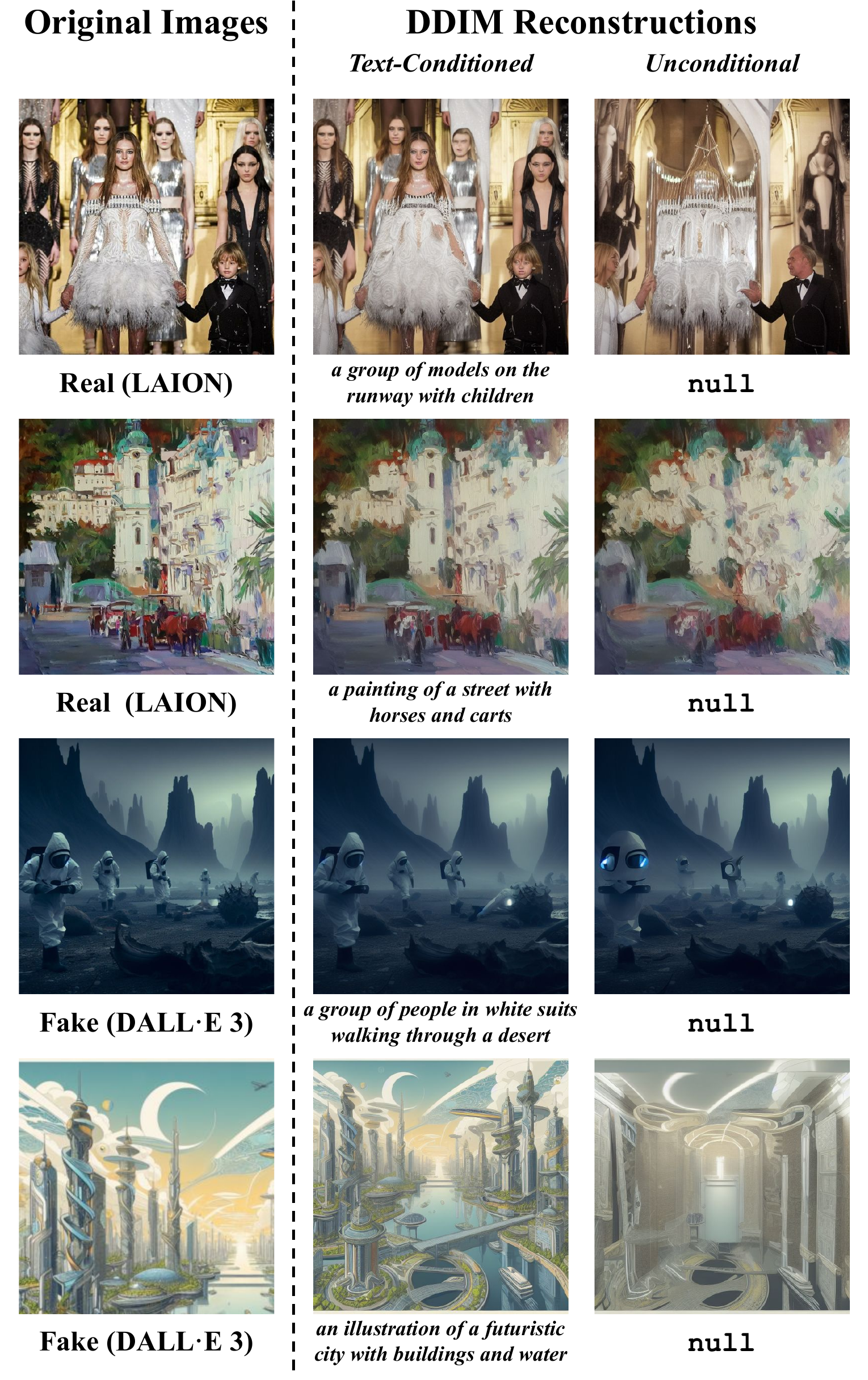}
    \caption{Unconditional DDIM inversion and reconstruction catastrophically fails to reconstruct the original image. To remedy this, we predict a caption using BLIP 2 \cite{blip} and use this for text conditioning.}
    \label{fig:text_conditioning}
\end{figure}

\begin{figure*}[p]
    \centering
    \includegraphics[]{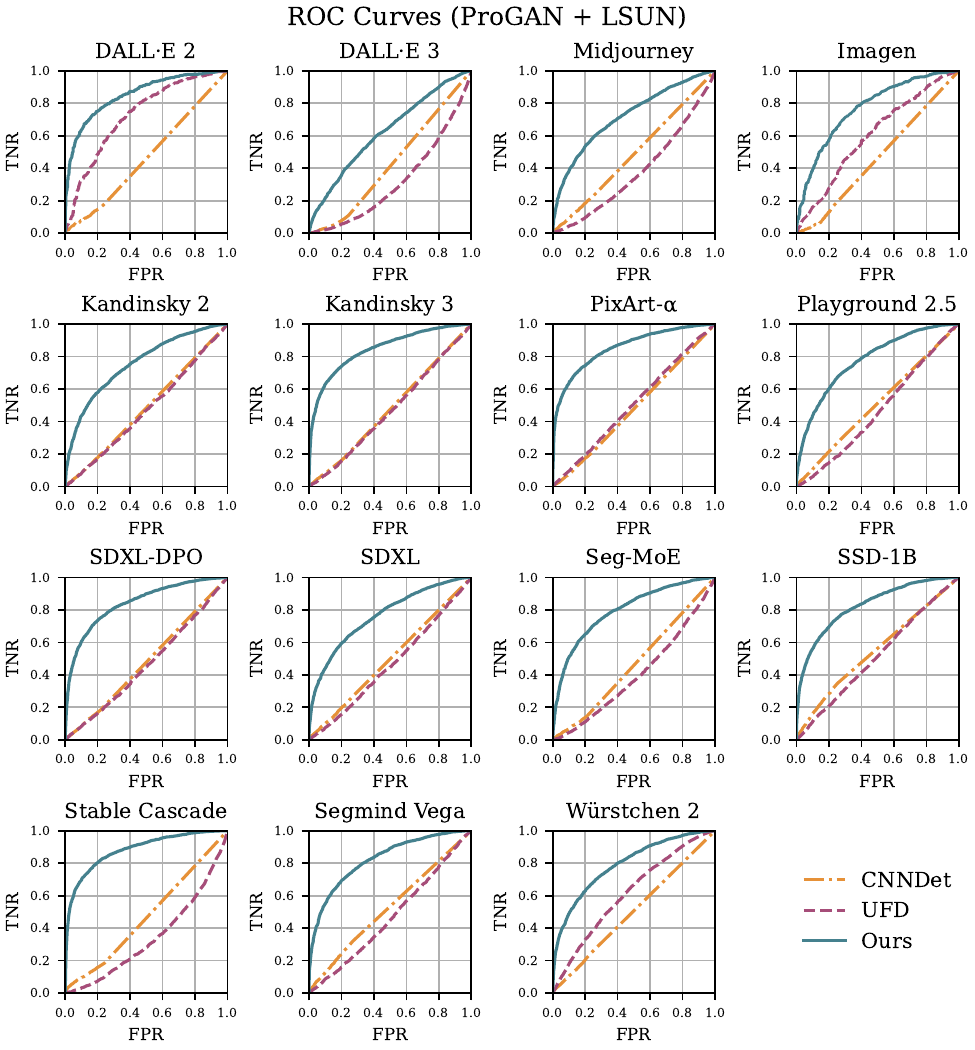}
    \caption{ROC curves for all 15 of our SynRIS evaluation datasets. Models trained on ProGAN+LSUN.}
    \label{fig:roc_progan}
\end{figure*}

\begin{figure*}
    \centering
    \includegraphics[]{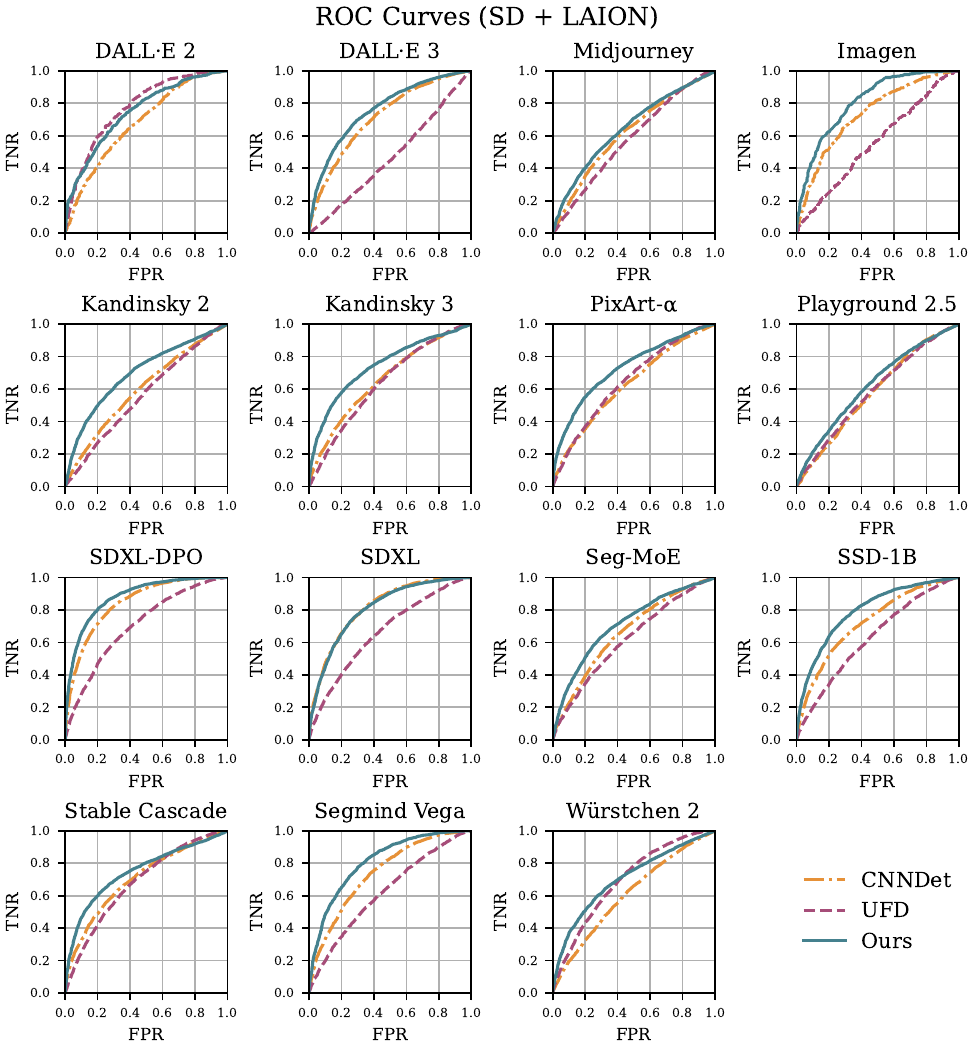}
    \caption{ROC curves for all 15 of our SynRIS evaluation datasets. Models trained on SD+LAION.}
    \label{fig:roc_ddblaion}
\end{figure*}

\begin{figure*}
    \centering
    \includegraphics[]{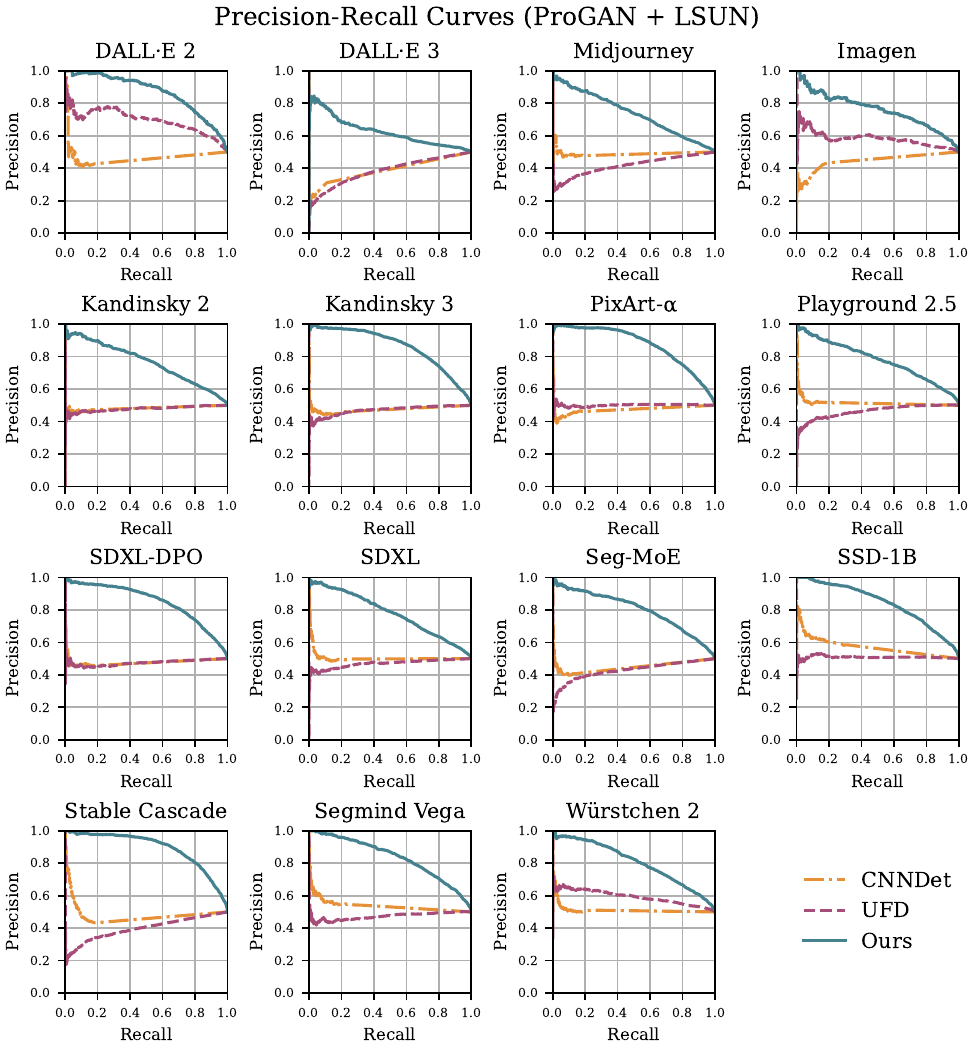}
    \caption{Precision-Recall curves for all 15 of our SynRIS evaluation datasets. Models trained on ProGAN+LSUN.}
    \label{fig:pr_progan}
\end{figure*}

\begin{figure*}
    \centering
    \includegraphics[]{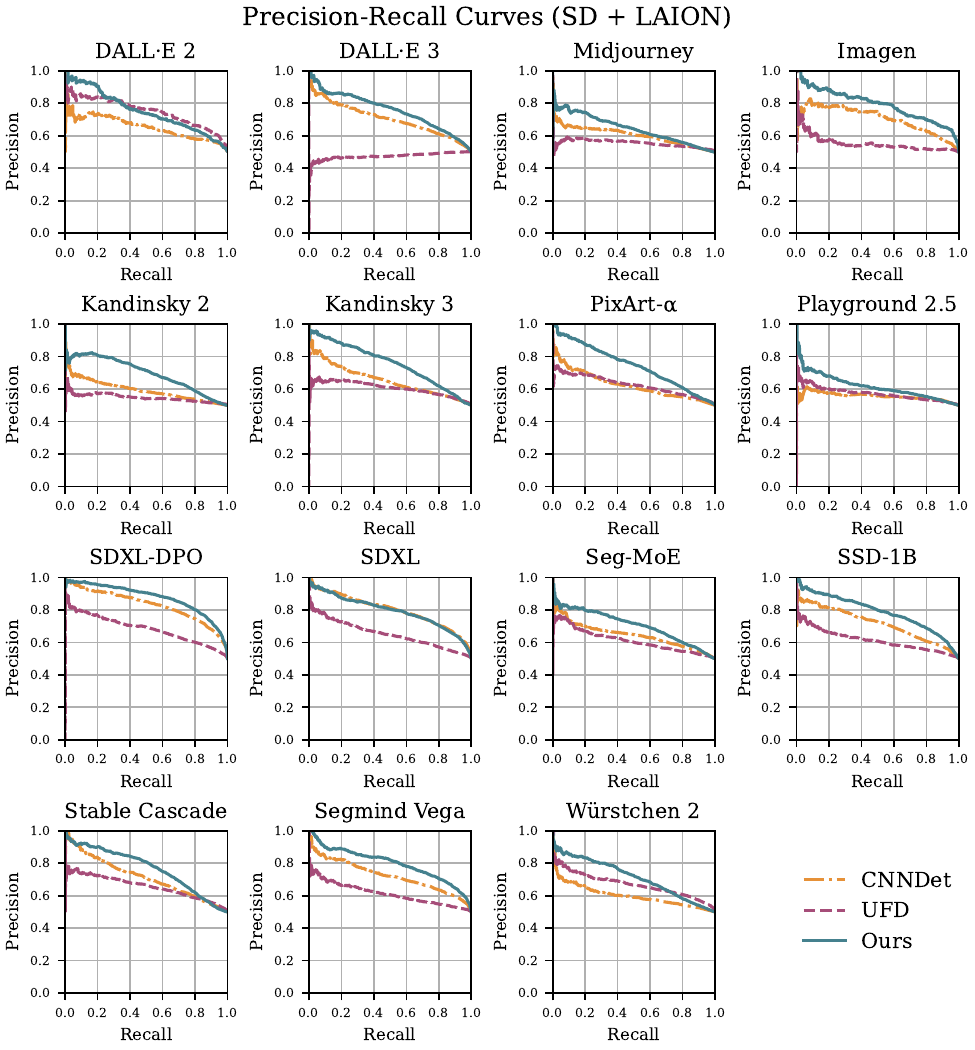}
    \caption{Precision-Recall curves for all 15 of our SynRIS evaluation datasets. Models trained on SD+LAION.}
    \label{fig:pr_ddblaion}
\end{figure*}

\begin{figure*}
    \centering
    \includegraphics[]{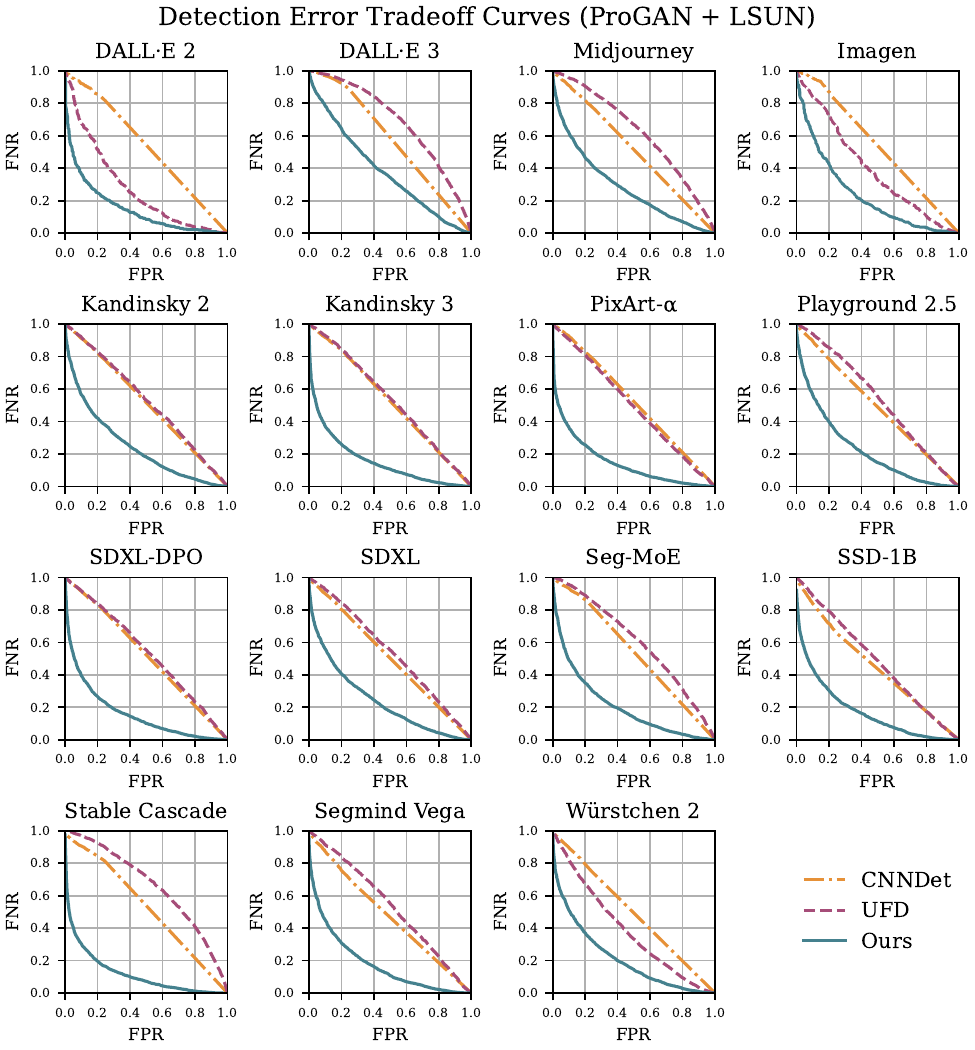}
    \caption{Detection Error Tradeoff curves for all 15 of our SynRIS evaluation datasets. Models trained on ProGAN+LSUN.}
    \label{fig:det_progan}
\end{figure*}

\begin{figure*}
    \centering
    \includegraphics[]{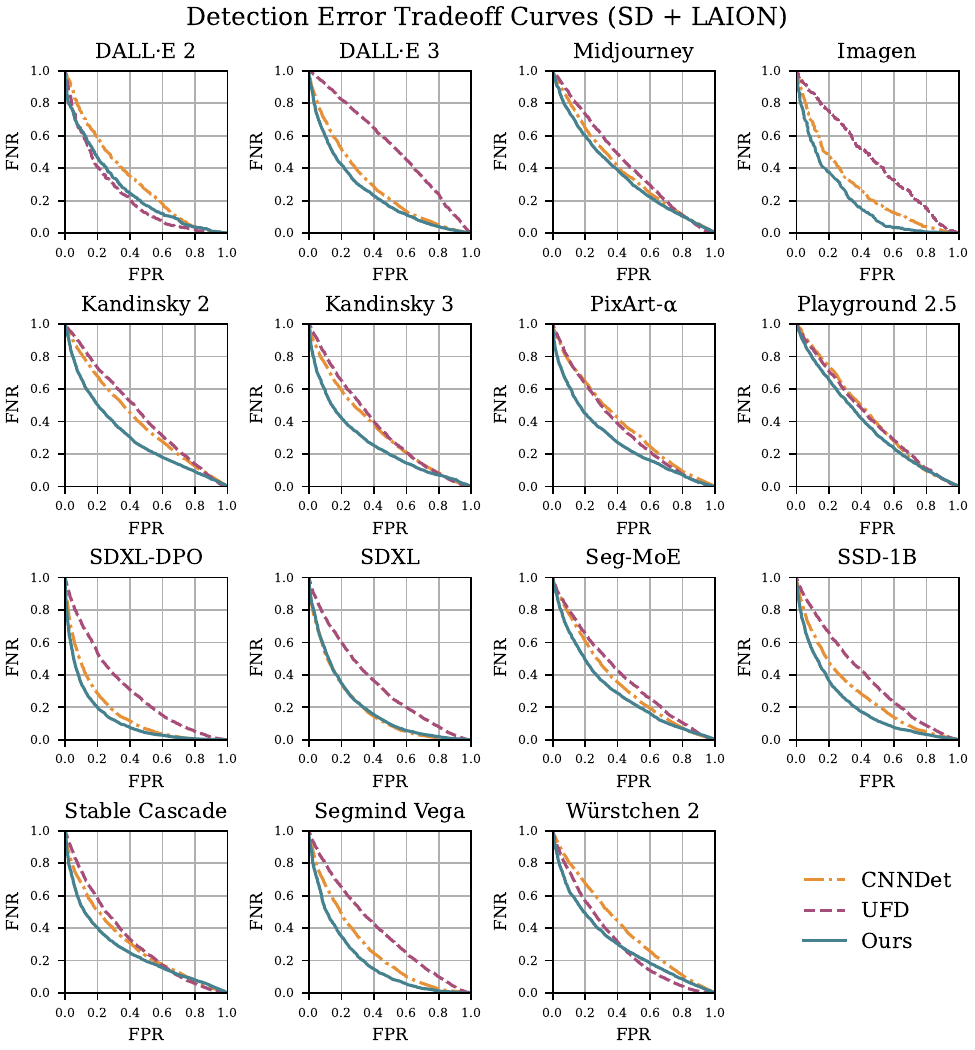}
    \caption{Detection Error Tradeoff curves for all 15 of our SynRIS evaluation datasets. Models trained on SD+LAION.}
    \label{fig:det_ddblaion}
\end{figure*}

\section{Baselines}\label{sup:baselines}
\subsection{DIRE \cite{wang2023dire}}\label{sup:dire}
\looseness=-1
The DIRE~\cite{wang2023dire} paper reports almost perfect detection performance on all unseen test sets. 
However, after their code was released, several researchers noticed a fundamental issue with the training and evaluation setup present in their released code and checkpoints (\href{https://github.com/ZhendongWang6/DIRE/issues/11}{link}) causing the in-the-wild performance to drop to near-random levels. More specifically, all pre-processed images used for training are saved with the \textbf{same extension} as the source images. 
Since all input \textit{real} images in the training set are saved as \texttt{*.JPG} files and all fake images are saved as \texttt{*.PNG} files, all real DIRE images used to train and evaluate the network were embedded with JPEG artifacts while the non of the fake images used for training and evaluation were.
As such, the model seemingly learned to detect the presence of JPEG artifacts. 
This holds even for the robustness experiments since augmented real and fake images are also saved as \texttt{JPG} and \texttt{PNG} respectively.
In all of \textit{our} datasets, both real and fake images are saved as lossless \texttt{*.PNG} files, explaining DIRE's poor performance on all our test sets.
To honor the contribution of DIRE authors, we conducted an ablation that used the exact same absolute residuals signals as reported in DIRE paper. 
It is evident from our results that the inversion signals we propose in this paper perform much better then DIRE signals across all evaluation benchmarks.

\vfill\null
\subsection{CNNDet \cite{wang2019cnngenerated}}
We trained using the original CNNDet training code (\href{https://github.com/peterwang512/CNNDetection}{link}) using following default parameters.
\begin{itemize}
    \item \texttt{--name blur\_jpg\_prob=0.5}
    \item \texttt{--blur\_prob=0.5}
    \item \texttt{--blur\_sigma=0.0,0.3}
    \item \texttt{--jpeg\_prob=0.5}
    \item \texttt{--jpeg\_method=cv2,pil}
    \item \texttt{--jpeg\_qual=30,100}
\end{itemize}
All training and eval images are resized to 256$\times$256 and saved as \texttt{.PNG} files.
\vfill\null
\subsection{UFD \cite{ufd}}
We trained using the original UFD training code (\href{https://github.com/Yuheng-Li/UniversalFakeDetect}{link})  using following default parameters.
\begin{itemize}
    \item \texttt{--name=clip\_vitl14}
    \item \texttt{--wang2020\_data\_path=datasets}
    \item \texttt{--data\_mode=wang2020}
    \item \texttt{--arch=CLIP:ViT-L/14}
    \item \texttt{--fix\_backbone}
\end{itemize}
All training and eval images are resized to 256$\times$256 and saved as \texttt{.PNG} files.
\vfill\null

\section{SynRIS Visualization\label{sup:data_viz}}
\cref{fig:imagen_samples,fig:mj_samples,fig:dalle2_samples,fig:dalle3_samples} contain samples from our evaluation sets using \textit{closed-source} models. 
\cref{fig:kandinsky2_samples,fig:kandinsky3_samples,fig:pixart_samples,fig:playground_samples,fig:dpo_samples,fig:sdxl_samples,fig:segmoe_samples,fig:ssd1b_samples,fig:cascade_samples,fig:vega_samples,fig:wurstchen_samples} contain samples from our evaluation sets using \textit{open-source} models. 
The corresponding images in each of these open-source model figures were generates using the same prompts.
The top panel consists of \textit{fake} images either found online or generated by us, and the bottom panel are the corresponding \textit{real} images found via reverse image search.

\cref{fig:pokemon_samples,fig:anime_samples} contain samples from our \pkmn{} and Anime datasets respectively. 
The real images were taken from a source dataset \cite{pinkney2022pokemon,danbooru2022} and the attached captions were used to generate the fake images with SD \pkmn{} Diffusers~\cite{pkmn_diffusers} and Animagine~\cite{animagine}.
\begin{figure*}
    \centering
    \centering
    \begin{center}
        \Large{Fake (Imagen \cite{imagen})}\\\vspace{0.25cm}
        \includegraphics[width=\linewidth]{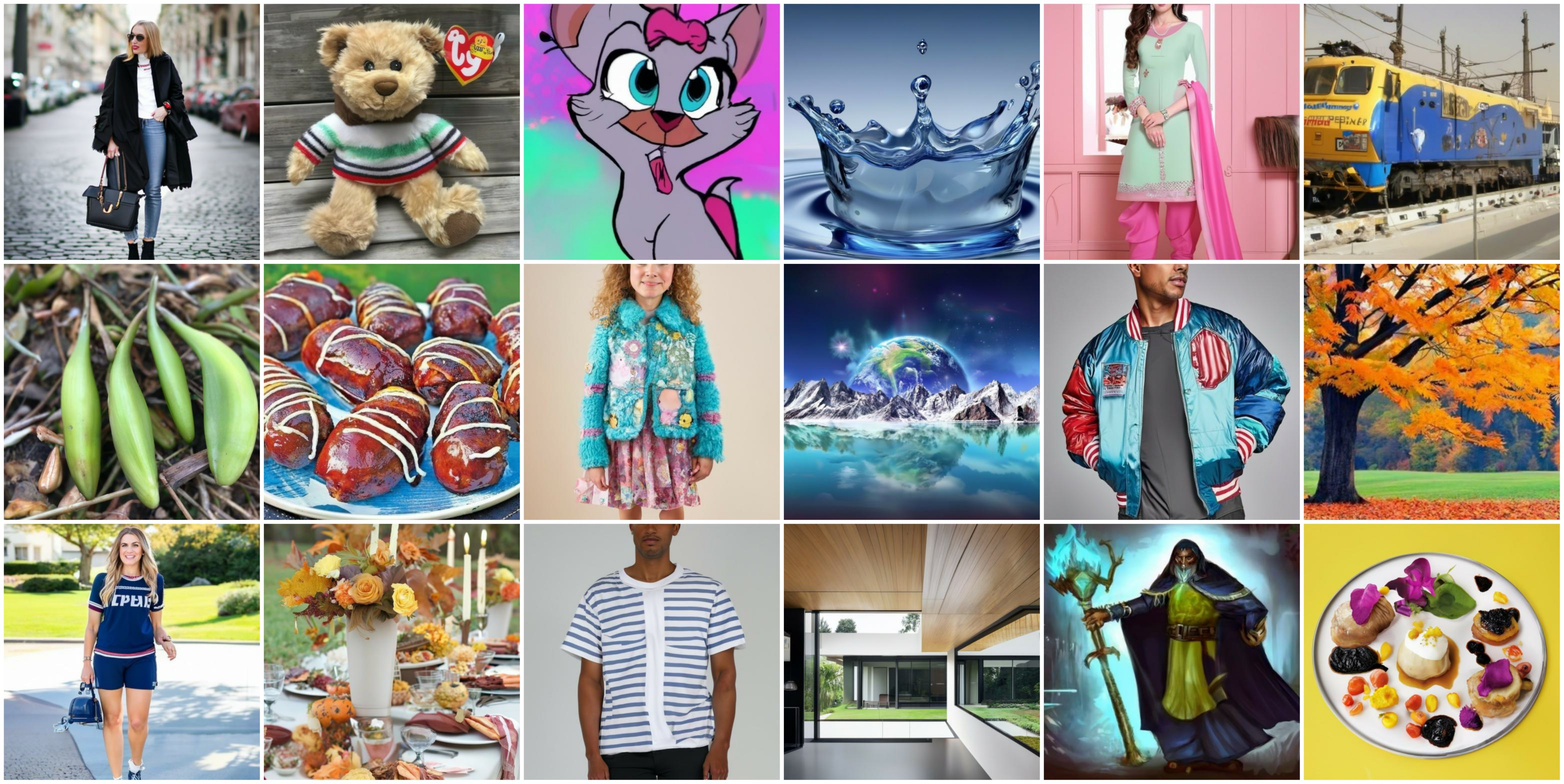}\\\vspace{0.8cm}
        \Large{Real (Reverse Image Search)}\\\vspace{0.25cm}
        \includegraphics[width=\linewidth]{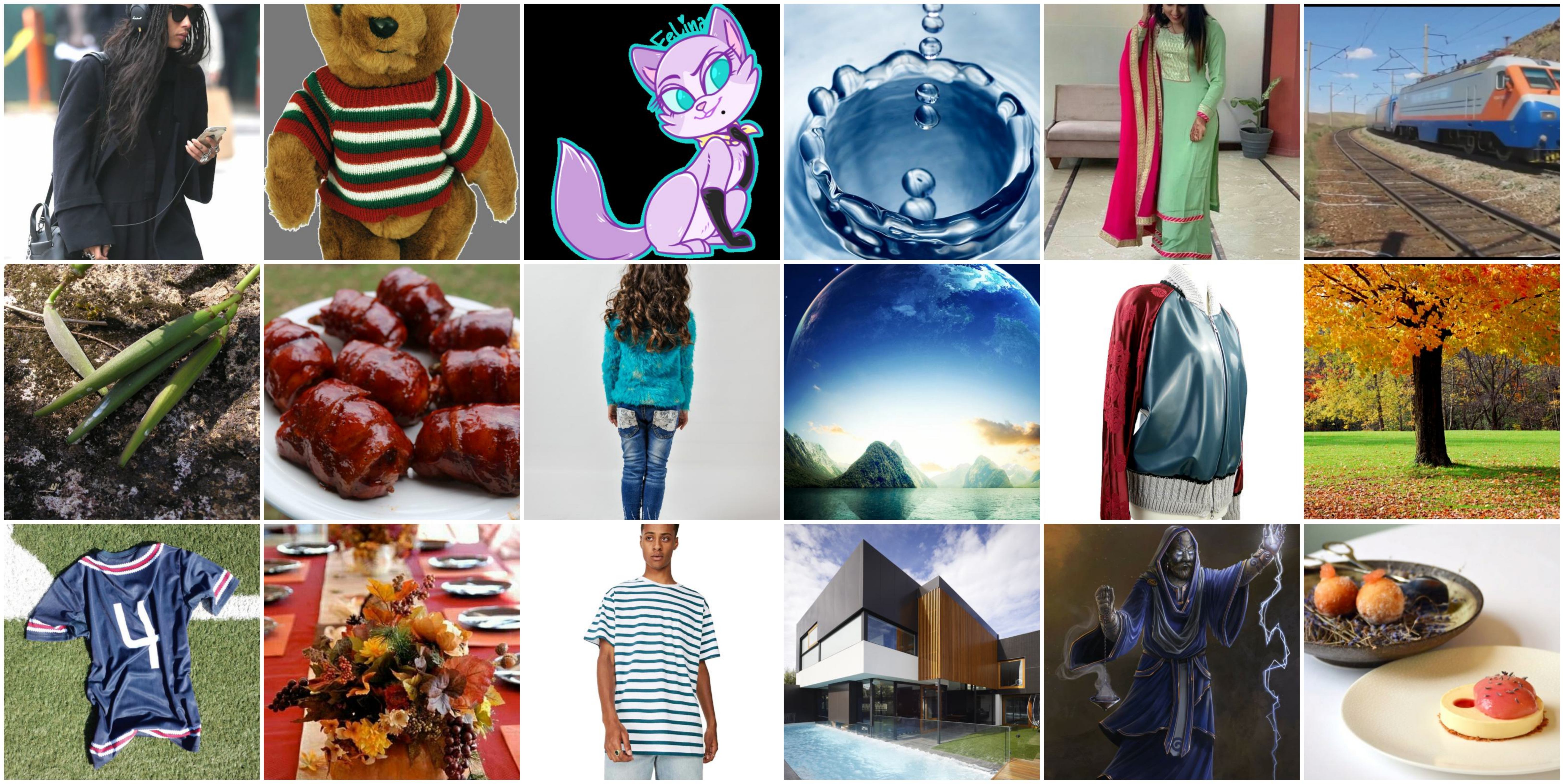}\\
    \end{center}
    \caption{\textbf{Top}: A sample of \textit{fake} Imagen \cite{imagen} images we generated for our dataset. \textbf{Bot}: Corresponding \textit{real} images found via RIS.}
    \label{fig:imagen_samples}
\end{figure*}
\begin{figure*}
    \centering
    \centering
    \begin{center}
        \Large{Fake (Midjourney \cite{midjourney})}\\\vspace{0.25cm}
        \includegraphics[width=\linewidth]{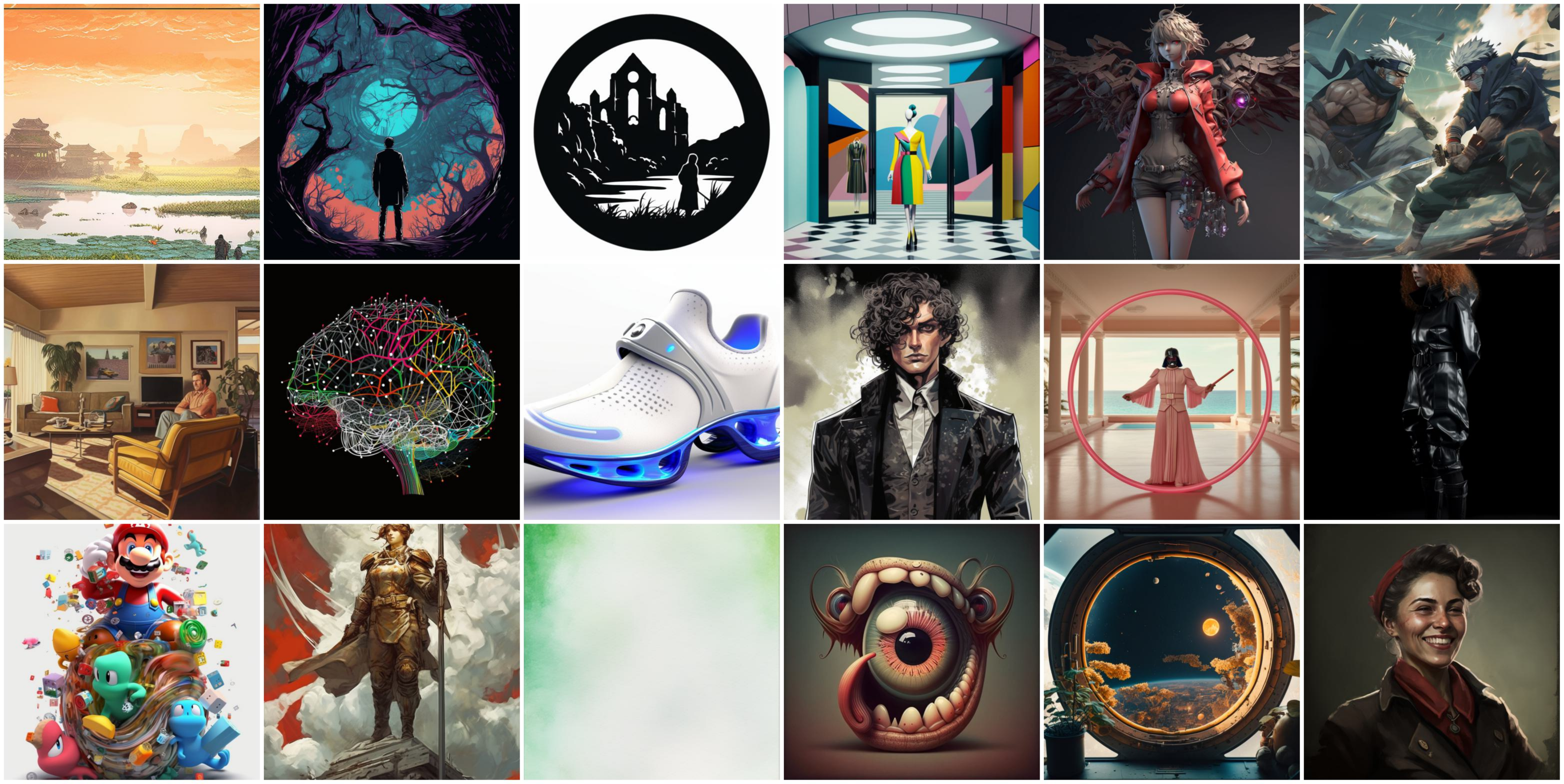}\\\vspace{0.8cm}
        \Large{Real (Reverse Image Search)}\\\vspace{0.25cm}
        \includegraphics[width=\linewidth]{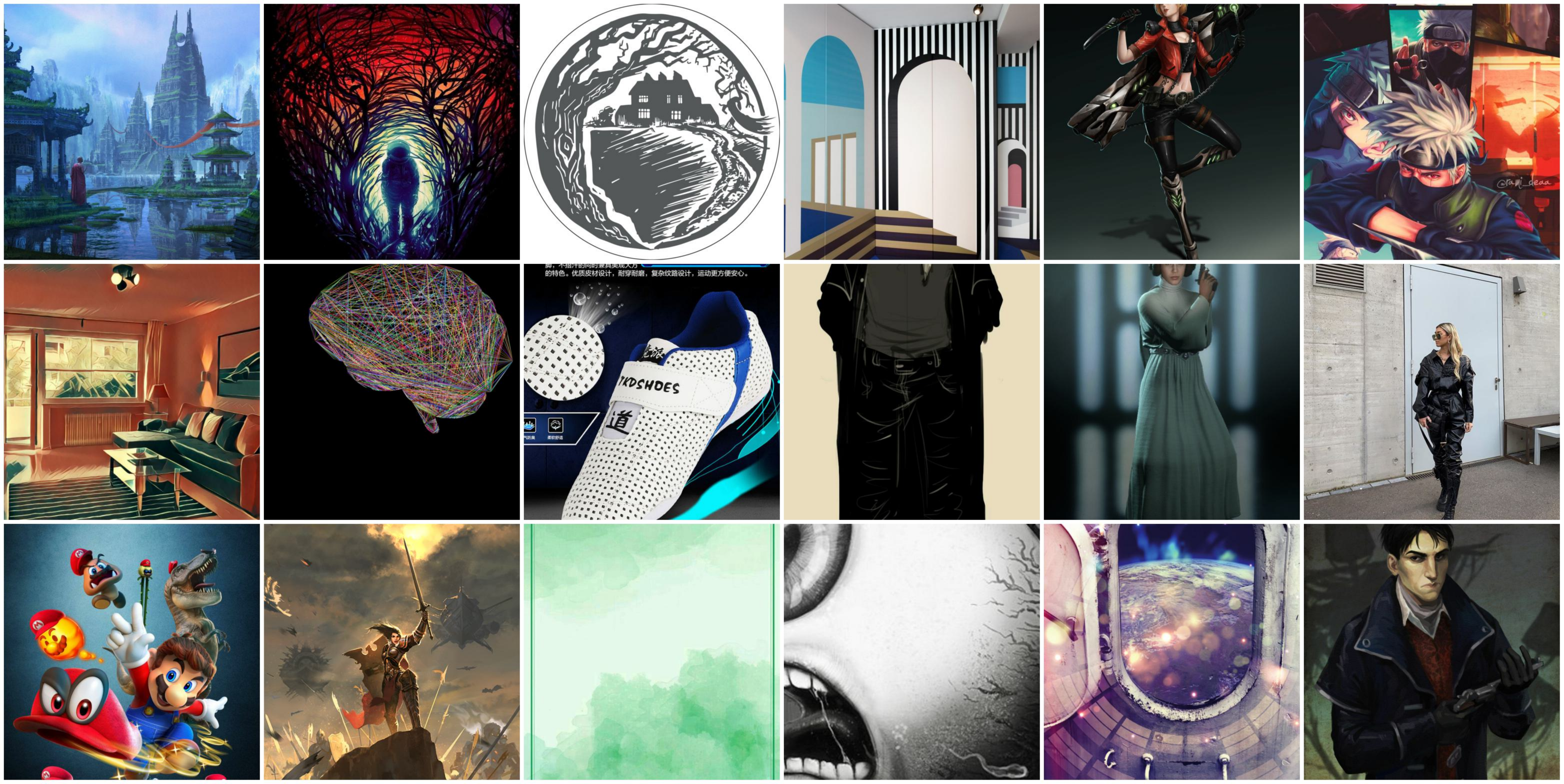}\\
    \end{center}
    \caption{\textbf{Top}: A sample of \textit{fake} Midjourney \cite{dalle3} images taken from Hugging Face (\href{https://huggingface.co/datasets/wanng/midjourney-v5-202304-clean}{link}). \textbf{Bot}: Corresponding \textit{real} images found via RIS.}
    \label{fig:mj_samples}
\end{figure*}

\begin{figure*}
    \centering
    \centering
    \begin{center}
        \Large{Fake (\dallet{} \cite{ramesh2022hierarchical})}\\\vspace{0.25cm}
        \includegraphics[width=\linewidth]{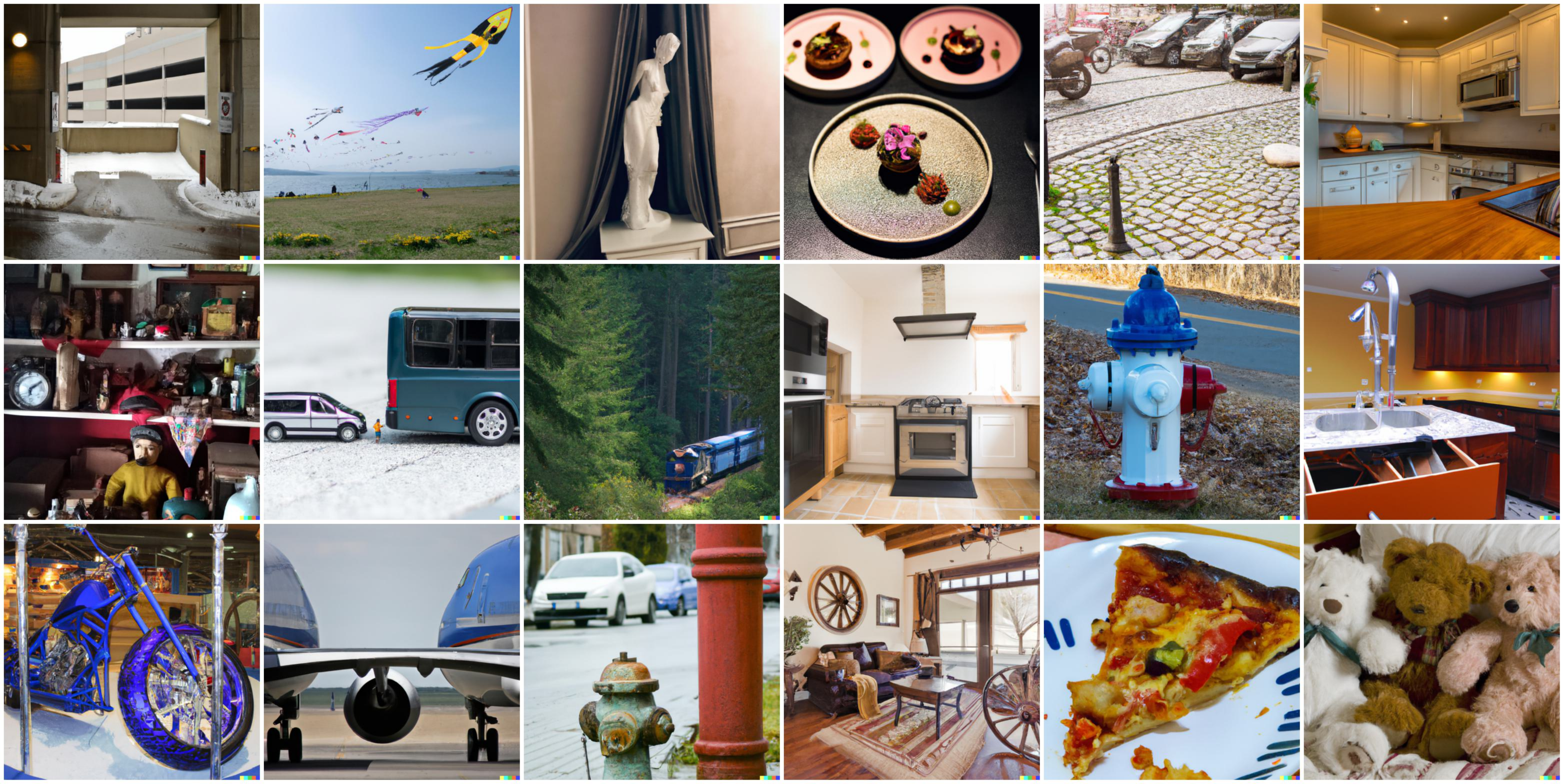}\\\vspace{0.8cm}
        \Large{Real (Reverse Image Search)}\\\vspace{0.25cm}
        \includegraphics[width=\linewidth]{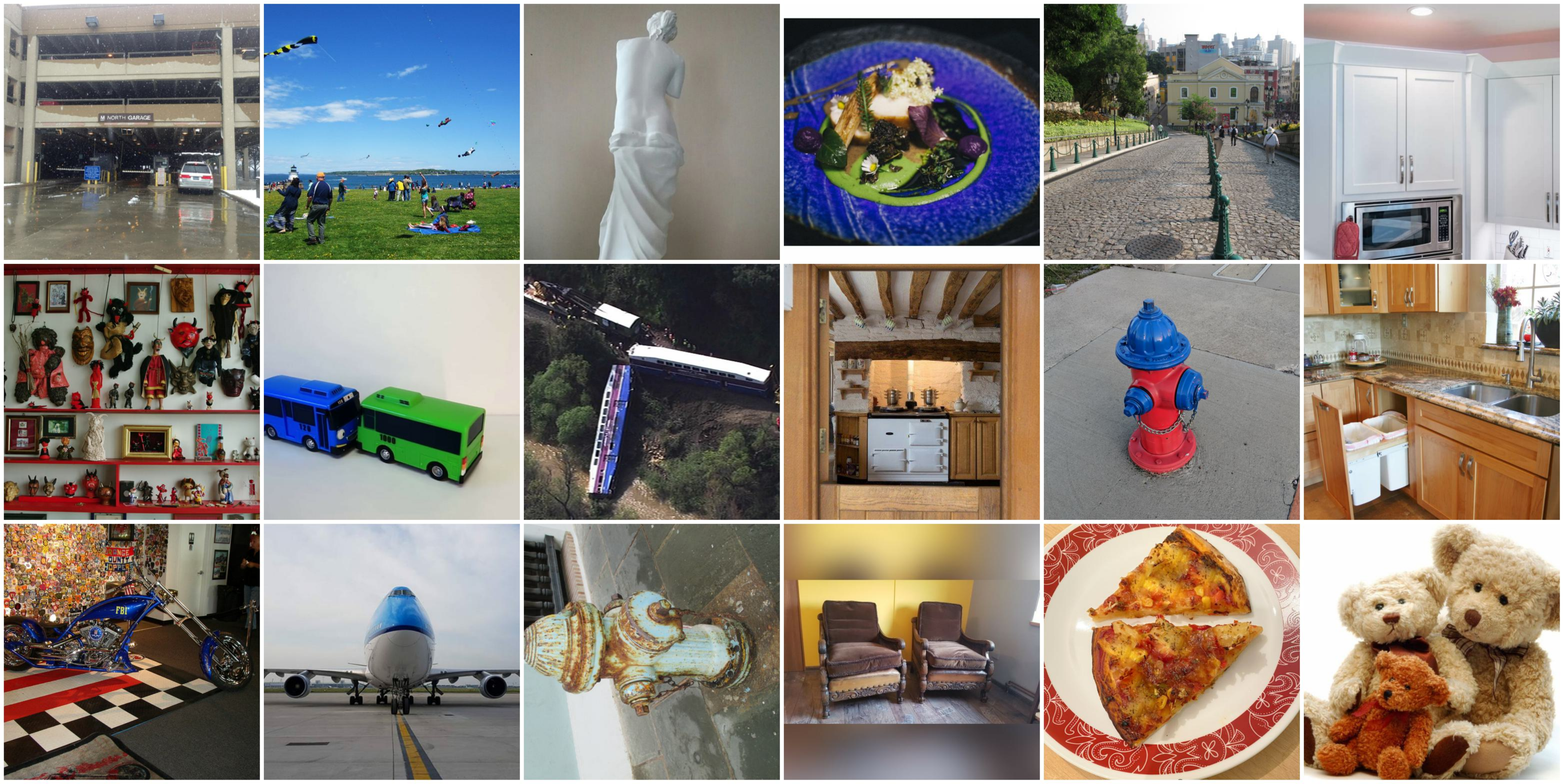}\\
    \end{center}
    \caption{\textbf{Top}: A sample of \textit{fake} \dallet{} \cite{ramesh2022hierarchical} images taken from the DMDetect \cite{corvi2023naples} repo (\href{https://github.com/grip-unina/DMimageDetection}{link}). \textbf{Bot}: Corresponding \textit{real} images found via RIS.}
    \label{fig:dalle2_samples}
\end{figure*}

\begin{figure*}
    \centering
    \centering
    \begin{center}
        \Large{Fake (\dalle{} \cite{dalle3})}\\\vspace{0.25cm}
        \includegraphics[width=\linewidth]{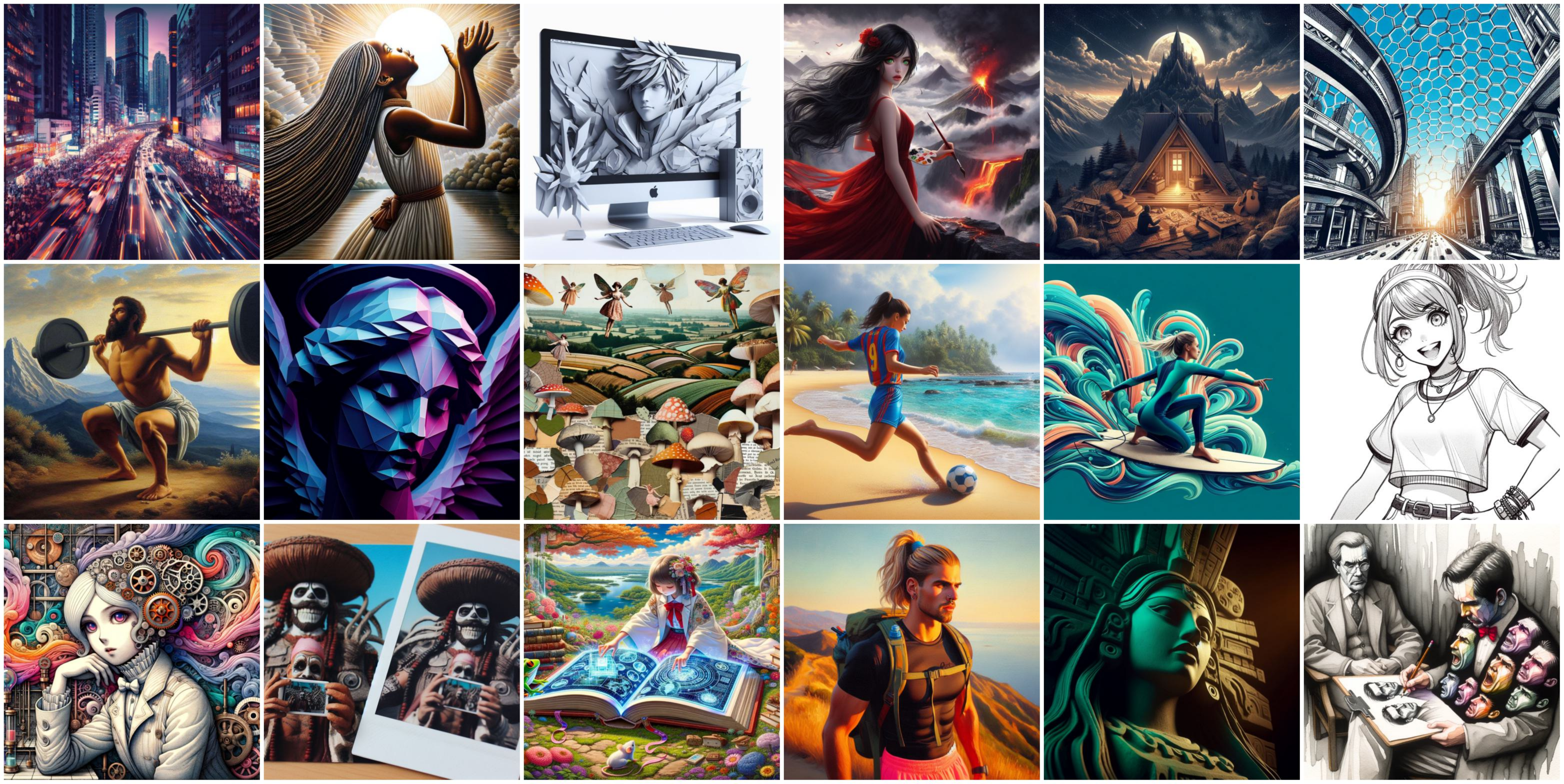}\\\vspace{0.8cm}
        \Large{Real (Reverse Image Search)}\\\vspace{0.25cm}
        \includegraphics[width=\linewidth]{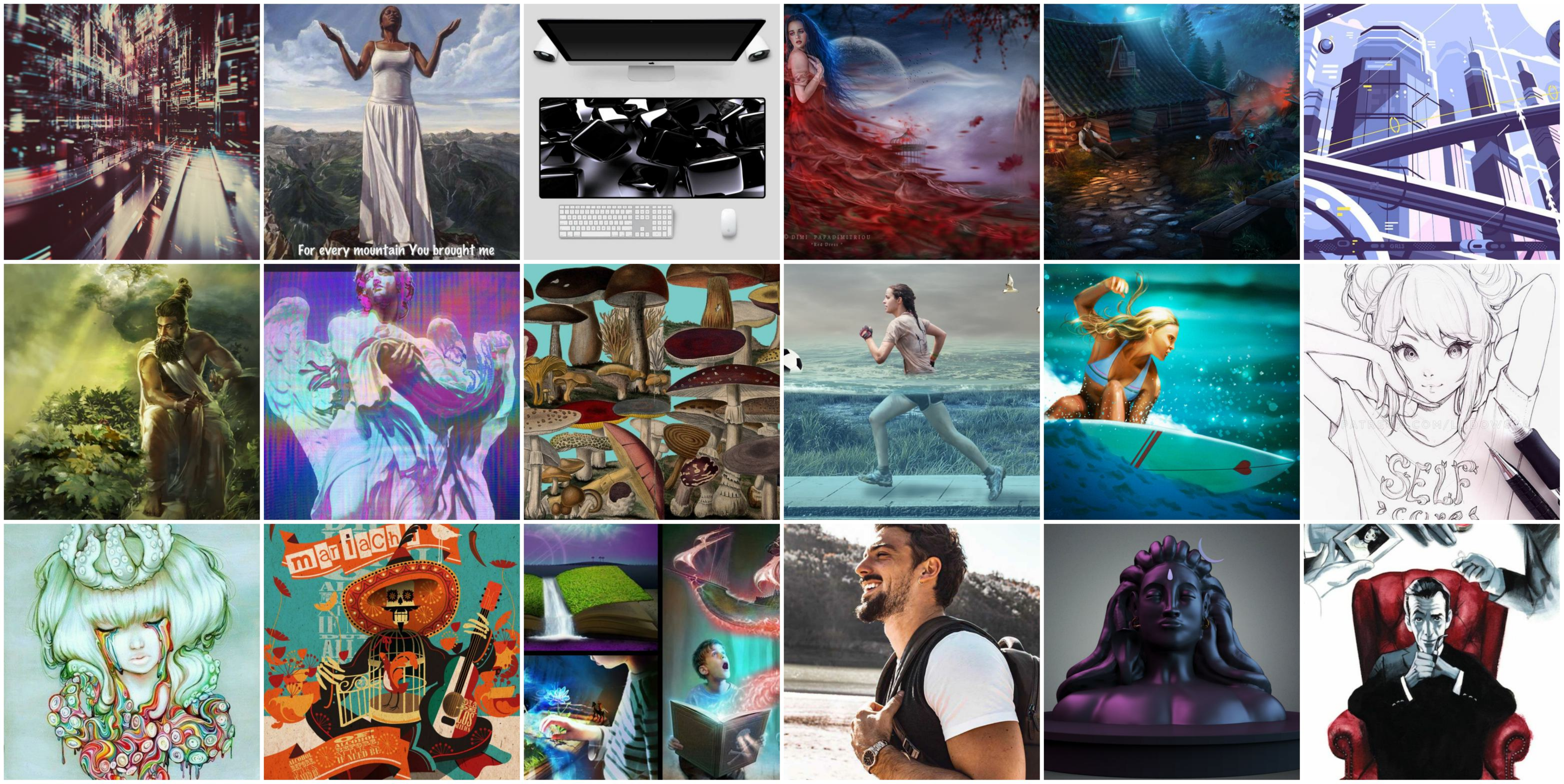}\\
    \end{center}
    \caption{\textbf{Top}: A sample of \textit{fake} \dalle{} \cite{dalle3} images taken from Hugging Face (\href{https://huggingface.co/datasets/laion/dalle-3-dataset}{link}). \textbf{Bot}: Corresponding \textit{real} images found via RIS.}
    \label{fig:dalle3_samples}
\end{figure*}

\begin{figure*}
    \centering
    \centering
    \begin{center}
        \Large{Fake (Kandinsky 2 \cite{kandinsky2_2})}\\\vspace{0.25cm}
        \includegraphics[width=\linewidth]{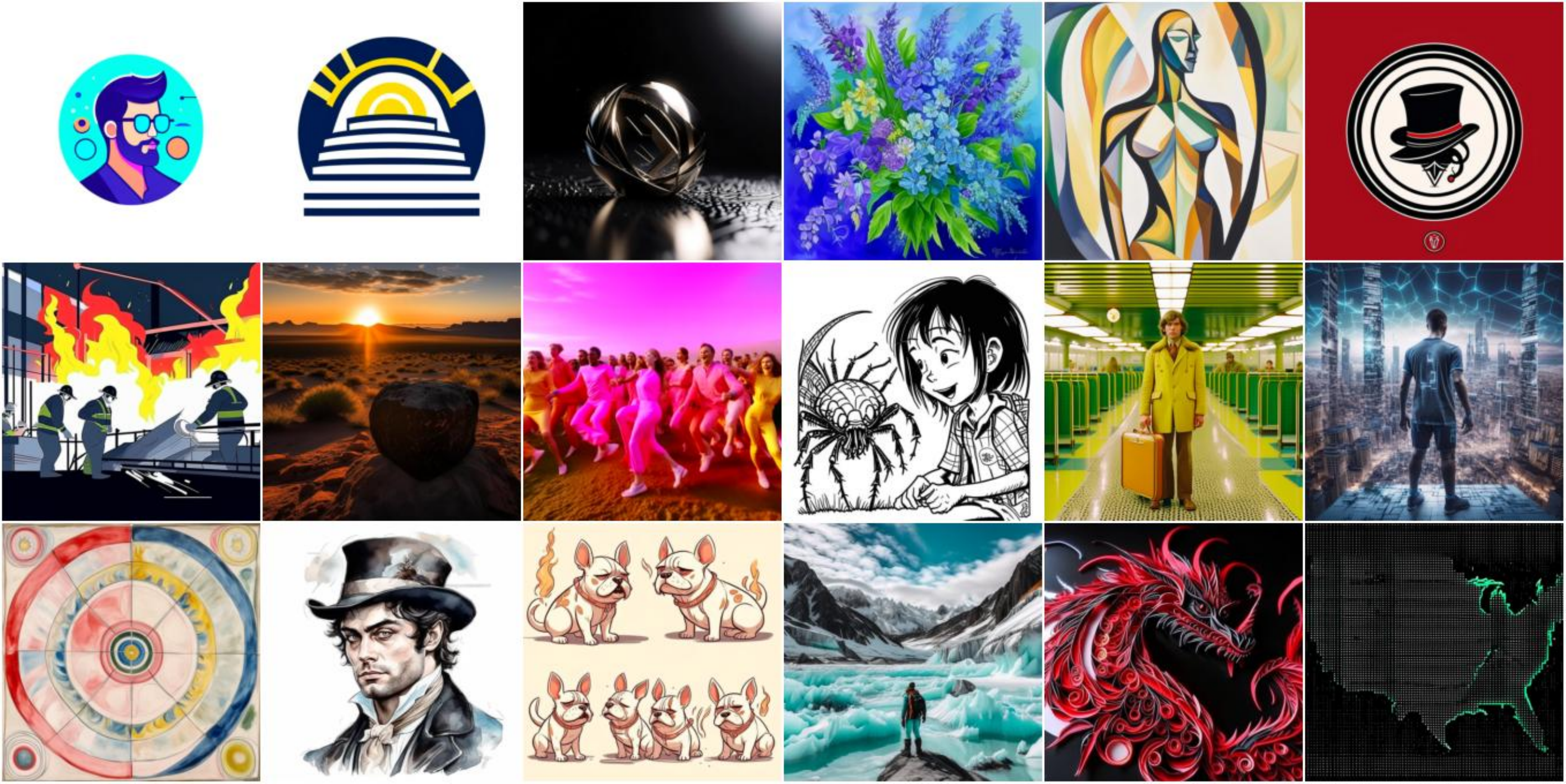}\\\vspace{0.8cm}
        \Large{Real (Reverse Image Search)}\\\vspace{0.25cm}
        \includegraphics[width=\linewidth]{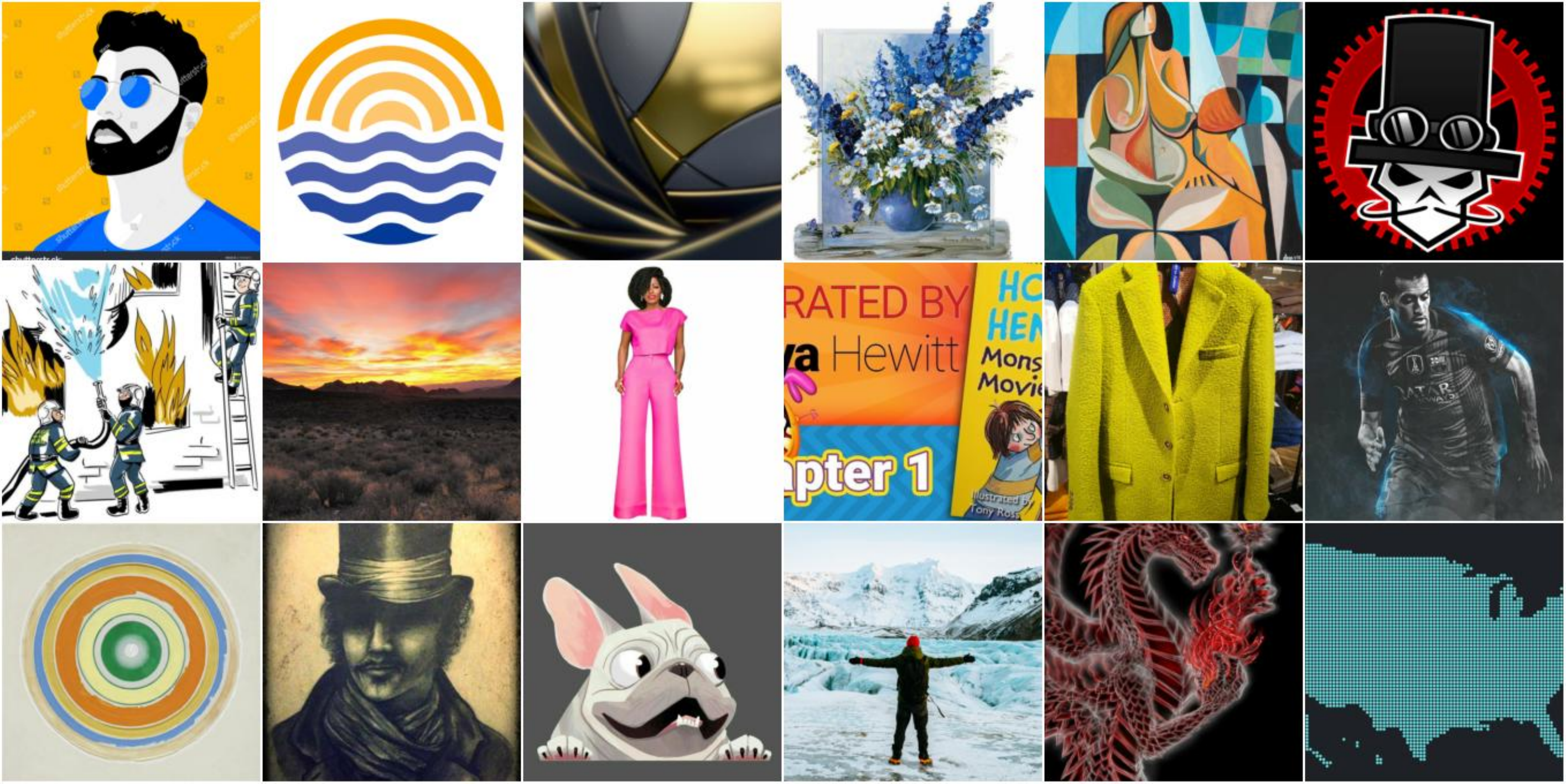}\\
    \end{center}
    \caption{\textbf{Top}: A sample of \textit{fake} Kandinsky 2 \cite{kandinsky2_2} images we generated for our dataset. \textbf{Bot}: Corresponding \textit{real} images found via RIS.}
    \label{fig:kandinsky2_samples}
\end{figure*}

\begin{figure*}
    \centering
    \centering
    \begin{center}
        \Large{Fake (Kandinsky 3 \cite{kandinsky3})}\\\vspace{0.25cm}
        \includegraphics[width=\linewidth]{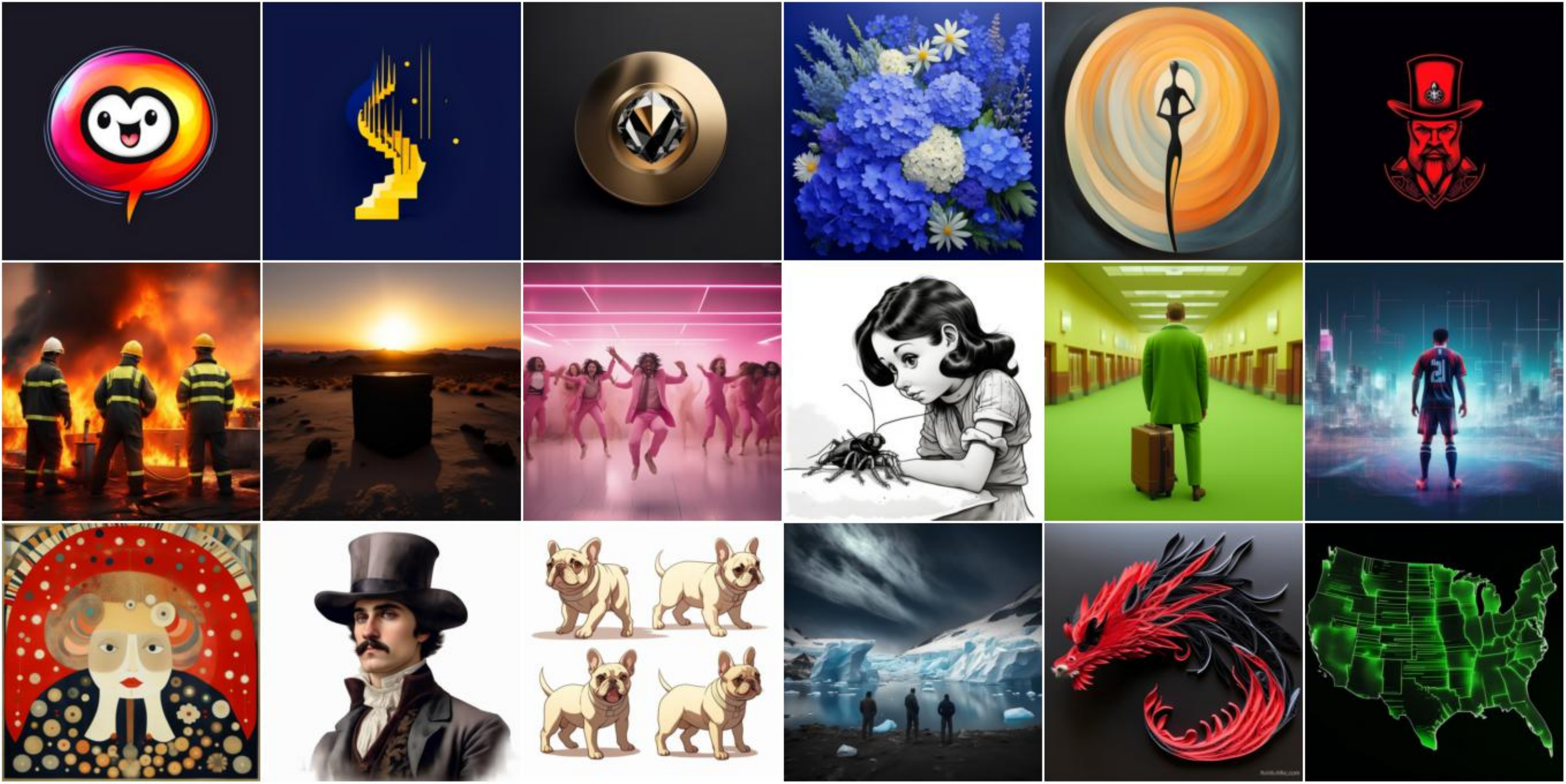}\\\vspace{0.8cm}
        \Large{Real (Reverse Image Search)}\\\vspace{0.25cm}
        \includegraphics[width=\linewidth]{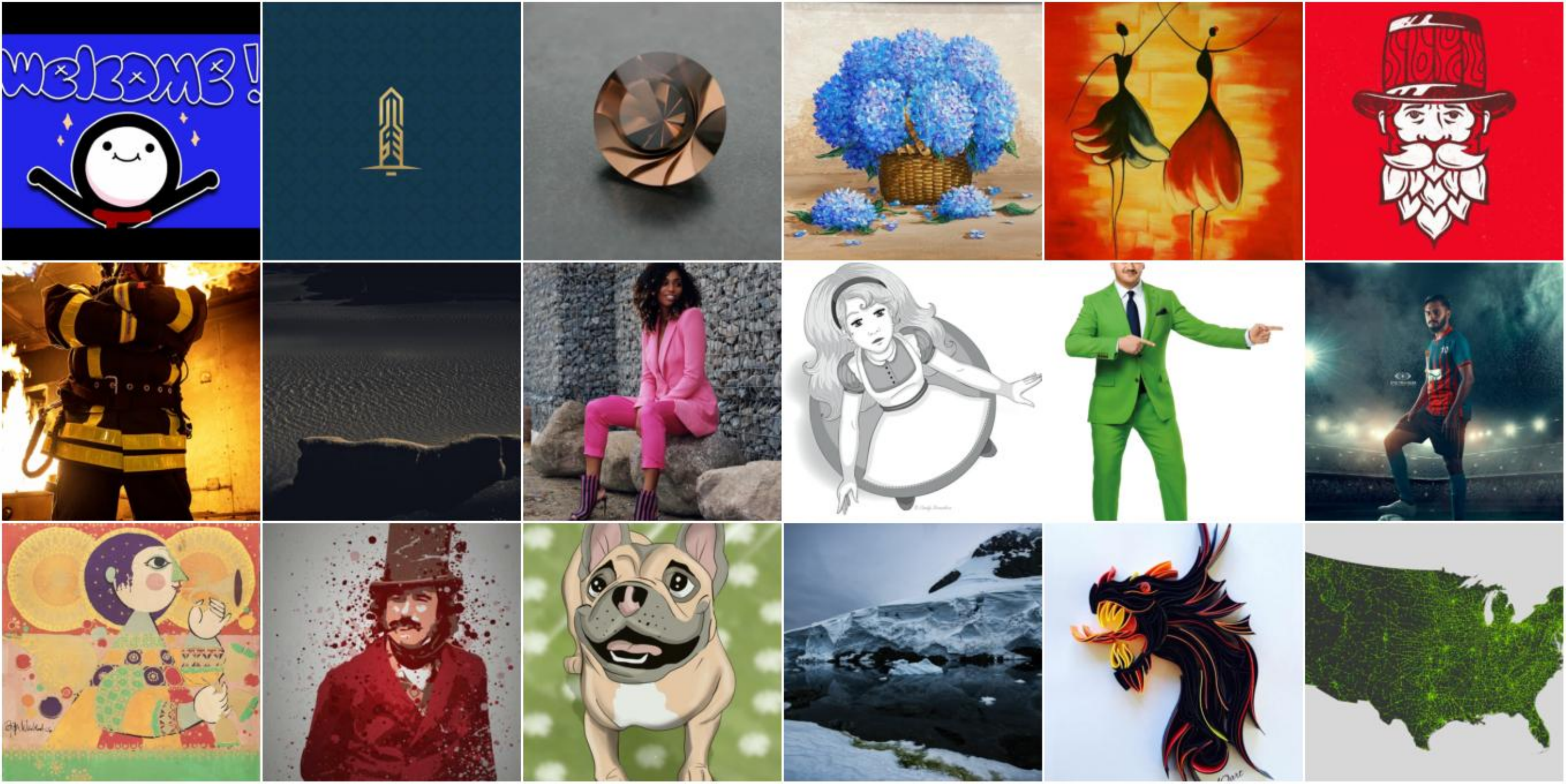}\\
    \end{center}
    \caption{\textbf{Top}: A sample of \textit{fake} Kandinsky 3 \cite{kandinsky3} images we generated for our dataset. \textbf{Bot}: Corresponding \textit{real} images found via RIS.}
    \label{fig:kandinsky3_samples}
\end{figure*}

\begin{figure*}
    \centering
    \centering
    \begin{center}
        \Large{Fake (PixArt-$\alpha$ \cite{chen2023pixart})}\\\vspace{0.25cm}
        \includegraphics[width=\linewidth]{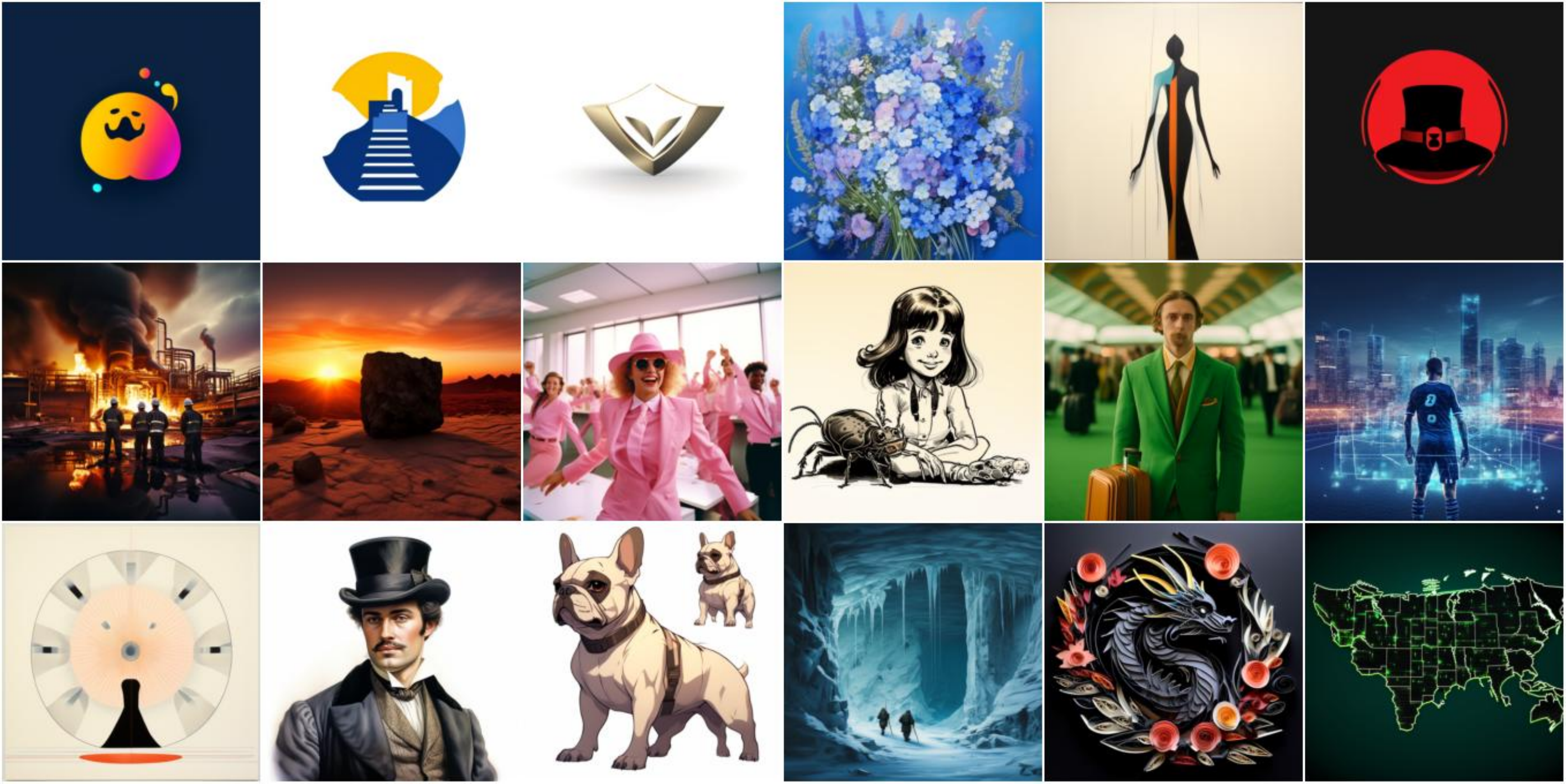}\\\vspace{0.8cm}
        \Large{Real (Reverse Image Search)}\\\vspace{0.25cm}
        \includegraphics[width=\linewidth]{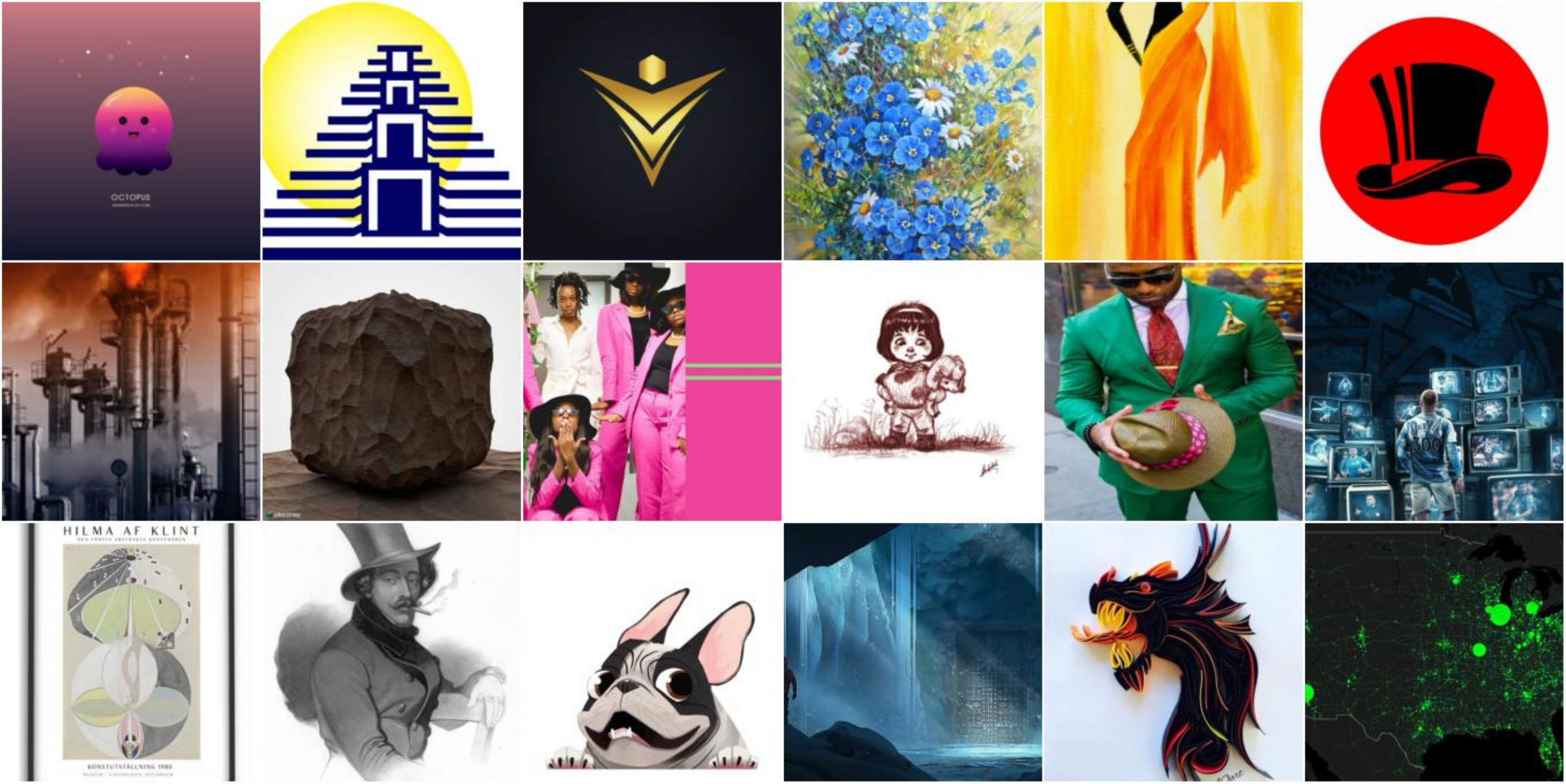}\\
    \end{center}
    \caption{\textbf{Top}: A sample of \textit{fake} PixArt-$\alpha$ \cite{chen2023pixart} images we generated for our dataset. \textbf{Bot}: Corresponding \textit{real} images found via RIS.}
    \label{fig:pixart_samples}
\end{figure*}

\begin{figure*}
    \centering
    \centering
    \begin{center}
        \Large{Fake (Playground 2.5 \cite{li2024playground})}\\\vspace{0.25cm}
        \includegraphics[width=\linewidth]{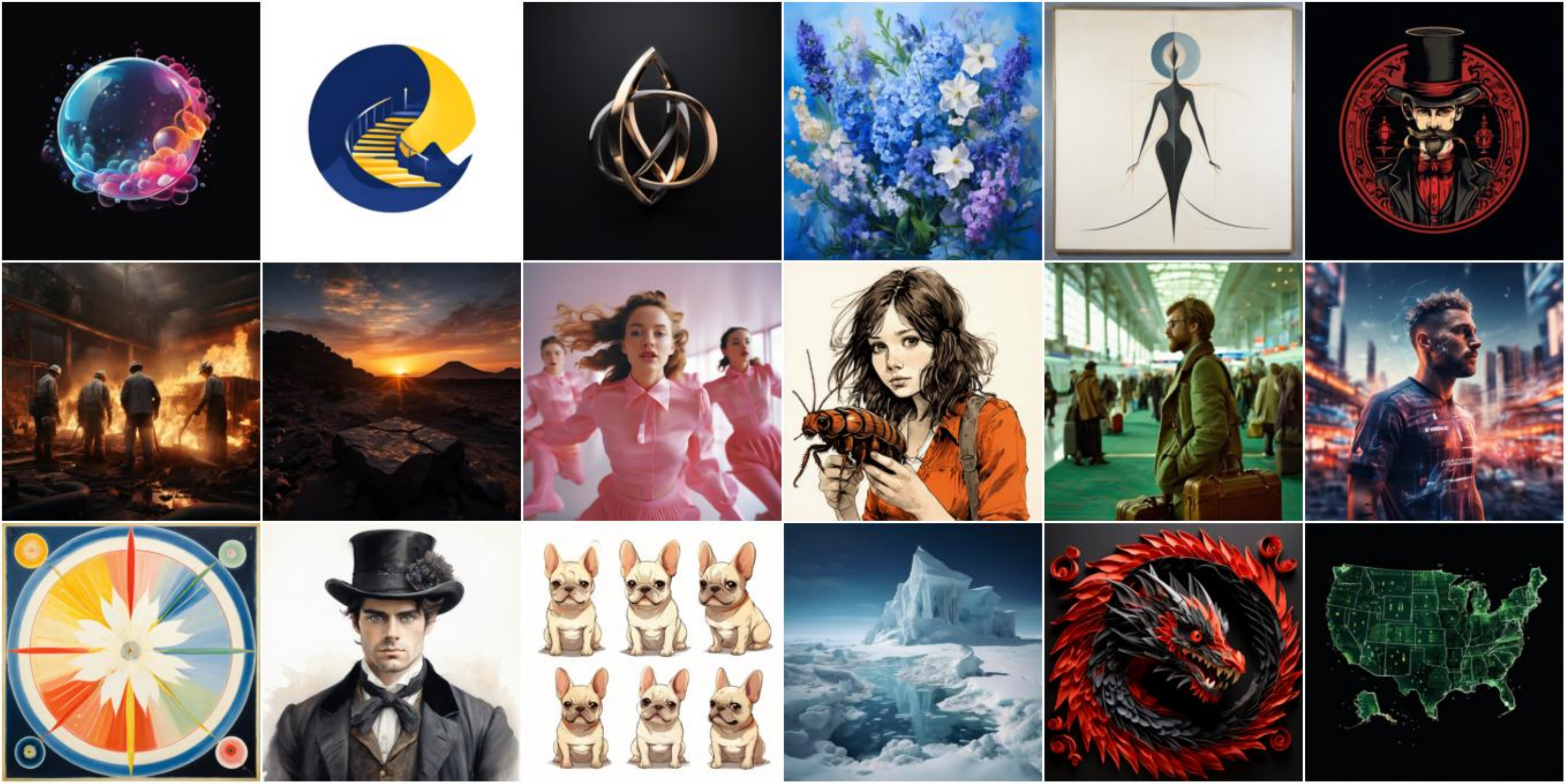}\\\vspace{0.8cm}
        \Large{Real (Reverse Image Search)}\\\vspace{0.25cm}
        \includegraphics[width=\linewidth]{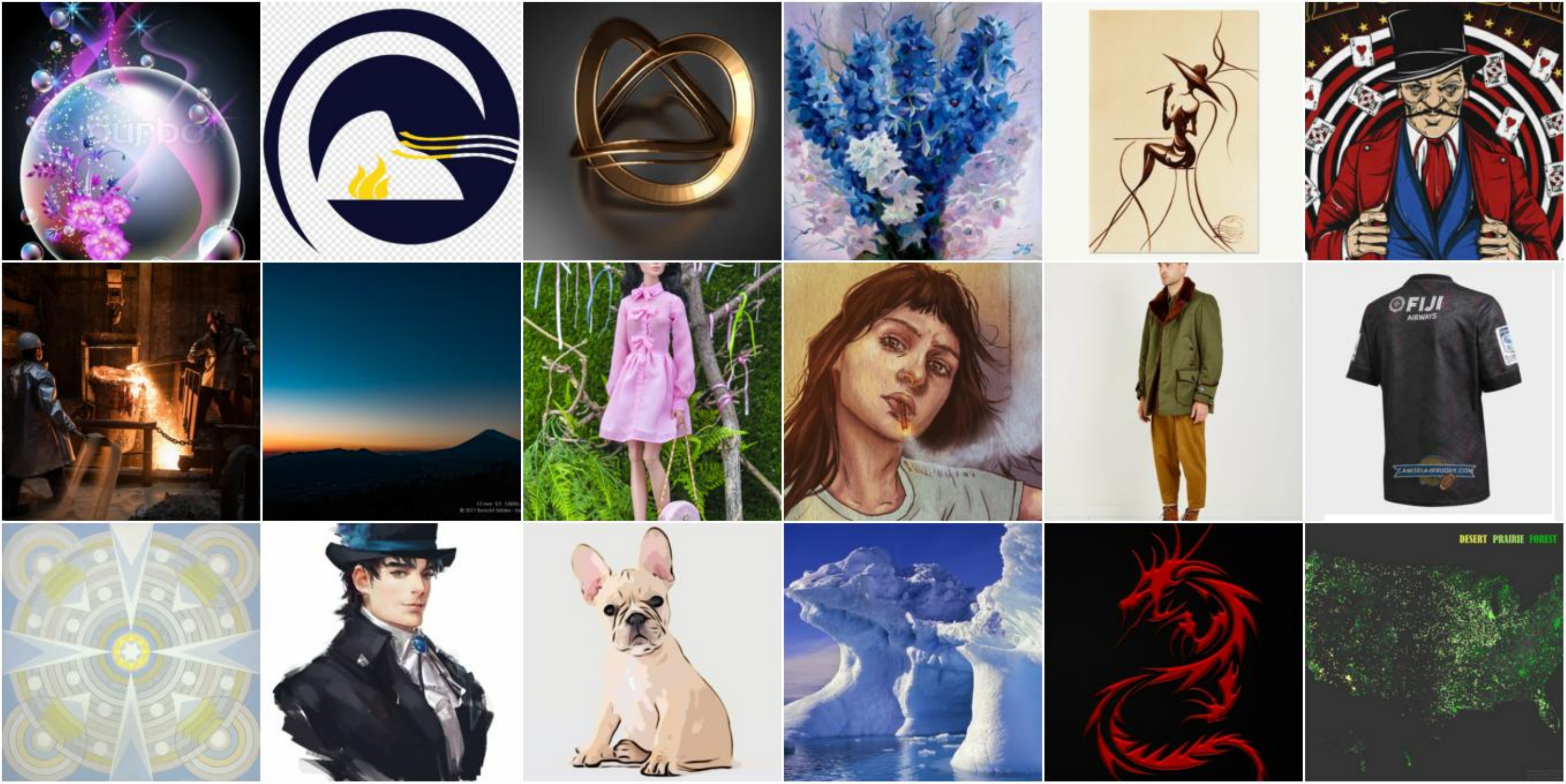}\\
    \end{center}
    \caption{\textbf{Top}: A sample of \textit{fake} Playground \cite{li2024playground} images we generated for our dataset. \textbf{Bot}: Corresponding \textit{real} images found via RIS.}
    \label{fig:playground_samples}
\end{figure*}

\begin{figure*}
    \centering
    \centering
    \begin{center}
        \Large{Fake (SDXL-DPO \cite{dpo})}\\\vspace{0.25cm}
        \includegraphics[width=\linewidth]{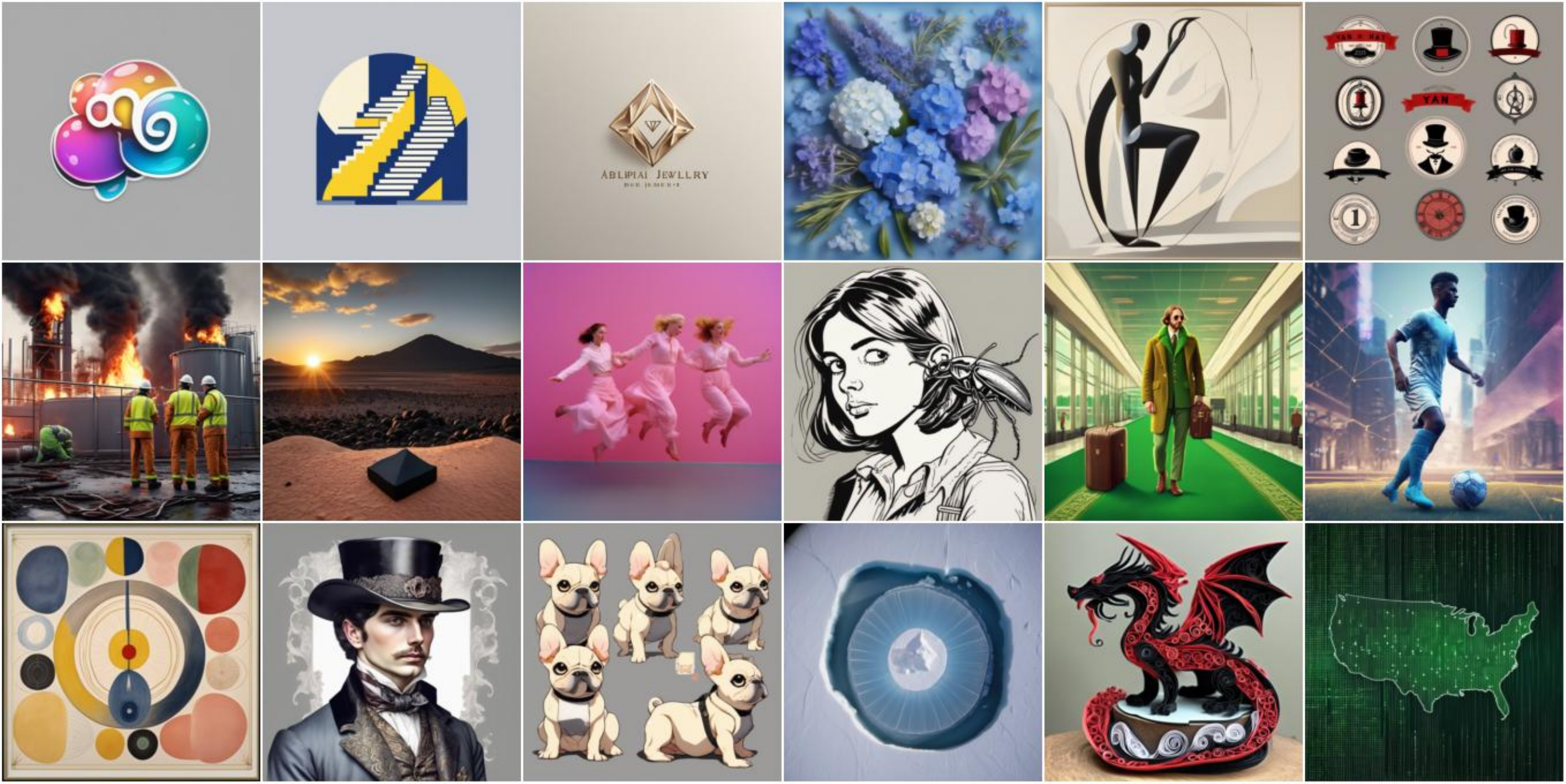}\\\vspace{0.8cm}
        \Large{Real (Reverse Image Search)}\\\vspace{0.25cm}
        \includegraphics[width=\linewidth]{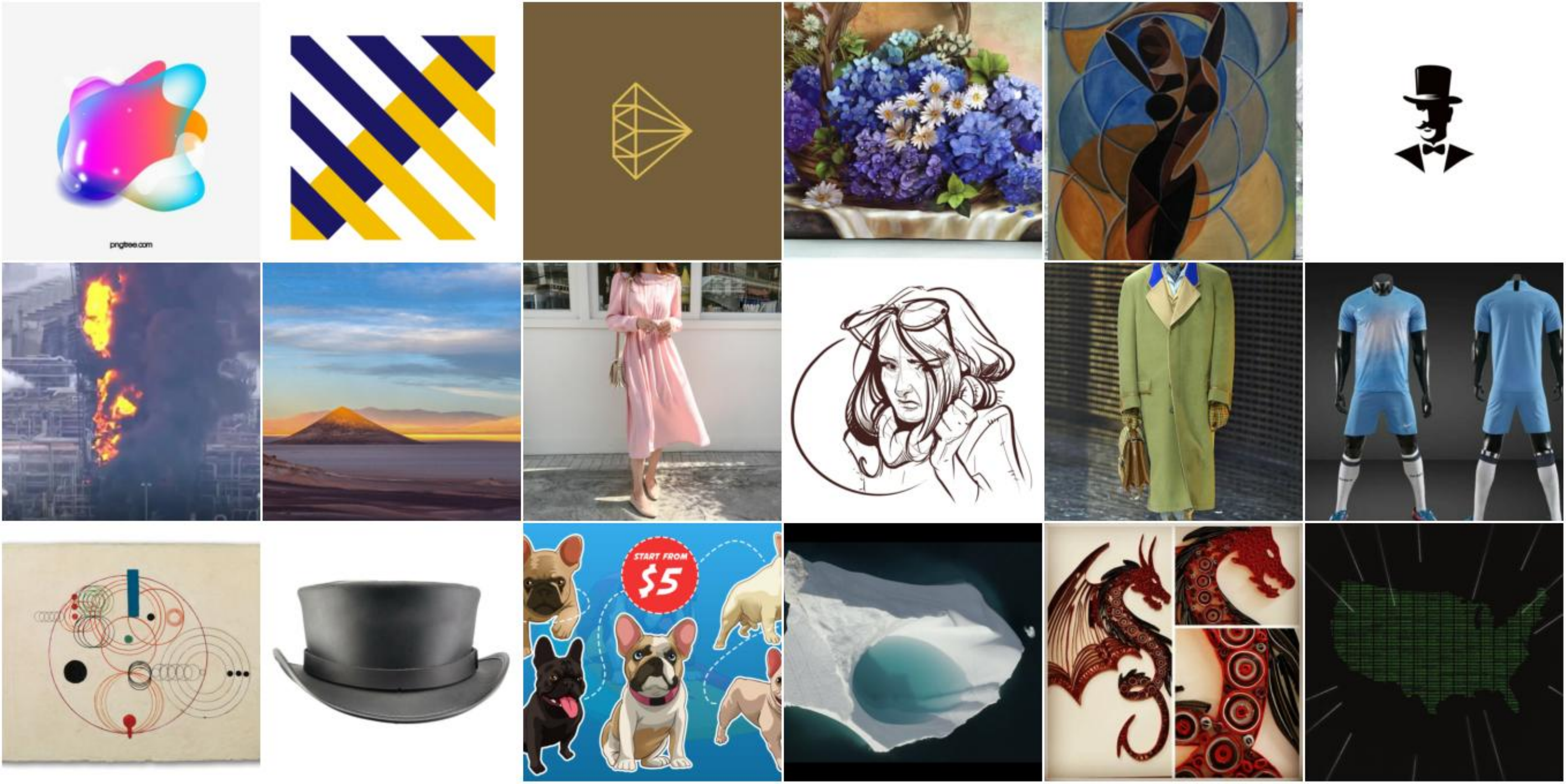}\\
    \end{center}
    \caption{\textbf{Top}: A sample of \textit{fake} SDXL-DPO \cite{dpo} images we generated for our dataset. \textbf{Bot}: Corresponding \textit{real} images found via RIS.}
    \label{fig:dpo_samples}
\end{figure*}

\begin{figure*}
    \centering
    \centering
    \begin{center}
        \Large{Fake (SDXL \cite{sdxl})}\\\vspace{0.25cm}
        \includegraphics[width=\linewidth]{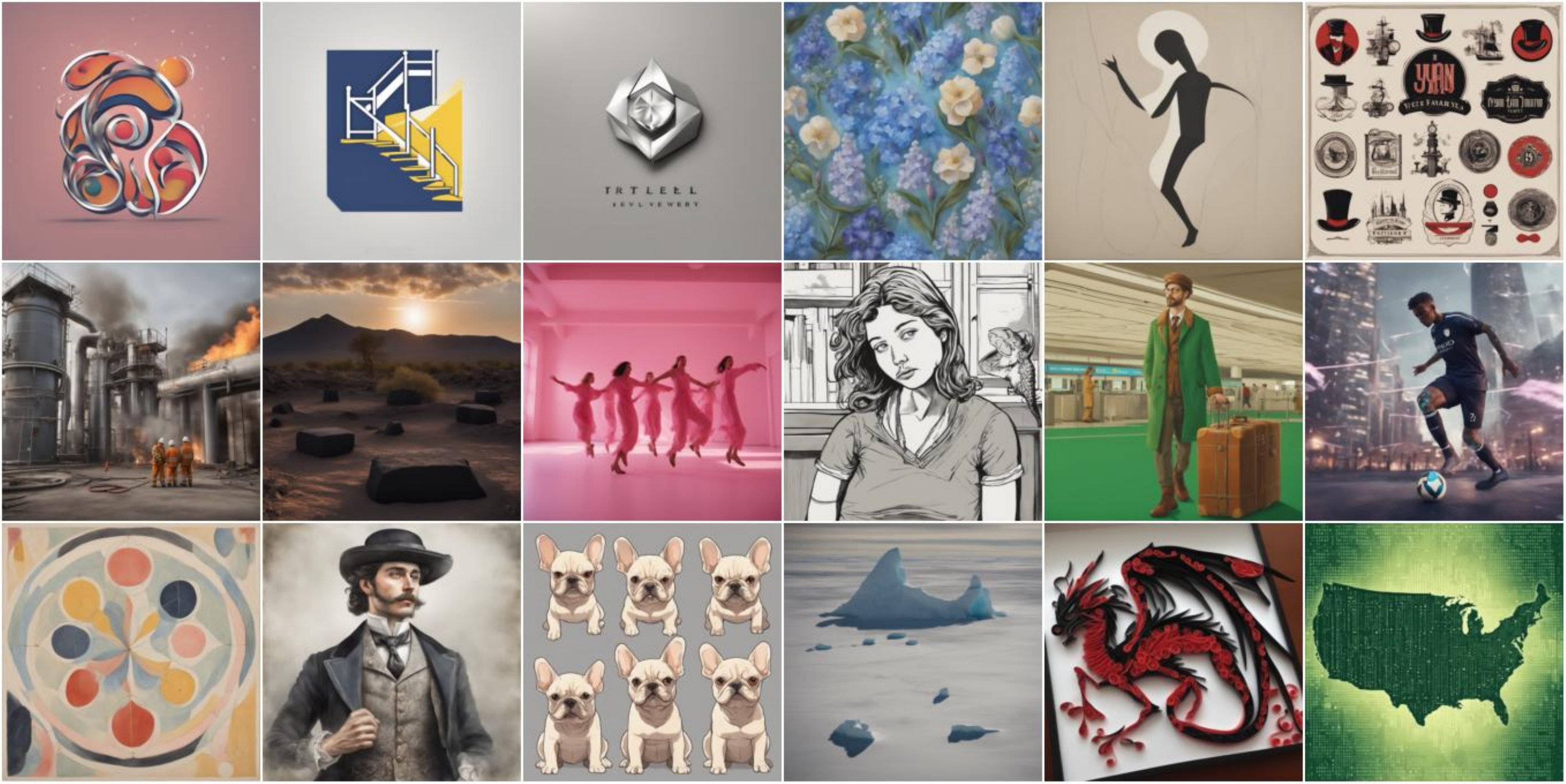}\\\vspace{0.8cm}
        \Large{Real (Reverse Image Search)}\\\vspace{0.25cm}
        \includegraphics[width=\linewidth]{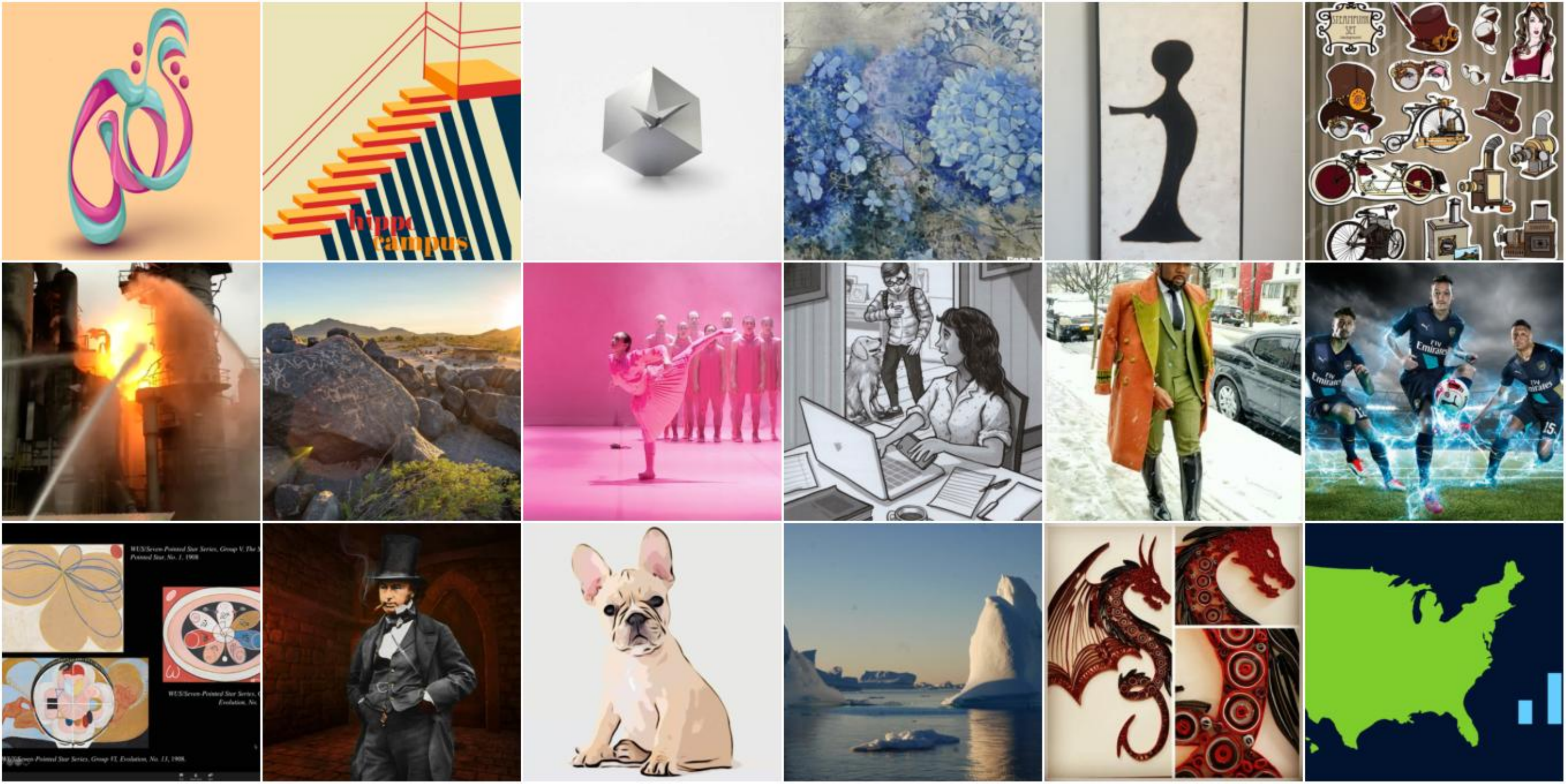}\\
    \end{center}
    \caption{\textbf{Top}: A sample of \textit{fake} SDXL \cite{sdxl} images we generated for our dataset. \textbf{Bot}: Corresponding \textit{real} images found via RIS.}
    \label{fig:sdxl_samples}
\end{figure*}

\begin{figure*}
    \centering
    \centering
    \begin{center}
        \Large{Fake (Seg-MoE \cite{segmoe})}\\\vspace{0.25cm}
        \includegraphics[width=\linewidth]{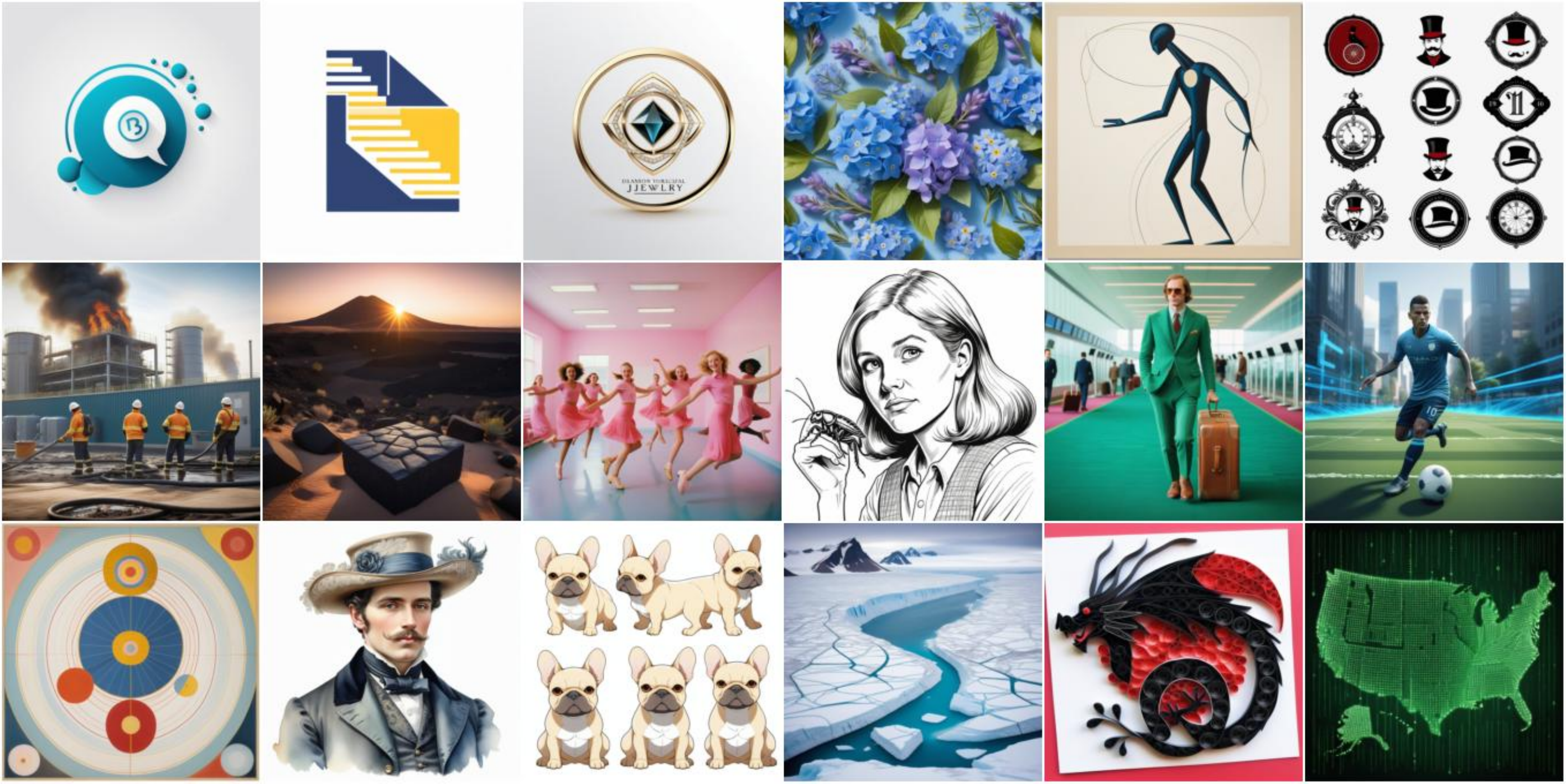}\\\vspace{0.8cm}
        \Large{Real (Reverse Image Search)}\\\vspace{0.25cm}
        \includegraphics[width=\linewidth]{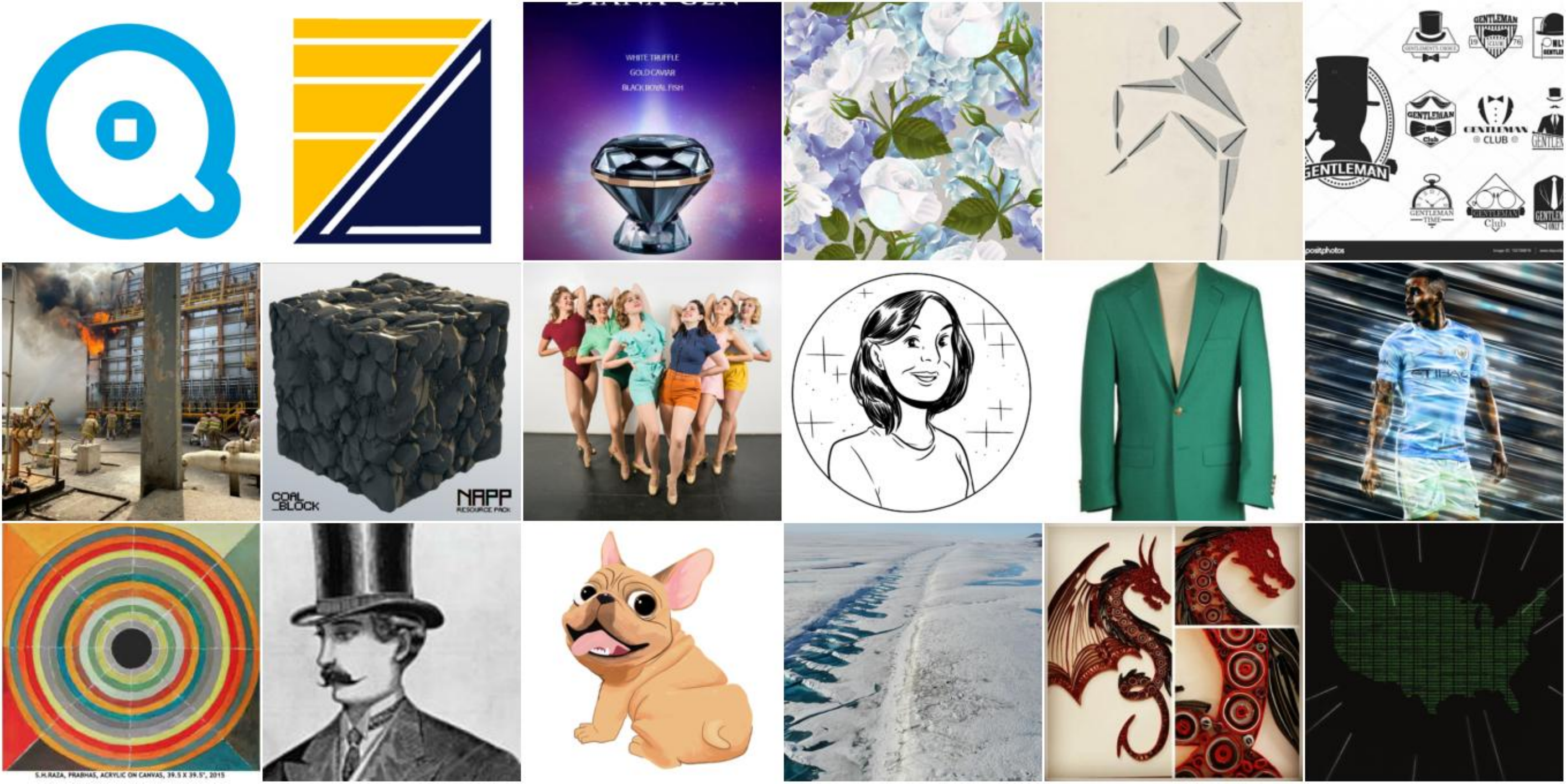}\\
    \end{center}
    \caption{\textbf{Top}: A sample of \textit{fake} Seg-MoE \cite{segmoe} images we generated for our dataset. \textbf{Bot}: Corresponding \textit{real} images found via RIS.}
    \label{fig:segmoe_samples}
\end{figure*}

\begin{figure*}
    \centering
    \centering
    \begin{center}
        \Large{Fake (SSD-1B \cite{segmind})}\\\vspace{0.25cm}
        \includegraphics[width=\linewidth]{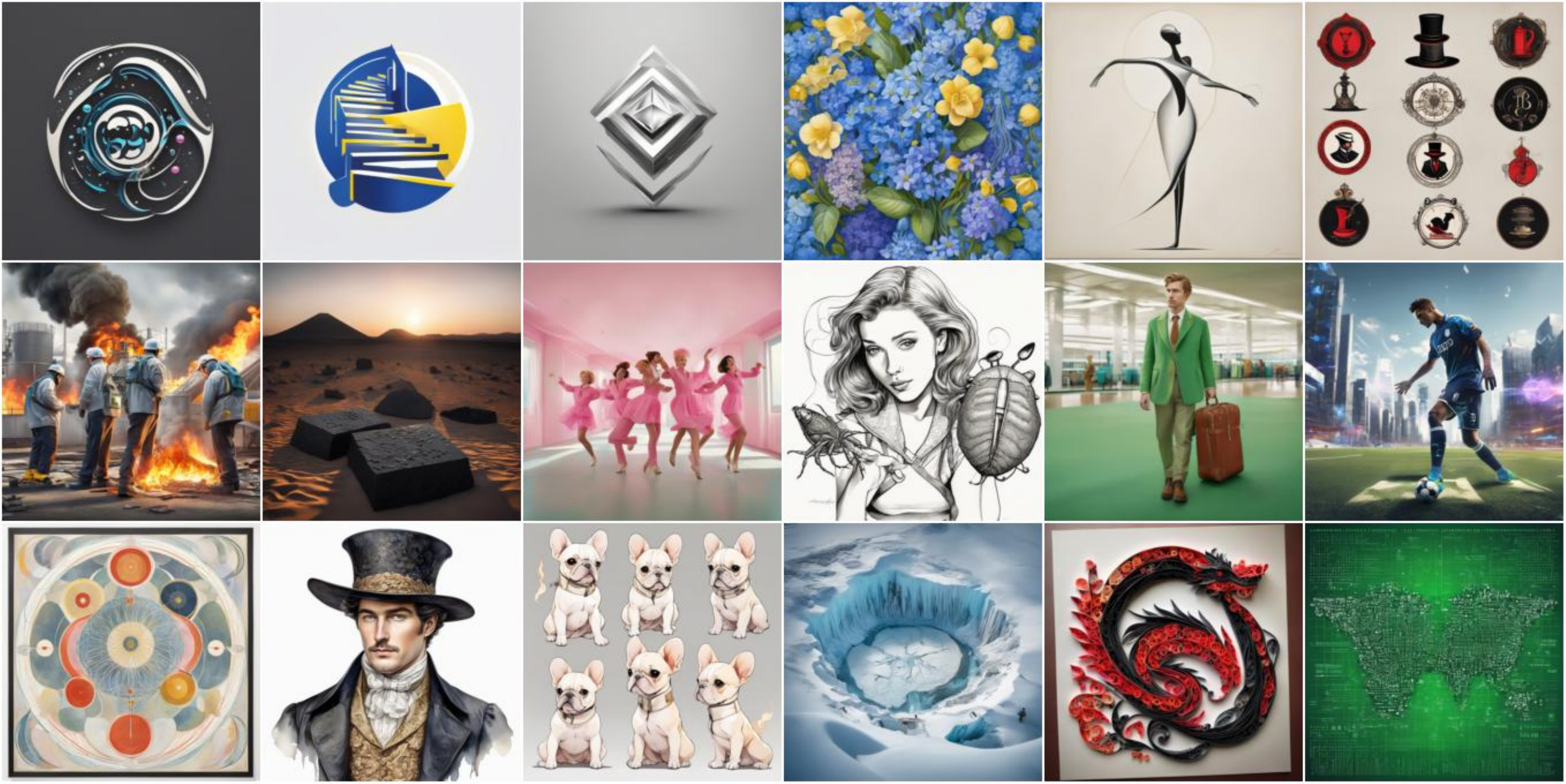}\\\vspace{0.8cm}
        \Large{Real (Reverse Image Search)}\\\vspace{0.25cm}
        \includegraphics[width=\linewidth]{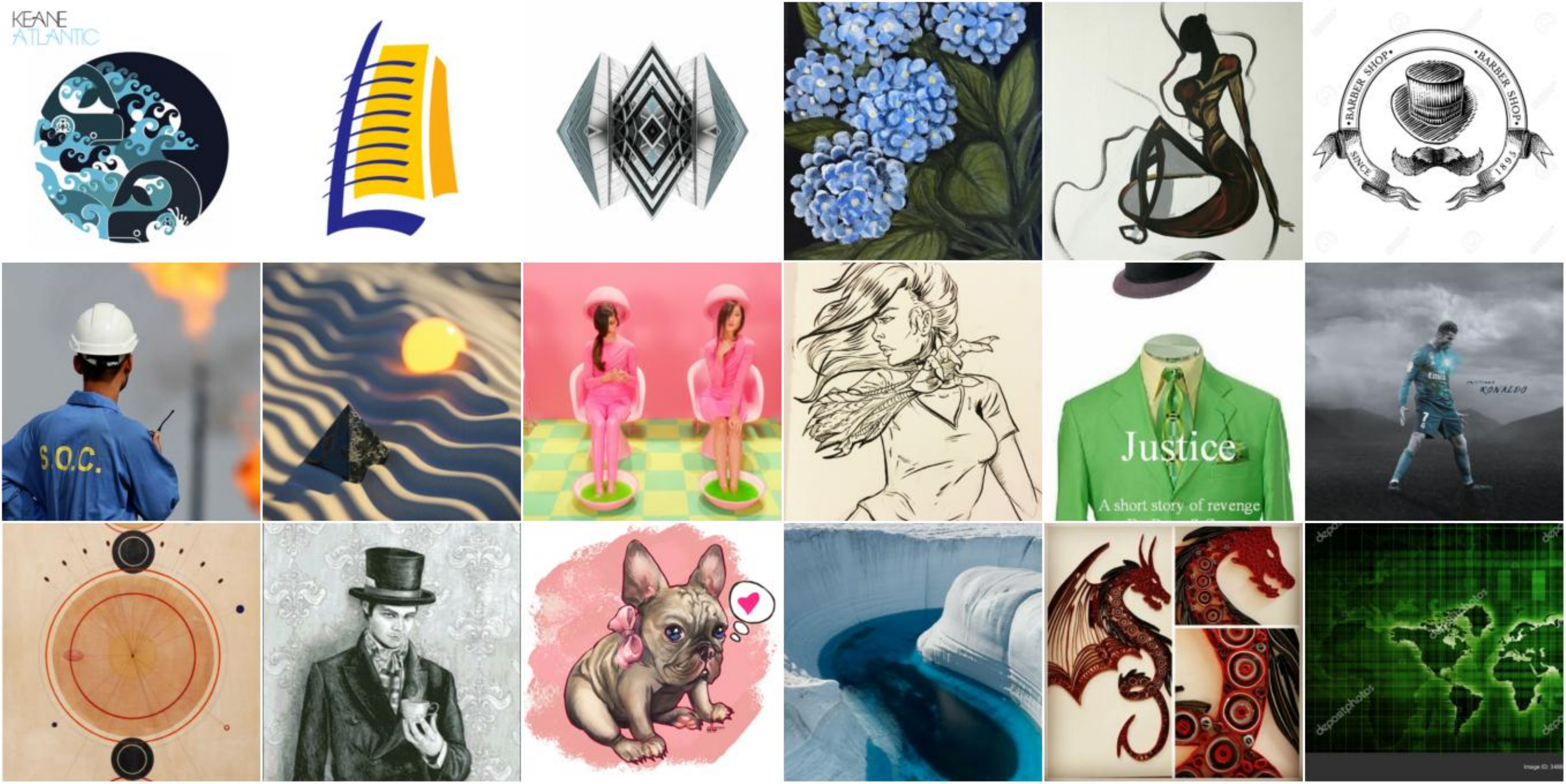}\\
    \end{center}
    \caption{\textbf{Top}: A sample of \textit{fake} SSD-1B \cite{segmind} images we generated for our dataset. \textbf{Bot}: Corresponding \textit{real} images found via RIS.}
    \label{fig:ssd1b_samples}
\end{figure*}

\begin{figure*}
    \centering
    \centering
    \begin{center}
        \Large{Fake (Stable-Cascade \cite{wurstchen})}\\\vspace{0.25cm}
        \includegraphics[width=\linewidth]{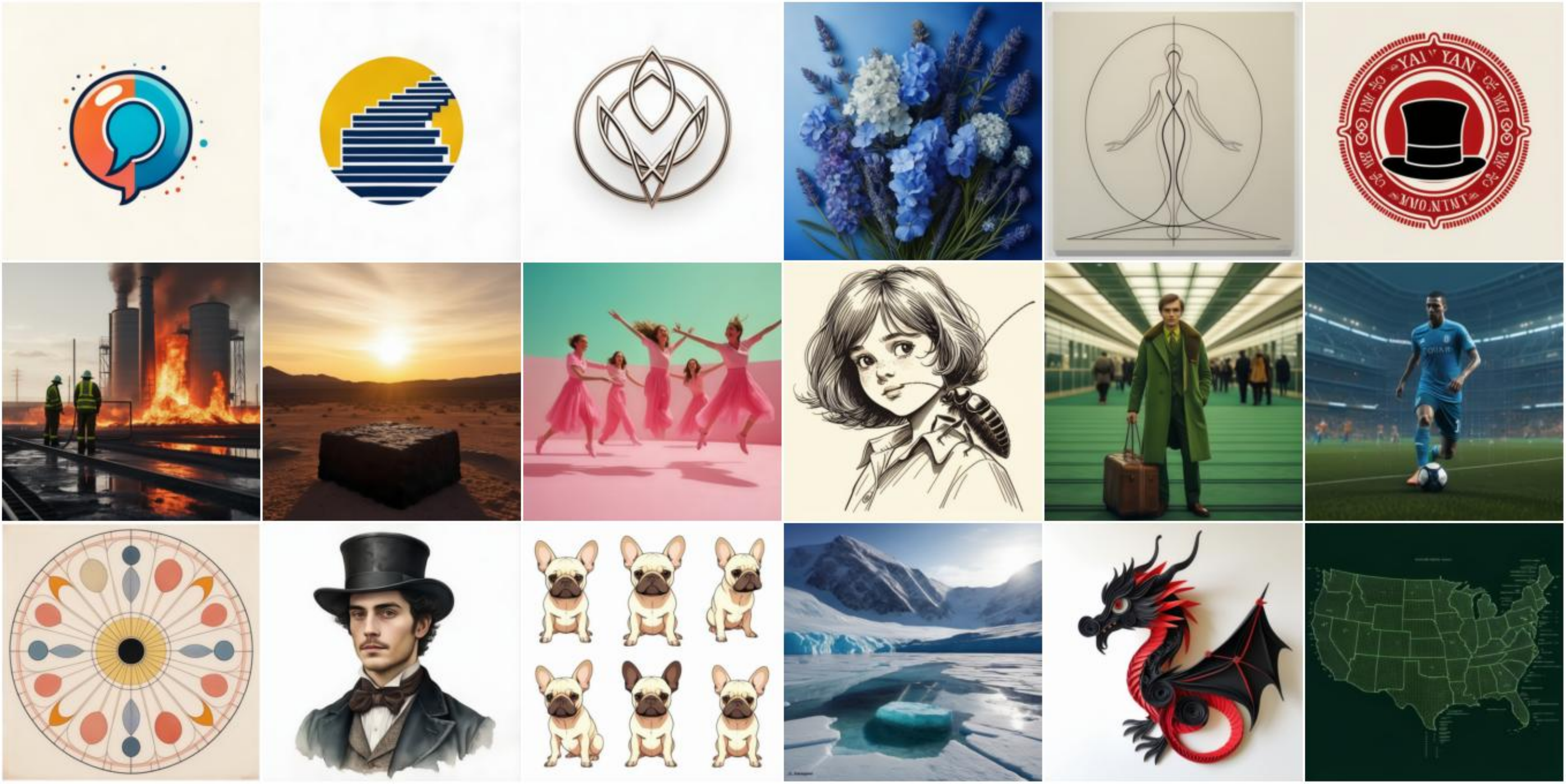}\\\vspace{0.8cm}
        \Large{Real (Reverse Image Search)}\\\vspace{0.25cm}
        \includegraphics[width=\linewidth]{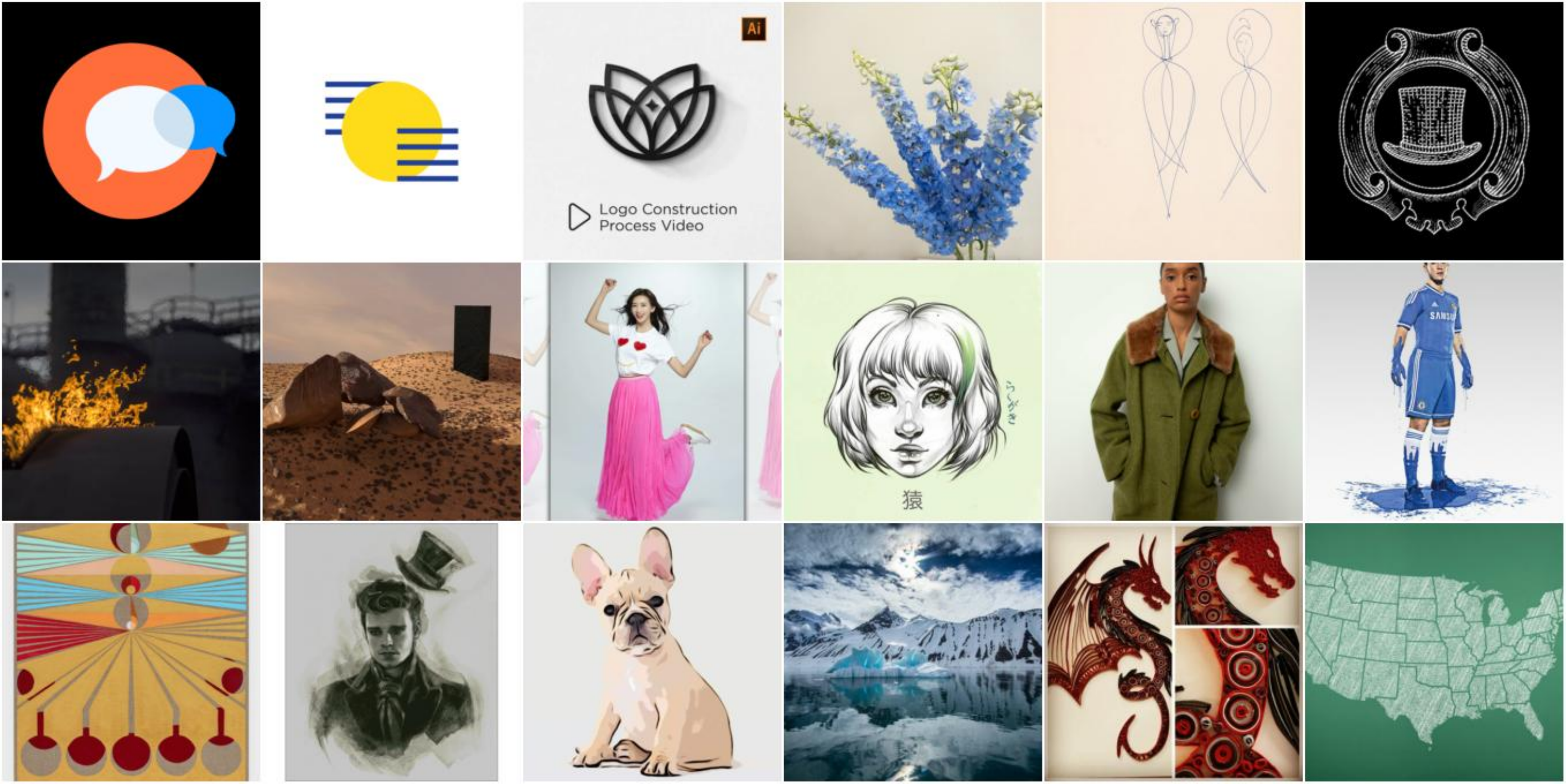}\\
    \end{center}
    \caption{\textbf{Top}: A sample of \textit{fake} Stable-Cascade \cite{wurstchen} images we generated for our dataset. \textbf{Bot}: Corresponding \textit{real} images found via RIS.}
    \label{fig:cascade_samples}
\end{figure*}

\begin{figure*}
    \centering
    \centering
    \begin{center}
        \Large{Fake (Segmind Vega \cite{segmind})}\\\vspace{0.25cm}
        \includegraphics[width=\linewidth]{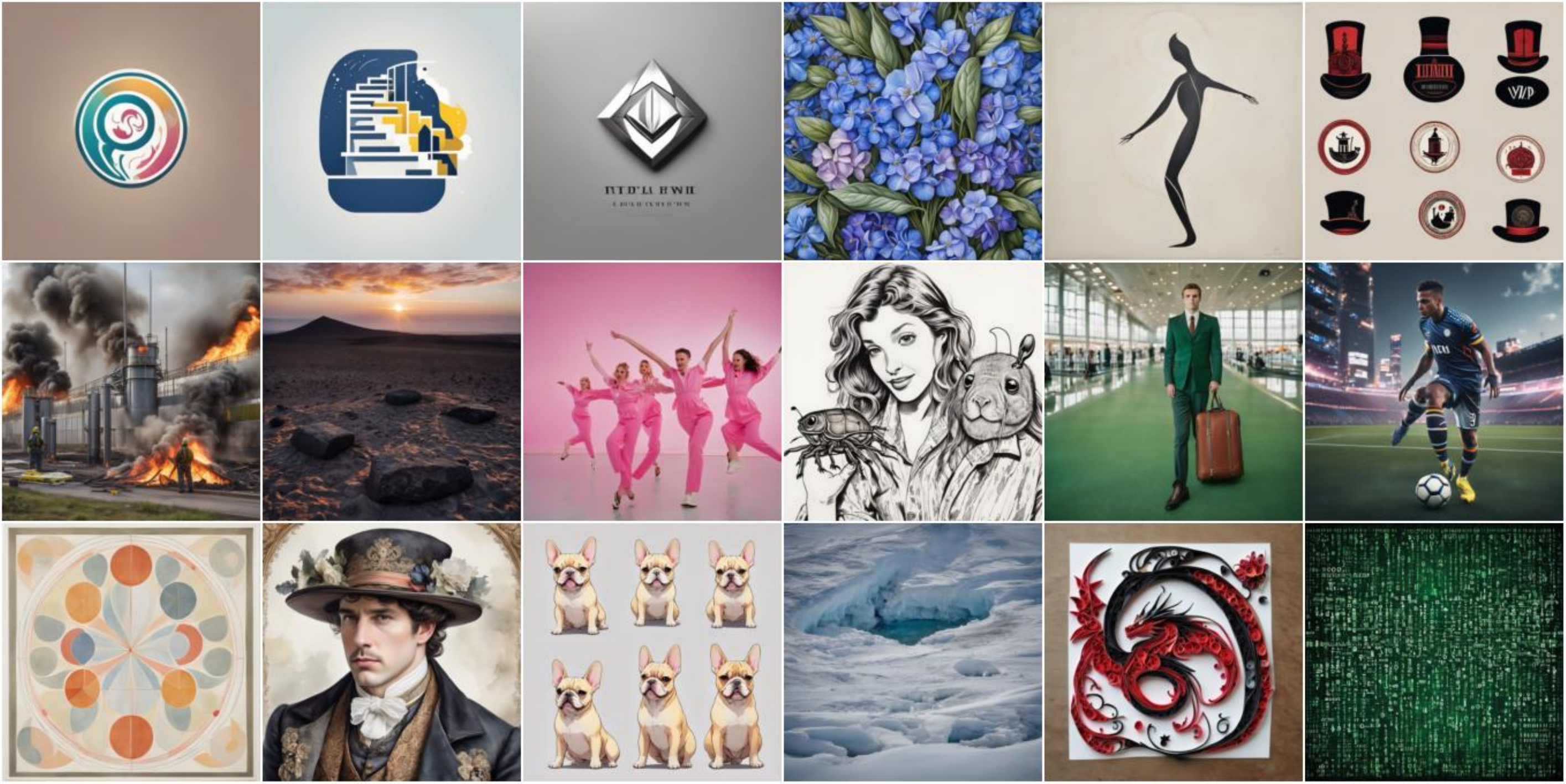}\\\vspace{0.8cm}
        \Large{Real (Reverse Image Search)}\\\vspace{0.25cm}
        \includegraphics[width=\linewidth]{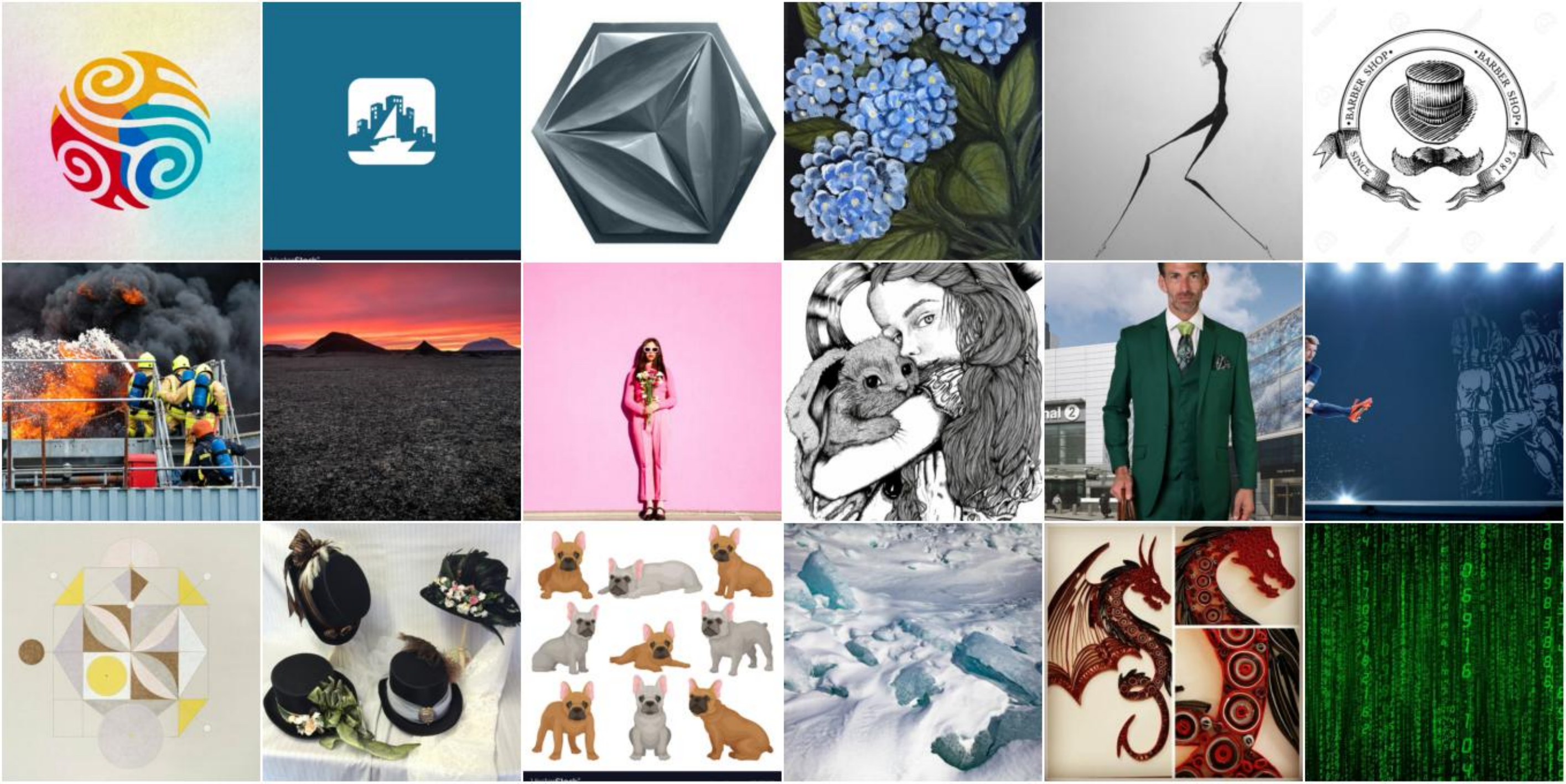}\\
    \end{center}
    \caption{\textbf{Top}: A sample of \textit{fake} Segmind Vega \cite{segmind} images we generated for our dataset. \textbf{Bot}: Corresponding \textit{real} images found via RIS.}
    \label{fig:vega_samples}
\end{figure*}

\begin{figure*}
    \centering
    \centering
    \begin{center}
        \Large{Fake (W\"urstchen 2 \cite{wurstchen})}\\\vspace{0.25cm}
        \includegraphics[width=\linewidth]{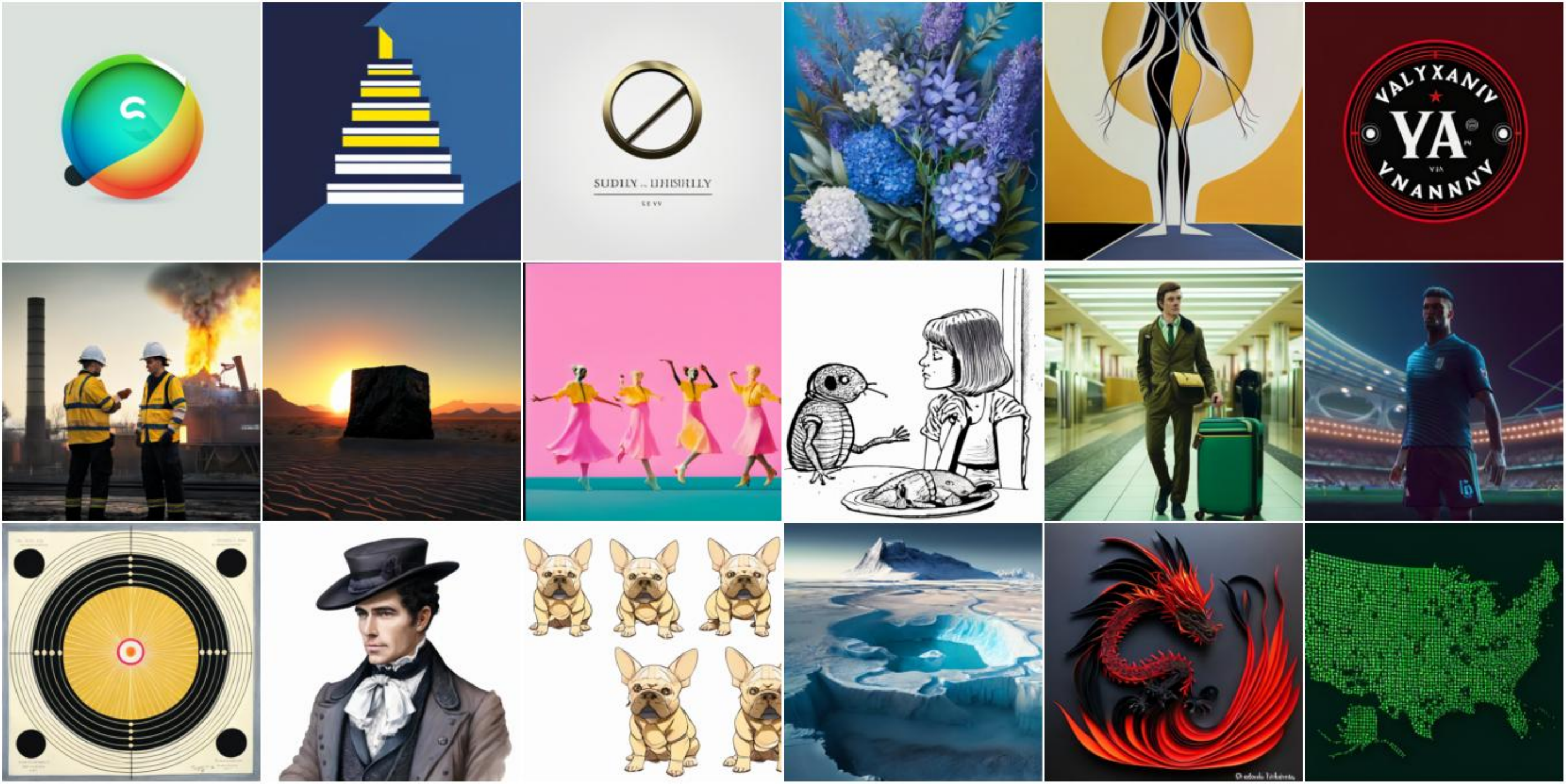}\\\vspace{0.8cm}
        \Large{Real (Reverse Image Search)}\\\vspace{0.25cm}
        \includegraphics[width=\linewidth]{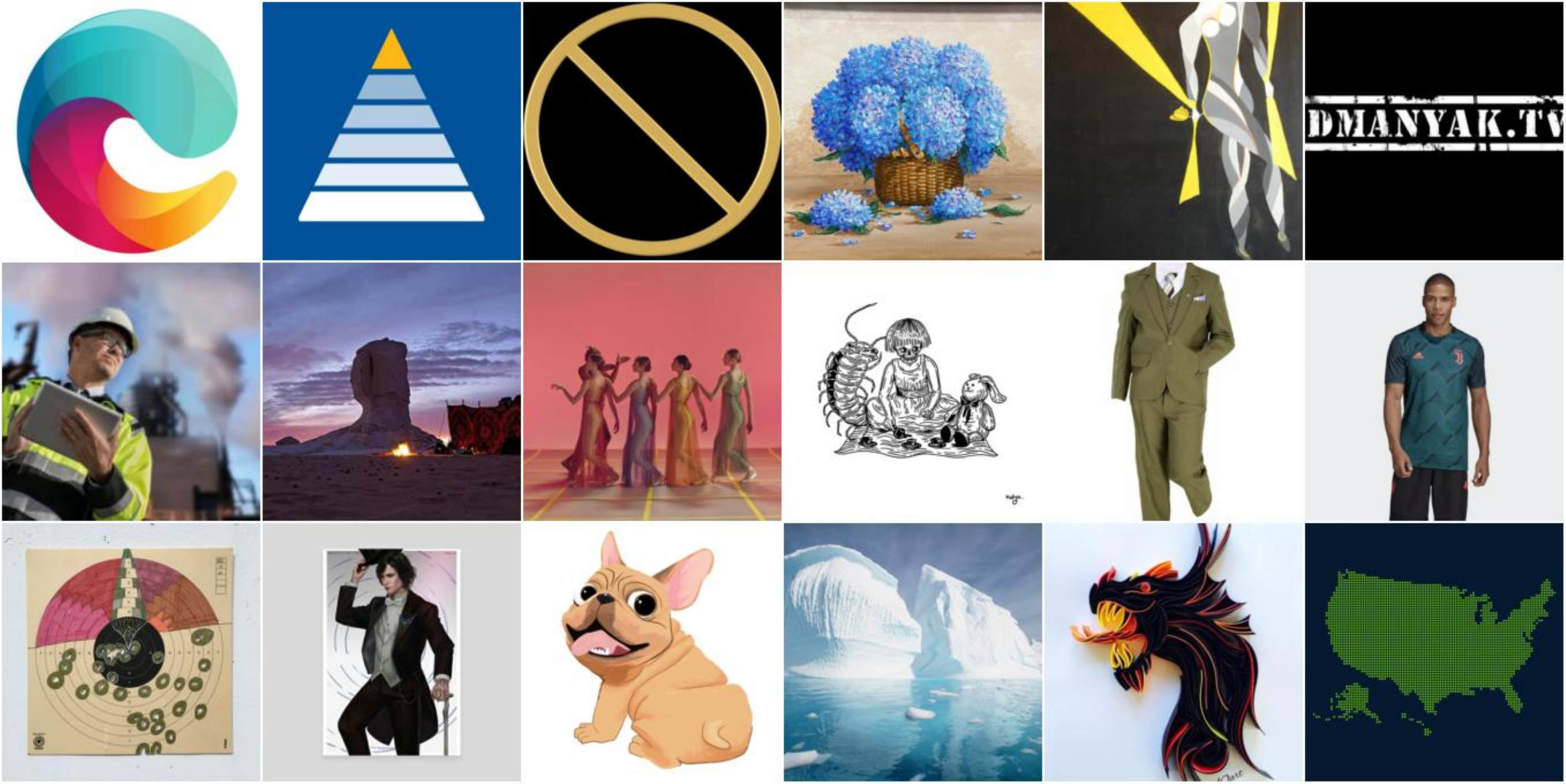}\\
    \end{center}
    \caption{\textbf{Top}: A sample of \textit{fake} W\"urstchen 2 \cite{wurstchen} images we generated for our dataset. \textbf{Bot}: Corresponding \textit{real} images found via RIS.}
    \label{fig:wurstchen_samples}
\end{figure*}

\begin{figure*}
    \centering
    \centering
    \begin{center}
        \Large{Fake (SD \pkmn{} Diffusers \cite{pkmn_diffusers})}\\\vspace{0.25cm}
        \includegraphics[width=\linewidth]{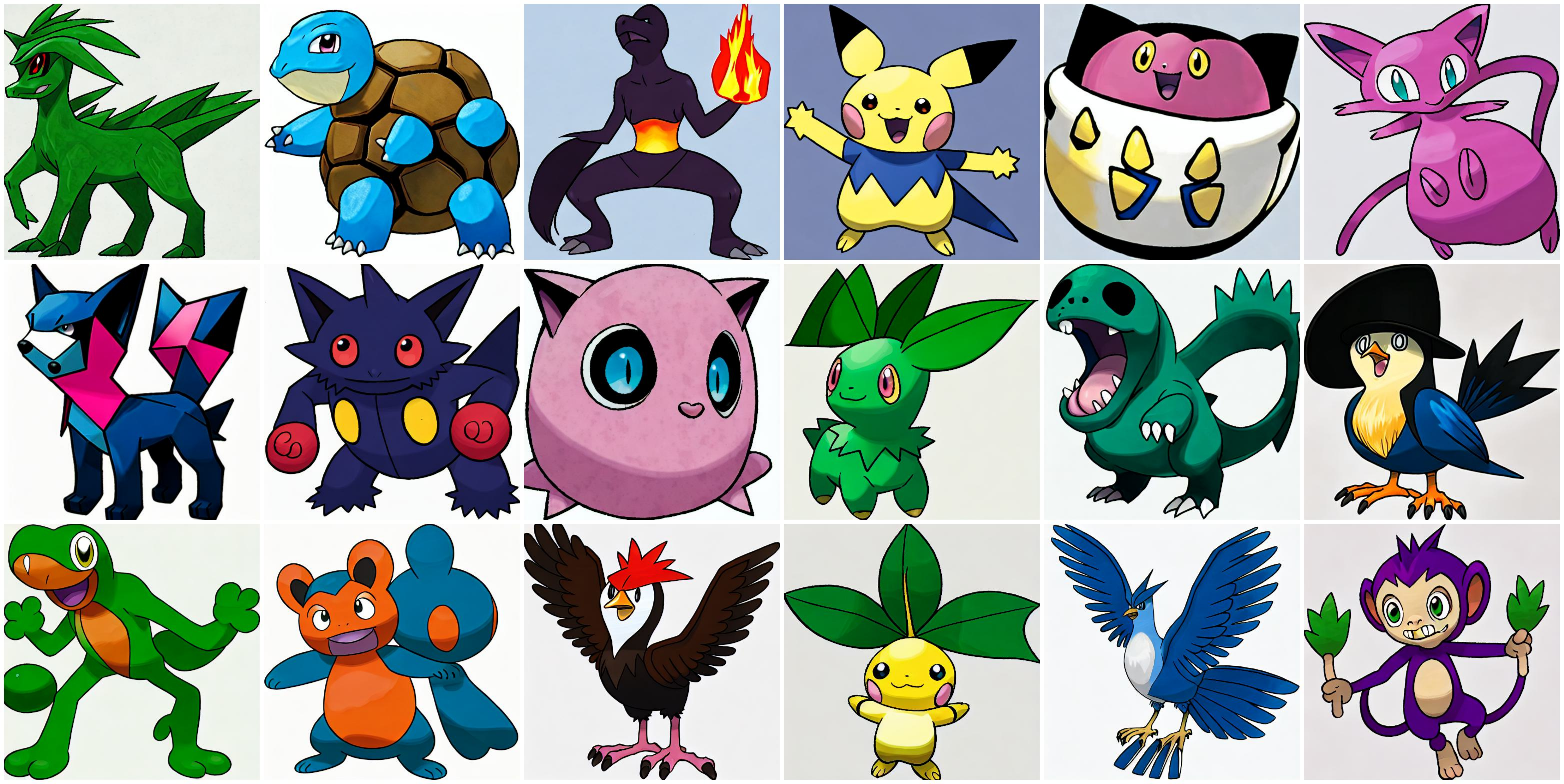}\\\vspace{0.8cm}
        \Large{Real (\pkmn{} \cite{pokemon})}\\\vspace{0.25cm}
        \includegraphics[width=\linewidth]{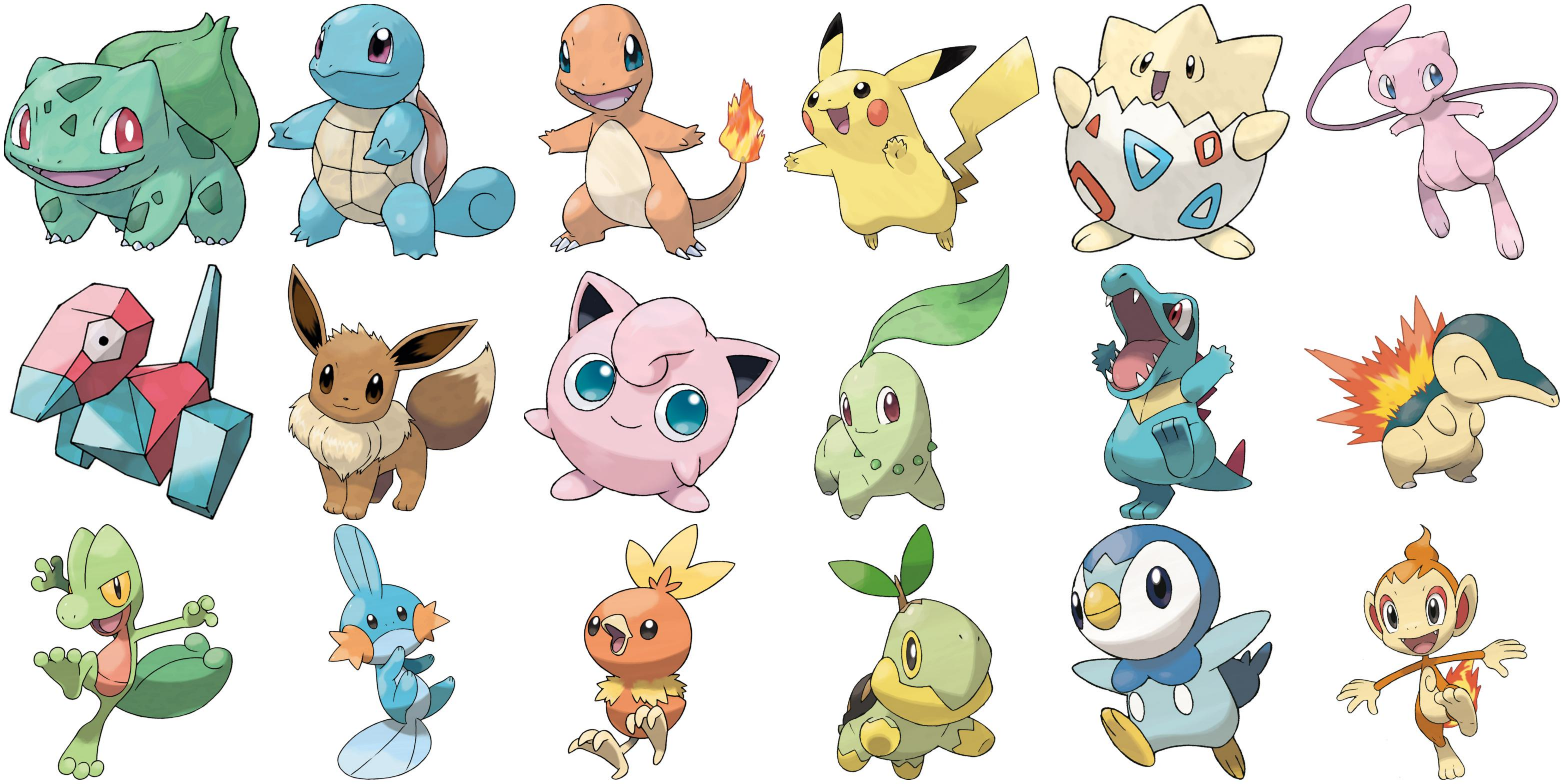}\\
    \end{center}
    \caption{\textbf{Top}: A sample of \textit{fake} \pkmn{} images we generated for our dataset. \textbf{Bot}: Corresponding \textit{real} images. The above \textit{fake} images were generated using the BLIP \cite{blip} captions of these \textit{real} images as prompts \cite{pinkney2022pokemon}.}
    \label{fig:pokemon_samples}
\end{figure*}

\begin{figure*}
    \centering
    \centering
    \begin{center}
        \Large{Fake (Animagine XL 2.0 \cite{animagine})}\\\vspace{0.25cm}
        \includegraphics[width=\linewidth]{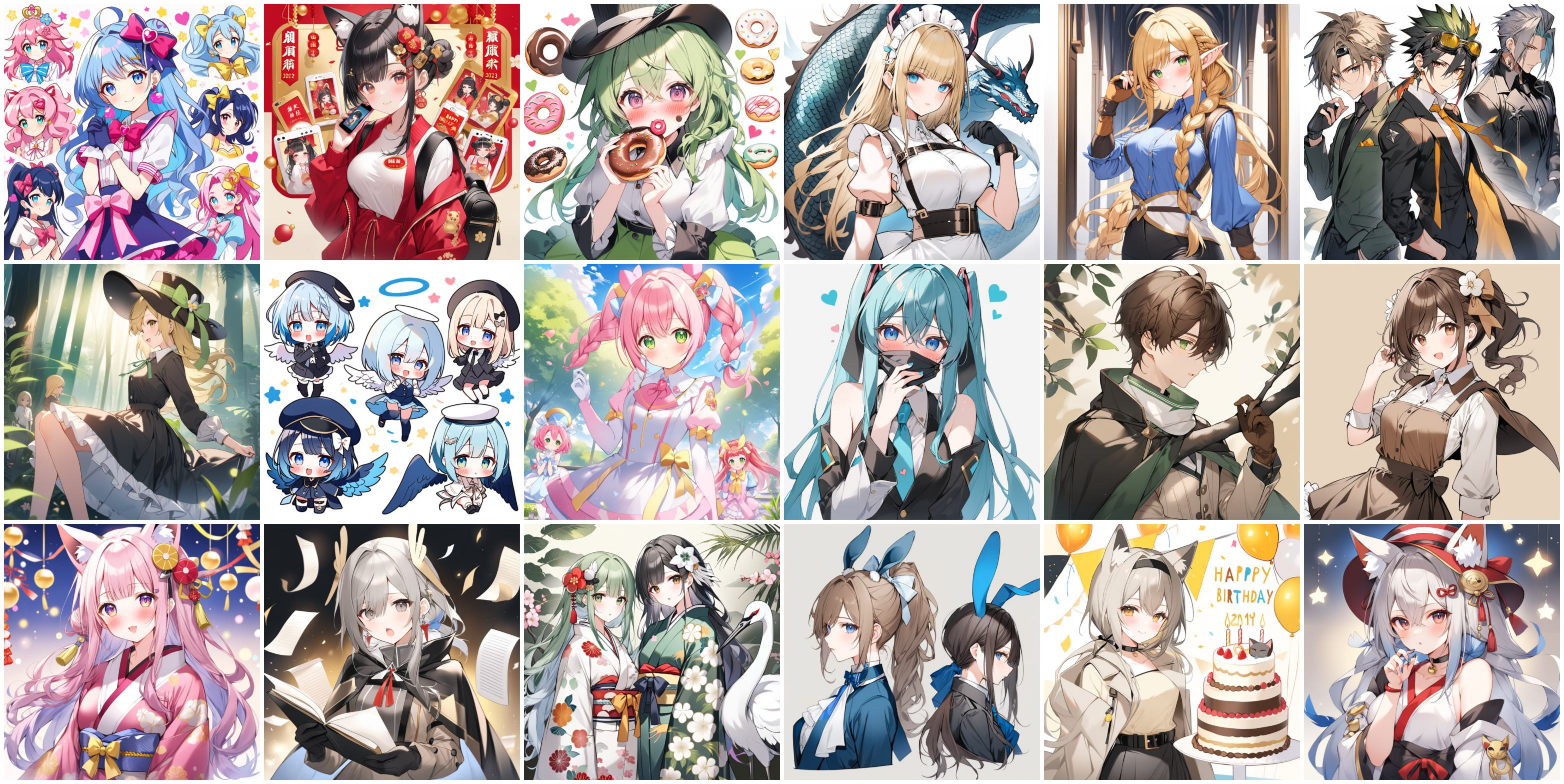}\\\vspace{0.8cm}
        \Large{Real (Danbooru \cite{danbooru2022})}\\\vspace{0.25cm}
        \includegraphics[width=\linewidth]{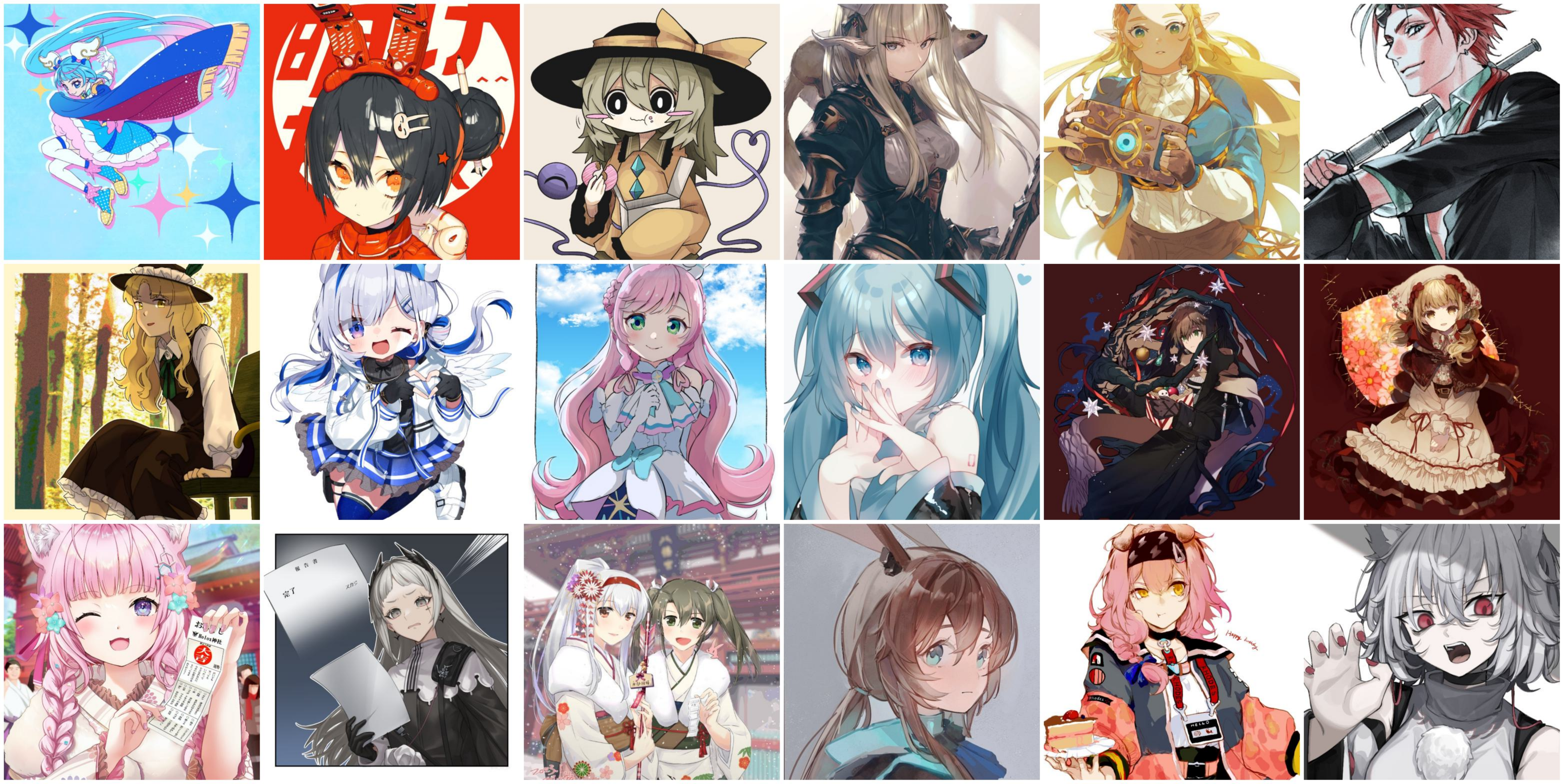}\\
    \end{center}
    \caption{\textbf{Top}: A sample of \textit{fake} Anime images we generated for our dataset. \textbf{Bot}: Corresponding \textit{real} images from the (SFW) Danbooru 2022 dataset \cite{danbooru2022}. The above \textit{fake} images were generated using the Danbooru tags of these \textit{real} images as prompts.}
    \label{fig:anime_samples}
\end{figure*}

\clearpage

\clearpage
\clearpage
 
\fi

\end{document}